\documentclass[a4paper,11pt]{report}

\usepackage{graphicx}
\usepackage{amsmath}
\usepackage{array}
\usepackage{float}
\usepackage{subfigure}
\usepackage{color}
\usepackage{listings}
\usepackage[utf8x]{inputenc}
\usepackage{wrapfig}
\usepackage{subeqnarray}
\usepackage{cancel}

\usepackage{amsfonts}  
\usepackage{amsthm}	

\newtheorem{mytheorem}{Theorem}

\usepackage{fancyhdr}               
\usepackage{titlesec}

\usepackage{etoolbox}
\makeatletter
\patchcmd{\ttlh@hang}{\parindent\z@}{\parindent\z@\leavevmode}{}{}
\patchcmd{\ttlh@hang}{\noindent}{}{}{}
\makeatother

\newcommand{\Paragraph}[1]{\underline{\textbf{#1}}\\ \noindent}

\usepackage{color}
\usepackage{colortbl}
\usepackage{lastpage}
\usepackage{array}
\usepackage{multirow}
\usepackage{datetime}

\usepackage{bm}   
\usepackage{amssymb,amsmath}

\titleformat{\chapter}
{\titlerule
\vspace{.8ex}%
\LARGE \bfseries}
{\thechapter.}{.5em}{}[\titlerule]

\titleformat{\section}
{\vspace{.8ex}%
\Large \bfseries}
{\thesection.}{.5em}{}

\newcommand{\HRuleW}{\rule{\linewidth}{1.2mm}}

\newcommand{\TRInfo}[6]{
    \begin{titlepage}
    \begin{center}

        \HRuleW \\[0.4cm]
        \LARGE \textbf{#1}\\[0.2cm]
        \Large #2\\
        \HRuleW

        \vspace{2.0cm}

        \textbf{Technical report \##5}\\
        \textbf{#6}

    \end{center}

        \vfill
        \begin{minipage}[b]{0.4\textwidth}
        \begin{flushleft}
        \includegraphics[width=0.3\textwidth]{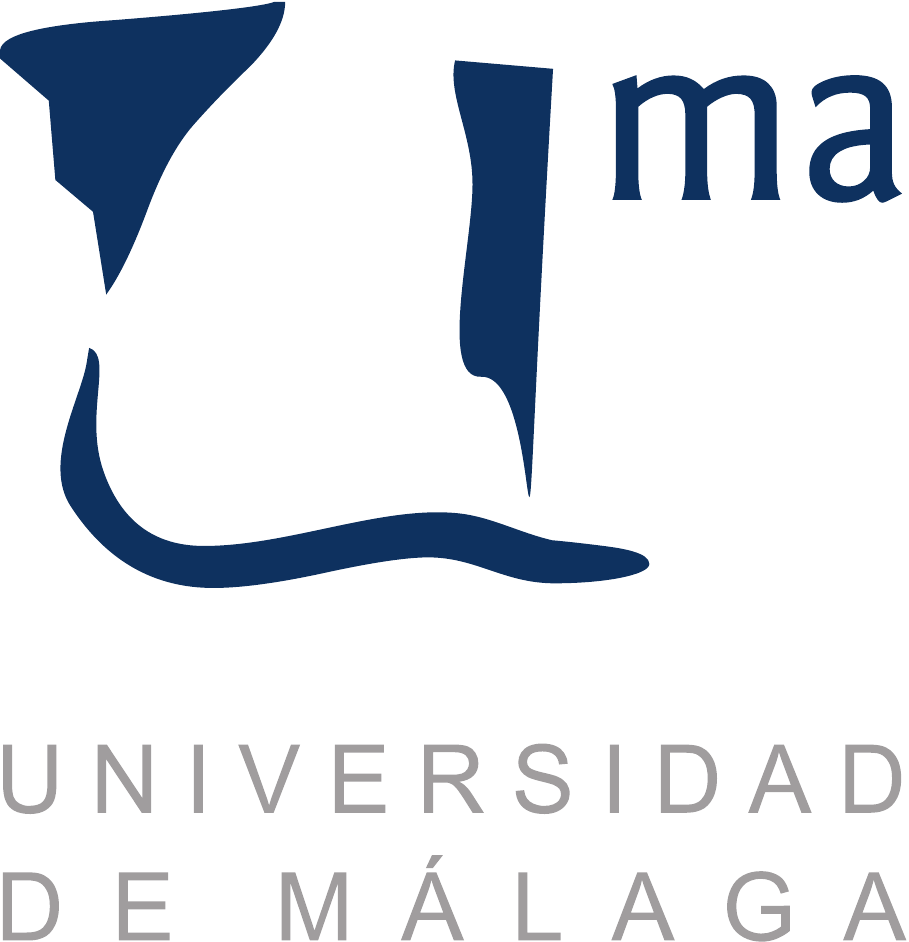}
        \end{flushleft}
        \end{minipage}
        \hfill
        \begin{minipage}[b]{0.45\textwidth}
        \begin{flushleft}
        \scriptsize \textbf{Málaga, \today}\\\medskip
        MAPIR: Grupo de Percepción y Robótica\\
        Dpto. de Ingeniería de Sistemas y Automática\\\medskip
        ETS Ingeniería Informática\\
        Universidad de Málaga\\
        Campus de Teatinos s/n \textbf - 29071 Málaga\\
        Tfno: 952132724 \textbf - Fax: 952133361\\
        http://mapir.isa.uma.es/ \textbf - http://www.isa.uma.es
        \end{flushleft}
        \end{minipage}

        \HRuleW

    \end{titlepage}

    \newcommand{\TRTitle}{#1}
    \newcommand{\TRPartners}{#3}
    \newcommand{\TRRef}{Technical report \##5}
} %

\newcommand\HdrTable{
\begin{tabular}{|m{0.75\textwidth}|m{0.25\textwidth}|}\hline
\centering \textbf{\textsc{\TRTitle}} \newline\newline \centering \TRPartners &
\scriptsize
 \TRRef \newline
 ~ \newline
 Dpto. de Ingeniería de Sistemas y Automática \newline
 \texttt{http://mapir.isa.uma.es/}
\\ \hline
\end{tabular}%
}

\newcommand{\PageStyle}{
    \pagestyle{fancyplain} %
    \fancyhead[CE,CO]{\HdrTable} %
} %

\newcommand{\E}{{\bm{\varepsilon}}}
\newcommand{\W}{{\bm{\omega}}}

\newcommand{\A}{{\mathbf{A}}}
\newcommand{\B}{{\mathbf{B}}}
\newcommand{\D}{{\mathbf{D}}}
\newcommand{\I}{{\mathbf{I}}}

\newcommand{\Pone}{\mathbf{P}_1}
\newcommand{\Ptwo}{\mathbf{P}_2}

\newcommand{\DEL}{{\bm{\delta}}}

\newcommand{\hatop}[1]{#1^\wedge}

\usepackage[bookmarks=true,breaklinks=true,linkbordercolor={0 0 1}]{hyperref}

\begin{document}

\definecolor{lightgray}{rgb}{0.93,0.93,0.93}
\definecolor{darkgray}{rgb}{0.5,0.5,0.5}
\lstset{language=C++}
\lstset{backgroundcolor=\color{lightgray}}
\lstset{keywordstyle=\color{blue}\bfseries\emph}
\lstset{commentstyle=\color{darkgray}\emph}
\lstset{basicstyle=\ttfamily\scriptsize} 

\title{}
\author{\\\\~\\Technical Report\\~\\
of M\'alaga, Spain}

\TRInfo{A tutorial on $\mathbf{SE}(3)$ transformation parameterizations and on-manifold optimization}
{José Luis Blanco Claraco\\ \begin{small} \texttt{jlblanco@ual.es} \\ \url{https://w3.ual.es/personal/jlblanco/} \end{small} }
{MAPIR Group}{}{012010}{Last update: \ddmmyyyydate \today }
\PageStyle

\pagestyle{plain}

\begin{abstract}
An arbitrary rigid transformation in $\mathbf{SE}(3)$ can be separated into two parts, namely,
a translation and a rigid rotation.
This technical report reviews, under a unifying viewpoint, three common alternatives to representing the
rotation part: sets of three (yaw-pitch-roll) Euler angles, orthogonal rotation matrices from
$\mathbf{SO}(3)$ and quaternions.
It will be described:
(i) the equivalence between these representations and the formulas for transforming one to each other
(in all cases considering the translational and rotational parts as a whole),
(ii) how to compose poses with poses and poses with points in each representation and
(iii) how the uncertainty of the poses (when modeled as Gaussian distributions)
is affected by these transformations and compositions.
Some brief notes are also given about the Jacobians required to implement least-squares optimization
on manifolds, an very promising approach in recent engineering literature.
The text reflects which MRPT C++ library\footnote{\texttt{https://www.mrpt.org/}} functions implement
each of the described algorithms.
All formulas and their implementation have been thoroughly validated 
by means of unit testing and numerical estimation of the Jacobians.
\end{abstract}

\newpage

\begin{wrapfigure}{r}{3cm}
    \includegraphics[width=2.7cm]{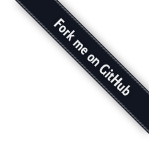}
\end{wrapfigure}

Feedback and contributions are welcome in: \\
\url{https://github.com/jlblancoc/tutorial-se3-manifold}.

\vspace{2cm}

\textbf{History of document versions:}
\begin{itemize}
 \item Apr/2022: Fixed missing transpose in Eq.~(\ref{eq:oplus.ab.wrt.a}), which was correctly set in Eq.~(\ref{eq:jacob_eD.2}) (Thanks to Frisch)
 \item Mar/2021: Added Eqs.~(\ref{eq:rodrigues_coordinate})--(\ref{eq:axis.angle.PR}) and 
 Eqs.~(\ref{eq:rotation_angle})--(\ref{eq:axis_around_pi}) (Thanks to \href{https://github.com/nurlanov-zh
 }{Nurlanov Zhakshylyk}).
 \item May/2020: Fix wrong terms in Eq.~\ref{eq:jacob.f_qrir_p} and a typo in Eq.~\ref{eq:jacob.inverse.quat} (Thanks to \href{https://github.com/YB27}{@YB27}).

 \item Mar/2019: Added new sections: \S\ref{sect:jacob.DinvP2invP2}, \S\ref{sect:eq:jacob_dLnSE3_dSE3}. Removed incorrect transpose in Eq.~\ref{eq:oplus.ab.wrt.a}. Add appendix~\ref{ch:apx:se2} for SE(2) GraphSLAM. Formally define pseudo exponential and logarithm maps in \S\ref{eq:exp.log.se3}.

 \item Oct/2018: Yaw-Pitch-Roll to Quaternion Jacobian gets its own equation number for easier reference: Eq.~\ref{eq:jac_quat_ypr}. Better references for boxplus and boxminus operators in \S\ref{ch:se3_optim}. Added \S\ref{sect:jacob.p6_p12}. Added exponential and logarithms for SO(3) in quaternion form to \S\ref{eq:exp.log.so3}.
	
 \item 29/May/2018: Adoption of the widespread notation for the "hat" and "vee" Lie group operators, as introduced now in \S\ref{sect:mat_deriv:ops}.

 \item 25/Mar/2018: Fixed minor typos.
  
 \item 28/Nov/2017: Fixed typos in \S\ref{eq:jacob_pmA_e_D} (Thanks to \href{https://github.com/gblack007}{@gblack007}).

 \item 10/Nov/2017: Sources published in GitHub: \\
    \url{https://github.com/jlblancoc/tutorial-se3-manifold}.

 \item 18/Oct/2017: Corrected typos in equations of \S\ref{sect:point_inv:quat} (Thanks to Otacílio Neto for detecting and reporting it).

 \item 18/Oct/2016: C++ code excerpts updated to MRPT 1.3.0 or newer.

 \item 8/Dec/2015: Fixed a few typos in matrix size legends.

 \item 21/Oct/2014: Fixed a typo in Eq.~\ref{eq:rodrigues_ln} (Thanks to Tanner Schmidt for reporting).

 \item 9/May/2013: Added the Jacobian of the SO(3) logarithm map, in \S\ref{sect:eq:jacob_dLnROT_dROT}.

 \item 14/Aug/2012: Added the explicit formulas for the logarithm map of SO(3) and SE(3),
    fixed error in Eq.~(\ref{eq:jacob_p_min_eD_e}), explained the equivalence between the yaw-pitch-roll
and roll-pitch-yaw forms and introduction of the $\left[ \log \mathbf{R} \right]^\vee$ notation
when discussing the logarithm maps.

 \item 12/Sep/2010: Added more Jacobians (\S\ref{sect:jacob_eDp},
\S\ref{sect:jacob_eDp_inv}, \S\ref{sect:jacob_De}),
         the Appendix \ref{ch:apx:cv} and approximation in \S\ref{eq:jacob_A_e_D_p}.

 \item 1/Sep/2010: First version.
\end{itemize}

\vfill
\textbf{Notice:} \\

Part of this report was also published within chapter 10 and appendix IV of the
book \cite{madrigal2012slambook}.

\vspace{1cm}

\begin{scriptsize}
\begin{center}
\includegraphics[width=3cm]{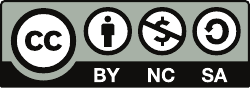}
\\
This work is licensed under
Attribution-NonCommercial-ShareAlike 4.0 International (CC BY-NC-SA 4.0)
License.
\end{center}

\end{scriptsize}

\newpage

\tableofcontents

\chapter{Rigid transformations in 3D}

\section{Basic definitions}
\label{sect:basic}

This report focuses on geometry for the most interesting case of an Euclidean space in
engineering: the three-dimensional space $\mathbb{R}^3$.
Over this space one can define an arbitrary \emph{transformation} through
a function or mapping:

\begin{equation}
 f: \quad \mathbb{R}^3 \rightarrow \mathbb{R}^3
\end{equation}

For now, assume that $f$ can be any $3 \times 3$ matrix $\mathbf{R}$,
such as the mapping function
from a point $\mathbf{x_1}=[x_1 ~ y_1 ~ z_1]^\top$ to $\mathbf{x_2}=[x_2 ~ y_2 ~ z_2]^\top$ is simply:

\begin{equation}
\left(
\begin{array}{c}
 x_2 \\ y_2 \\ z_2
\end{array}
\right)
=
 \mathbf{x_2}
=
\mathbf{R} \mathbf{x_1}
=
\mathbf{R}
\left(
\begin{array}{c}
 x_1 \\ y_1 \\ z_1
\end{array}
\right)
\end{equation}

The set of all invertible $3 \times 3$ matrices forms the
general linear group $\mathbf{GL}(3,\mathbb{R})$.
From all the infinite possibilities for $\mathbf{R}$, the set
of \emph{orthogonal matrices} with determinant of $\pm 1$
(i.e. $\mathbf{R}\mathbf{R}^\top=\mathbf{R}^\top\mathbf{R} = \mathbf{I_3}$)
forms the so called \emph{orthogonal group} or $\mathbf{O}(3) \subset \mathbf{GL}(3,\mathbb{R})$.
Note that the group operator is the standard matrix product, since multiplying
any two matrices from $\mathbf{O}(3)$ gives another member of $\mathbf{O}(3)$.
All these matrices define \emph{isometries}, that is, transformations that
preserve distances between any pair of points.
From all the isometries, we are only interested here in those with a
determinant of $+1$, named \emph{proper} isometries.
They constitute the group of proper orthogonal transformations, or
\emph{special orthogonal group} $\mathbf{SO}(3) \subset \mathbf{O}(3)$
\cite{gallier2001geometric}.

The group of matrices in $\mathbf{SO}(3)$ represents \emph{pure} rotations only.
In order to also handle translations, we can take into account $4 \times 4$
transformation matrices $\mathbf{T}$ and extend 3D points with a fourth
\emph{homogeneous} coordinate (which in this report will be always the unity),
thus:

\begin{eqnarray}
\left(
\begin{array}{c}
 \mathbf{x_2} \\ 1
\end{array}
\right)
&=&
\mathbf{T}
\left(
\begin{array}{c}
 \mathbf{x_1} \\ 1
\end{array}
\right)
\nonumber \\
\label{eq:x2_T_x1}
\left(
\begin{array}{c}
 x_2 \\ y_2 \\ z_2 \\ 1
\end{array}
\right)
&=&
\left(
\begin{array}{c|c}
\mathbf{R} &
 \begin{array}{c}
   t_x \\ t_y \\ t_z
 \end{array} \\
\hline
 \begin{array}{ccc}
  0 & 0 & 0
 \end{array}
 &
 1
\end{array}
\right)
\left(
\begin{array}{c}
 x_1 \\ y_1 \\ z_1 \\ 1
\end{array}
\right)
\\
\mathbf{x_2} &=& \mathbf{R} \mathbf{x_1} +
\left(
t_x ~~ t_y ~~ t_z
\right)^\top \nonumber
\end{eqnarray}

In general, any invertible $4 \times 4$ matrix belongs to the
general linear group $\mathbf{GL}(4,\mathbb{R})$, but in the particular case of
the so defined set of transformation matrices $\mathbf{T}$
(along with the group operation of matrix product),
they form the group of affine rigid motions which, with \emph{proper} rotations
($|\mathbf{R}|=+1$), is denoted as the special Euclidean group $\mathbf{SE}(3)$.
It turns out that $\mathbf{SE}(3)$ is also a Lie group, and
a manifold with structure $\mathbf{SO}(3) \times \mathbb{R}^3$ (see \S\ref{sect:lie:linear}).
Chapters \ref{chap:mat_deriv}-\ref{ch:se3_optim} will explain
what all this means and how to exploit it in engineering optimization problems.

\begin{figure}[h!]
\centering
\includegraphics[width=0.70\textwidth]{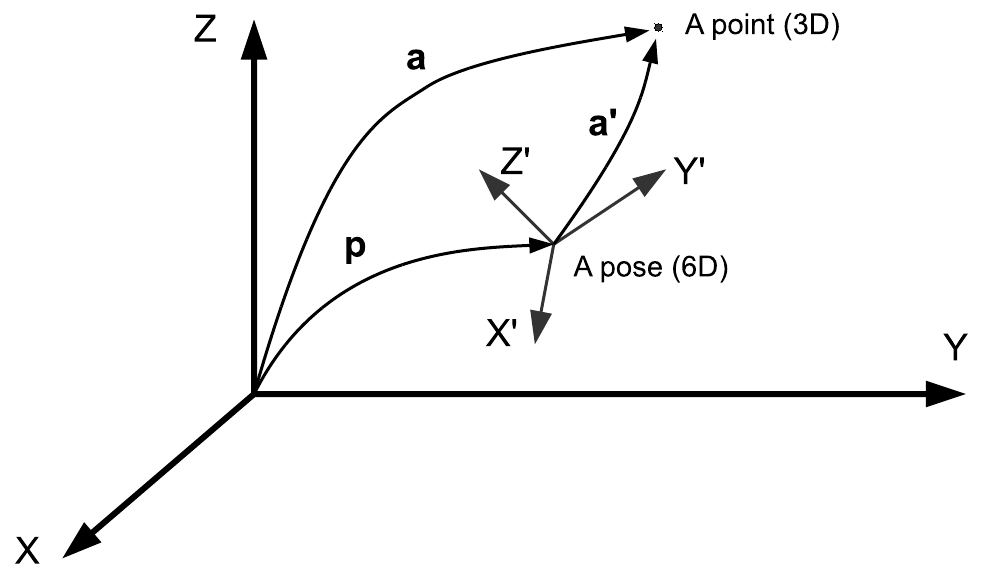}
\caption{Schematic representation of a 6D pose $\mathbf{p}$ and its role in defining
the relative coordinates $\mathbf{a}'$ of the 3D point $\mathbf{a}$.}
\label{fig:1}
\end{figure}

In this report we will refer to $\mathbf{SE}(3)$ transformations as \emph{poses}.
As seen in Eq.~(\ref{eq:x2_T_x1}), a pose can be described by means of a 3D translation
plus an orthonormal vector base (the columns of $\mathbf{R}$),
or coordinate frame, relative to any other arbitrary coordinate reference system.
The overall number of degrees of freedom is six, hence they can be also referred
to as \emph{6D poses}.
The Figure~\ref{fig:1} illustrates this definition, where the pose $\mathbf{p}$ is represented
by the axes $\{\mathbf{X}',\mathbf{Y}',\mathbf{Z}' \}$ with respect to a reference frame
$\{\mathbf{X},\mathbf{Y},\mathbf{Z} \}$.

Given a 6D pose $\mathbf{p}$ and a 3D point $\mathbf{a}$, both relative to some arbitrary
global frame of reference, and being $\mathbf{a}'$ the coordinates of $\mathbf{a}$ relative
to $\mathbf{p}$, we define
the composition $\oplus$ and inverse composition $\ominus$ operations as follows:

\begin{eqnarray*}
\mathbf{a} & \equiv & \mathbf{p} \oplus \mathbf{a}'   ~~~~~~~ \textrm{Pose composition} \\
\mathbf{a'} & \equiv & \mathbf{a} \ominus \mathbf{p}  ~~~~~~~~ \textrm{Pose inverse composition} \\
\end{eqnarray*}

These operations are intensively applied in a number of robotics and computer vision
problems, for example, when computing the relative position of a 3D visual landmark
with respect to a camera while computing the perspective projection of the landmark
into the image plane.

\begin{figure}[h!]
\centering
\subfigure[The pose $\mathbf{p1}$]{\includegraphics[width=0.4\textwidth]{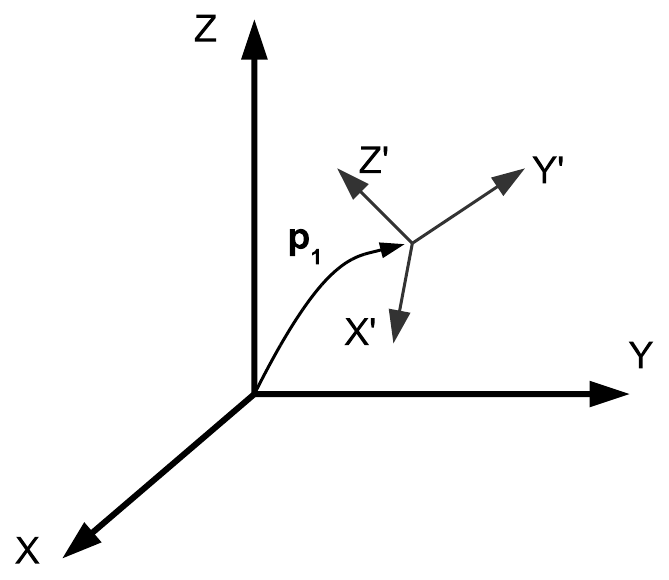}}
\subfigure[The pose $\mathbf{p2}$]{\includegraphics[width=0.4\textwidth]{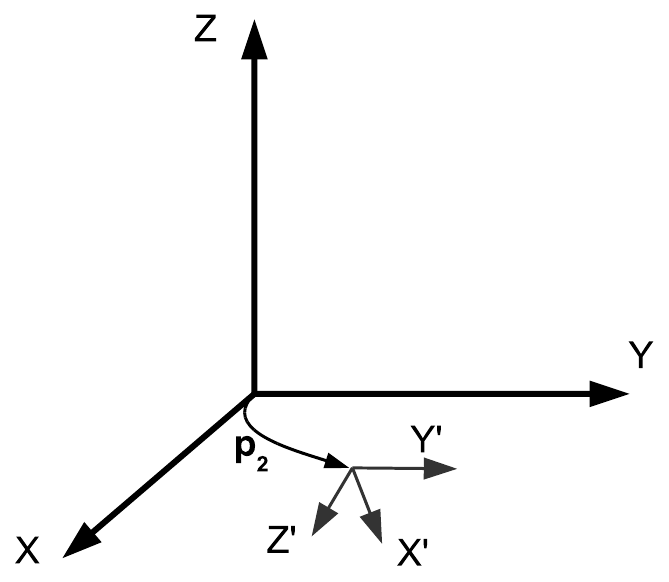}} \\
\subfigure[Composition $\mathbf{p1} \oplus \mathbf{p2}$]{\includegraphics[width=0.6\textwidth]{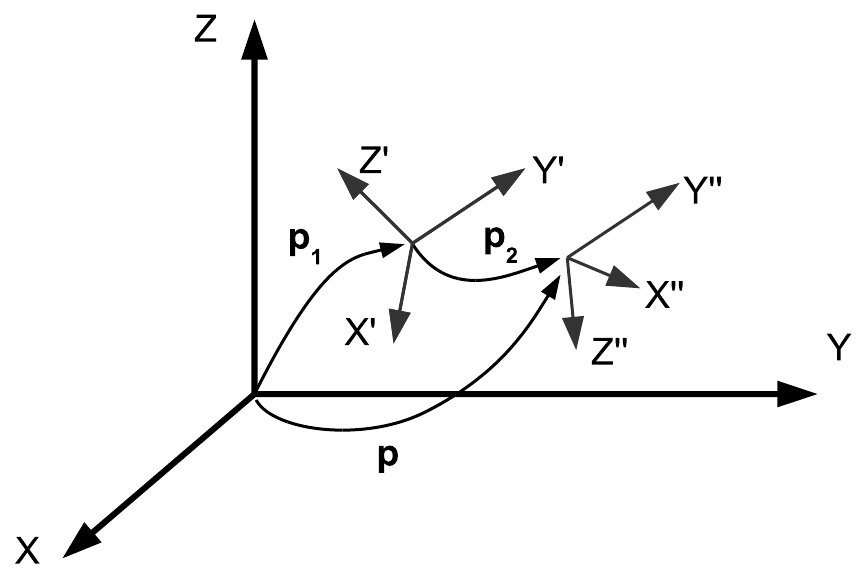}}
\caption{The composition of two 6D poses $\mathbf{p1}$ and $\mathbf{p2}$ leads to $\mathbf{p}$.}
\label{fig:comp_2poses}
\end{figure}

The composition operators can be also applied to pairs of 6D poses (above we described a combination
of \emph{6D poses} and {3D points}).
The meaning of composing two poses $\mathbf{p1}$ and $\mathbf{p2}$
obtaining a third pose $\mathbf{p} = \mathbf{p1} \oplus \mathbf{p2}$
is that of concatenating the transformation of the second pose to the reference system
\emph{already transformed} by the first pose.
This is illustrated in Figure~\ref{fig:comp_2poses}.

The inverse pose composition can be also applied to 6D poses, in this case meaning that
the pose $\mathbf{p}$ (in global coordinates) ``is seen'' as $\mathbf{p2}$ with respect
to the reference frame of $\mathbf{p1}$ (this one, also in global coordinates), a
relationship expressed as $\mathbf{p2} = \mathbf{p} \ominus \mathbf{p1}$.

Up to this point, poses, pose/point and pose/pose compositions have been mostly described
under a purely geometrical point of view.
The next section introduces some of the most commonly employed parameterizations.

\newpage

\section{Common parameterizations}

\subsection{3D translation plus yaw-pitch-roll (3D+YPR)}

A 6D pose $\mathbf{p_6}$ can be described as a displacement in 3D plus a rotation defined by
means of a specific case of Euler angles: yaw ($\phi$), pitch
($\chi$) and roll ($\psi$), that is:

\begin{eqnarray}
\mathbf{p_6} &=& [x ~ y ~ z ~ \phi ~ \chi ~ \psi]^\top
\end{eqnarray}

\begin{figure}[h]
\centering
\includegraphics[width=0.40\textwidth]{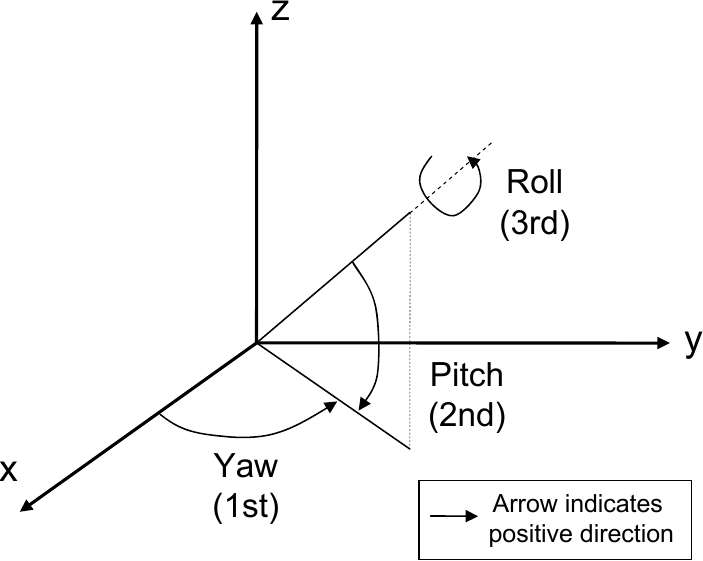}
\caption{A common convention for the angles yaw, pitch and roll.}
\label{fig:ypr}
\end{figure}

The geometrical meaning of the angles is represented in Figure~\ref{fig:ypr}.
There are other alternative conventions about triplets of angles to represent a rotation in 3D, but
the one employed here is the one most commonly used in robotics.

Note that the overall rotation is represented as a sequence of three individual rotations,
each taking a different axis of rotation.
In particular, the order is: yaw around the Z axis, then pitch around the \emph{modified} Y axis,
then roll around the \emph{modified} X axis. It is also common to find in the literature
the roll-pitch-yaw (RPY) parameterization (versus YPR), where rotations apply over the same angles (e.g. yaw
around the Z axis) but in inverse order and around the \emph{unmodified} axes instead of
the successively modified axes of the yaw-pitch-roll form. In any case, it can be shown that
the numeric values of the three rotations are identical for any given 3D rotation \cite{madrigal2012slambook},
thus both forms are completely equivalent.

This representation is the most compact since it only requires 6 real parameters
to describe a pose (the minimum number of parameters, since a pose has 6 degrees of freedom).
However, in some applications it may be more advantageous to employ other representations,
even at the cost of maintaining more parameters.

\subsubsection{Degenerate cases: gimbal lock}
\label{sect:ypr:gimbal}

One of the important disadvantages of the yaw-pitch-roll representation of rotations is the existence
of two degenerate cases, specifically, when pitch ($\chi$) approaches $\pm 90^\circ$. In this case,
it is easy to realize that a change in roll becomes a change in yaw.

This means that, for $ \chi = \pm 90^\circ$, there is not a unique correspondence between any possible
rotation in 3D and a triplet of yaw-pitch-roll angles.
The practical consecuences of this characteristic is the need for detecting and handling these
special cases, as will be seen in some of the transformations described later on.

\subsubsection{Implementation in MRPT}

Poses based on yaw-pitch-roll angles are implemented in the C++ class \texttt{mrpt::poses::CPose3D}:

\begin{lstlisting}
#include <mrpt/poses/CPose3D.h>
using namespace mrpt::poses;
using mrpt::utils::DEG2RAD;

CPose3D   p(1.0 /* x */,2.0 /* y */,3.0 /* z */,
            DEG2RAD(30.0) /* yaw */, DEG2RAD(20.0) /* pitch */, DEG2RAD(90.0) /* roll */ );
\end{lstlisting}


\subsection{3D translation plus quaternion (3D+Quat)}

A pose $\mathbf{p_7}$ can be also described with a displacement in 3D plus a rotation
defined by a quaternion, that is:

\begin{eqnarray}
\mathbf{p_7} &=& [x ~ y ~ z ~  q_r ~ q_x ~ q_y ~ q_z ] ^ \top
\end{eqnarray}

\noindent where the unit quaternion elements are $[q_r, (q_x,q_y,q_z)]$. A useful interpretation of quaternions
is that of a rotation of $\theta$ radians around the axis defined by the vector $\vec{v} = (v_x,v_y,v_z) \propto (q_x,q_y,q_z)$.
The relation between $\theta$, $\vec{v}$ and the elements in the quaternion is:

\begin{equation*}
\begin{array}{cc}
q_r = \cos\frac{\theta}{2}  &
  \begin{array}{rcl}
    q_x &=& \sin\frac{\theta}{2} v_x  \\
    q_y &=& \sin\frac{\theta}{2} v_y  \\
    q_z &=& \sin\frac{\theta}{2} v_z
  \end{array}
\end{array}
\end{equation*}

\begin{figure}[h]
\centering
\includegraphics[width=0.50\textwidth]{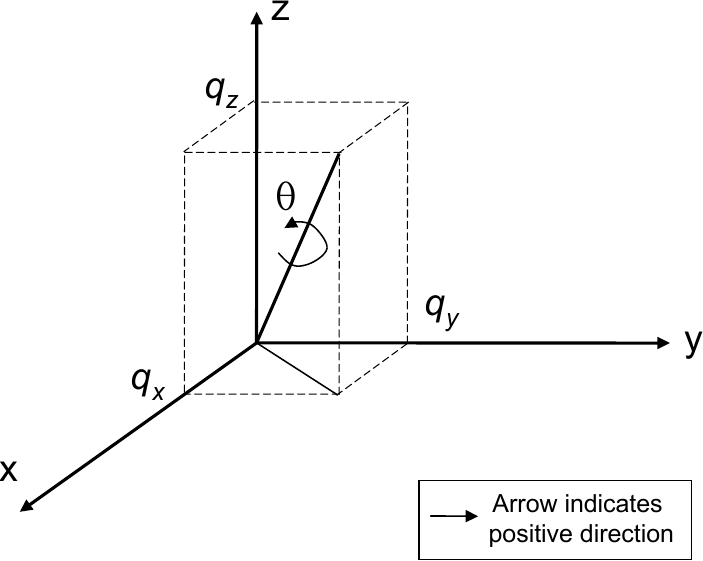}
\caption{A quaternion can be seen as a rotation around an arbitrary 3D axis.}
\label{fig:quat}
\end{figure}

This interpretation is also represented in Figure~\ref{fig:quat}.
The convention is $q_r$ (and thus $\theta$) to be non-negative.
A quaternion has 3 degrees of freedom in spite of having four components
due to the unit length constraint, which can be interpreted as a unit hyper-sphere, 
hence its topology being that of the special unitary group $SU(2)$, diffeomorphic to $S(3)$.

\subsubsection{Implementation in MRPT}

Poses based on quaternions are implemented in the class \texttt{mrpt::poses::CPose3DQuat}.
The quaternion part of the pose is always normalized (i.e. $q_r^2+q_x^2+q_y^2+q_z^2=1$).

\begin{lstlisting}
#include <mrpt/poses/CPose3DQuat.h>
using namespace mrpt::poses;
using namespace mrpt::math;

CPose3DQuat   p(1.0 /* x */,2.0 /* y */,3.0 /* z */,
                CQuaternionDouble(1.0 /* qr */,  0.0,0.0,0.0 /* vector part */) );
\end{lstlisting}

\subsubsection{Normalization of a quaternion}
\label{sect:quat:norm}

In many situations, the quaternion part of a 3D+Quat 7D representation
of a pose may drift away of being unitary.
This is specially true if each component of the quaternion is estimated
independently, such as within a Kalman filter or
any other Gauss-Newton iterative optimizer
(for an alternative, see \S\ref{ch:se3_optim}).

The normalization function is simply:

\begin{equation}
\mathbf{q'}(\mathbf{q})
=
\left(
\begin{array}{c}
 q_r' \\ q_x'\\ q_y'\\ q_z'
\end{array}
\right)
=
\frac{\mathbf{q}}{|\mathbf{q}|}
=
\frac{1}{(q_r^2+q_x^2+q_y^2+q_z^2)^{1/2}}
\left(
\begin{array}{c}
 q_r \\ q_x\\ q_y\\ q_z
\end{array}
\right)
\end{equation}

\noindent and its $4 \times 4$ Jacobian is given by:

\begin{equation}
\frac{\partial \mathbf{q'}(q_r,q_x,q_y,q_z)}{\partial q_r,q_x,q_y,q_z}
=
\frac{1}{(q_r^2+q_x^2+q_y^2+q_z^2)^{3/2}}
\left(
\begin{array}{cccc}
q_x^2 +q_y^2+q_z^2  & -q_r q_x   & -q_r q_y  & -q_r q_z  \\
-q_x q_r & q_r^2 +q_y^2+q_z^2 & -q_x q_y & - q_x q_z \\
-q_y q_r & -q_y q_x &  q_r^2 +q_x^2 +q_z^2 & -q_y q_z \\
-q_z q_r  & -q_z q_x & -q_z q_y  &  q_r^2+q_x^2+q_y^2
\end{array}
\right)
\end{equation}

%

\subsection{$4 \times 4$ transformation matrices}

Any rigid transformation in 3D can be described by means of a $4 \times 4$
matrix $\mathbf{P}$ with the following structure:

\begin{equation}
\mathbf{P}=\left(
  \begin{array}{ccc|c}
   & & & x \\
   & \mathbf{R} & & y \\
   & & & z \\ \hline
   0 & 0 & 0& 1
  \end{array}
\right)
\end{equation}

\noindent where the $3 \times 3$ orthogonal matrix $\mathbf{R} \in \mathbf{SO}(3)$
is the \emph{rotation matrix}\footnote{Also called direction cosine matrix (DCM).}
(the only part of $\mathbf{P}$ related to
the 3D rotation) and the vector $(x,y,z)$ represents the translational part of the 6D pose.
For such a matrix to be applicable to 3D points, they must be first represented in
homogeneous coordinates \cite{bloomenthal1994homogeneous} which, in our case, will consist in just considering a fourth,
extra dimension to each point which will be always equal to the unity -- examples of this will be discussed later on.

\subsubsection{Implementation in MRPT}

Transformation matrices themselves can be managed as any other normal $4\times 4$ matrix:

\begin{lstlisting}
#include <mrpt/utils/types_math.h>
using namespace mrpt::math;

CMatrixDouble44  P;
\end{lstlisting}

Note however that the 3D+YPR type \texttt{CPose3D} also holds a cached matrix representation
of the transformation which can be retrieved with \texttt{CPose3D::getHomogeneousMatrix()}.


\chapter{Equivalences between representations}

In this chapter the focus will be on the transformation of the rotational
part of 6D poses, since the 3D translational part is always represented as an
unmodified vector in all the parameterizations.

Another point to be discussed here is how the transformation between different
parameterizations affects the \emph{uncertainty} for the case of probability distributions over poses.
Assuming a multivariate Gaussian model, first order
linearization of the transforming functions
is proposed as a simple and effective approximation.
In general, having a multivariate Gaussian distribution of the variable
$\mathbf{x} \sim N(\bar{\mathbf{x}},\mathbf{\Sigma_x})$ (where $\bar{x}$ and $\mathbf{\Sigma_x}$
are its mean and covariance matrix, respectively), we can approximate
the distribution of $\mathbf{y} = f(\mathbf{x})$ as another Gaussian
with parameters:

\begin{eqnarray}
 \bar{\mathbf{y}} &=& f(\bar{\mathbf{x}}) \\
 \mathbf{\Sigma_y} &=&
\left.\frac{\partial f(\mathbf{x})}{\partial \mathbf{x}}\right|_{\mathbf{x}=\bar{\mathbf{x}}}
\mathbf{\Sigma_x}
\left.\frac{\partial f(\mathbf{x})}{\partial \mathbf{x}}\right|_{\mathbf{x}=\bar{\mathbf{x}}}^\top
\end{eqnarray}

Note that an alternative to this method is using the scaled unscented
transform (SUT) \cite{julier2002sut}, which may give more exact results for
large levels of the uncertainty but typically requires more computation time
and can cause problems for semidefinite positive (in contrast to definite positive)
covariance matrices.

\section{3D+YPR to 3D+Quat }
\label{sect:ypr2quat}

\subsection{Transformation}

Any given rotation described as a combination of yaw ($\phi$),
pitch ($\chi$) and roll ($\psi$) can
be expressed as a quaternion with components $(q_r, q_x,q_y,q_z)$
given by \cite{horn2001some}:

\begin{eqnarray}
 \mathbf{q}(\phi,\chi,\psi) &=& 
 \begin{bmatrix} 
	 q_r(\phi,\chi,\psi) \\
	 q_x(\phi,\chi,\psi) \\
	 q_y(\phi,\chi,\psi) \\
	 q_z(\phi,\chi,\psi)
 \end{bmatrix}
 \quad \quad 
 \mathbf{q}(\phi,\chi,\psi) : \mathcal{R}^3 \rightarrow \mathcal{R}^4
\\
  q_r(\phi,\chi,\psi) &=& \cos\frac{\psi}{2} \cos\frac{\chi}{2}  \cos\frac{\phi}{2}  +
  \sin\frac{\psi}{2} \sin\frac{\chi}{2} \sin\frac{\phi}{2}    \\
  q_x(\phi,\chi,\psi) &=& \sin\frac{\psi}{2} \cos\frac{\chi}{2}  \cos\frac{\phi}{2}  -
  \cos\frac{\psi}{2} \sin\frac{\chi}{2} \sin\frac{\phi}{2}    \\
  q_y(\phi,\chi,\psi) &=& \cos\frac{\psi}{2} \sin\frac{\chi}{2}  \cos\frac{\phi}{2}  +
  \sin\frac{\psi}{2} \cos\frac{\chi}{2} \sin\frac{\phi}{2}    \\
  q_z(\phi,\chi,\psi) &=& \cos\frac{\psi}{2} \cos\frac{\chi}{2}  \sin\frac{\phi}{2}  -
  \sin\frac{\psi}{2} \sin\frac{\chi}{2} \cos\frac{\phi}{2}
\end{eqnarray}

\subsubsection{Implementation in MRPT}

Transformation of a \texttt{CPose3D} pose object based on yaw-pitch-roll angles
into another of type \texttt{CPose3DQuat} based on quaternions can be done transparently
due the existence of an implicit conversion constructor:

\begin{lstlisting}
#include <mrpt/poses/CPose3D.h>
#include <mrpt/poses/CPose3DQuat.h>
using namespace mrpt::poses;

CPose3D  p6;
...
CPose3DQuat   p7 = CPose3DQuat(p6);  // Transparent conversion
\end{lstlisting}

\subsection{Uncertainty}

Given a Gaussian distribution over a 6D pose in yaw-pitch-roll form with
mean ${\mathbf{\bar{p}_6}}$ and being  $cov(\mathbf{p_6})$ its $6 \times 6$ covariance matrix,
the $7 \times 7$ covariance matrix of the equivalent quaternion-based form
is approximated by:

\begin{equation}
cov(\mathbf{p_7}) =
\frac{\partial \mathbf{p_7}(\mathbf{p_6}) }{\partial \mathbf{p_6}} ~
cov(\mathbf{p_6})  ~
\frac{\partial \mathbf{p_7}(\mathbf{p_6}) }{\partial \mathbf{p_6}}^\top
\end{equation}

\noindent where the Jacobian matrix is given by:

\begin{subeqnarray}
\label{eq:jac_p7_p6}
\frac{\partial \mathbf{p_7}(\mathbf{p_6}) }{\partial \mathbf{p_6}} &=&
\left(
\begin{array}{c|cc}
 \mathbf{I_3} & \mathbf{0_{3\times 3}} \\ \hline
 \mathbf{0_{4\times 3}} & \frac{\partial \mathbf{q}(\phi,\chi,\psi) }{\partial \{ \phi,\chi,\psi \} }
\end{array}
\right)_{7 \times 6}
\\
\slabel{eq:jac_quat_ypr}
\frac{\partial \mathbf{q}(\phi,\chi,\psi) }{\partial \{ \phi,\chi,\psi \} } &=& 
\begin{bmatrix}
	(ssc-ccs)/2 & (scs-csc)/2 & (css-scc)/2 \\
	-(csc+scs)/2 & -(ssc+ccs)/2 & (ccc+sss)/2 \\
	(scc-css)/2 & (ccc-sss)/2 & (ccs-ssc)/2 \\
	(ccc+sss)/2 & -(css+scc)/2 & -(csc+scs)/2
\end{bmatrix} _{4 \times 3}
\end{subeqnarray}

\noindent where the following abbreviations have been used:

\begin{equation*}
\begin{array}{ccc}
ccc = \cos\frac{\psi}{2}\cos\frac{\chi}{2}\cos\frac{\phi}{2}  &
ccs = \cos\frac{\psi}{2}\cos\frac{\chi}{2}\sin\frac{\phi}{2}  &
csc = \cos\frac{\psi}{2}\sin\frac{\chi}{2}\cos\frac{\phi}{2}  \\
 & ... & \\
scc = \sin\frac{\psi}{2}\cos\frac{\chi}{2}\cos\frac{\phi}{2}  &
ssc = \sin\frac{\psi}{2}\sin\frac{\chi}{2}\cos\frac{\phi}{2}  &
sss = \sin\frac{\psi}{2}\sin\frac{\chi}{2}\sin\frac{\phi}{2}  \\
\end{array}
\end{equation*}

\subsubsection{Implementation in MRPT}

Gaussian distributions over 6D poses described as yaw-pitch-roll and quaternions
are implemented in the classes \texttt{CPose3DPDFGaussian} and \texttt{CPose3DQuatPDFGaussian}, respectively.
Transforming between them is possible via an explicit transform constructor, which
converts both the mean and the covariance matrix:

\begin{lstlisting}
#include <mrpt/poses/CPose3DPDFGaussian.h>
#include <mrpt/poses/CPose3DQuatPDFGaussian.h>
using namespace mrpt::poses;

CPose3DPDFGaussian     p6( p6_mean, p6_cov );
...
CPose3DQuatPDFGaussian p7 = CPose3DQuatPDFGaussian(p6);  // Explicit constructor
\end{lstlisting}

\section{3D+Quat to 3D+YPR  }

\subsection{Transformation}

As mentioned in \S \ref{sect:ypr:gimbal}, the existence of degenerate cases in
the yaw-pitch-roll representation forces us to consider special cases in many formulas,
as it happens in this case when a quaternion must be converted into these angles.

Firstly, assuming a normalized quaternion, we define the \emph{discriminant} $\Delta$ as:

\begin{equation}
 \Delta = q_r q_y - q_x q_z
\end{equation}

Then, in most situations we will have $|\Delta|<1/2$, hence we can recover the
yaw ($\phi$),
pitch ($\chi$) and roll ($\psi$) angles as:

\begin{eqnarray}
\left\{
\begin{array}{rcl}
 \phi &=& \tan^{-1} \left( 2 \frac{q_r q_z + q_x q_y}{1-2(q_y^2+q_z^2)}  \right) \nonumber \\
 \chi &=& \sin^{-1} \left( 2 \Delta \right) \label{eq:quat2ypr_1} \\
 \psi &=& \tan^{-1} \left( 2 \frac{q_r q_x + q_y q_z}{1-2(q_x^2+q_y^2)}  \right) \nonumber
\end{array}
\right.
\end{eqnarray}

\noindent which can be obtained from trigonometric identities and
the transformation matrices associated to a quaternion and a triplet of angles yaw-pitch-roll
(see \S \ref{sect:ypr2mat}--\ref{sect:quat2mat}).
On the other hand, the special cases when $|\Delta| \approx 1/2$ can be solved as:

\begin{equation}
\begin{array}{c|c}
  \Delta = -1/2 & \Delta = 1/2 \\ \hline
  \begin{array}{rcl}
    \phi &=& 2 \tan^{-1} \frac{q_x}{q_r} \\
    \chi &=& -\pi /2 \\
    \psi &=& 0
  \end{array}
&
  \begin{array}{rcl}
    \phi &=& -2 \tan^{-1} \frac{q_x}{q_r} \\
    \chi &=& \pi /2 \\
    \psi &=& 0
  \end{array}
\end{array}
\label{eq:quat2ypr_2}
\end{equation}

\subsubsection{Implementation in MRPT}

Transforming a 6D pose from a quaternion to a yaw-pitch-roll representation is
achieved transparently via an implicit transform constructor:

\begin{lstlisting}
#include <mrpt/poses/CPose3D.h>
#include <mrpt/poses/CPose3DQuat.h>
using namespace mrpt::poses;

CPose3DQuat   p7;
...
CPose3D       p6 = p7;  // Transformation
\end{lstlisting}

\subsection{Uncertainty}

Given a Gaussian distribution over a 7D pose in quaternion form with
mean ${\mathbf{\bar{p}_7}}$ and being $cov(\mathbf{p_7})$ its $7 \times 7$ covariance matrix,
we can estimate the $6 \times 6$ covariance matrix of the equivalent yaw-pitch-roll-based
form by means of:

\begin{equation}
cov(\mathbf{p_6}) =
\frac{\partial \mathbf{p_6}(\mathbf{p_7}) }{\partial \mathbf{p_7}} ~
cov(\mathbf{p_7})  ~
\frac{\partial \mathbf{p_6}(\mathbf{p_7}) }{\partial \mathbf{p_7}}^\top
\end{equation}

\noindent where the Jacobian matrix has the following block structure:

\begin{equation}
\label{eq:jac_p6_p7}
\frac{\partial \mathbf{p_6}(\mathbf{p_7}) }{\partial \mathbf{p_7}} =
\left(
\begin{array}{c|c}
 \mathbf{I_3} & \mathbf{0_{3\times 4}} \\ \hline
 \mathbf{0_{3\times 3}} &  \frac{\partial (\phi,\chi,\psi)(q_r,q_x,q_y,q_z)}{\partial q_r,q_x,q_y,q_z}
\end{array}
\right)_{6 \times 7}
\end{equation}

In turn, the bottom-right sub-Jacobian matrix must account for two consecutive transformations:
normalization of the Jacobian (since each element has an uncertainty, but we need it normalized
for the transformation formulas to hold), then transformation to yaw-pitch-roll form. That is:

\begin{equation}
\frac{\partial (\phi,\chi,\psi)(q_r,q_x,q_y,q_z)}{\partial q_r,q_x,q_y,q_z} =
\frac{\partial (\phi,\chi,\psi)(q_r',q_x',q_y',q_z')}{\partial q_r',q_x',q_y',q_z'}
\frac{\partial (q_r',q_x',q_y',q_z')(q_r,q_x,q_y,q_z)}{\partial q_r,q_x,q_y,q_z}
\end{equation}

\noindent where the second term in the product is the Jacobian of the quaternion
normalization (see \S \ref{sect:quat:norm}). Here, and in the rest of this report,
it can be replaced by an identity Jacobian $\mathbf{I}_4$ if it is known for sure that
the quaternion is normalized.

Regarding the first term in the product, it is the
Jacobian of the functions in Eq. \ref{eq:quat2ypr_1}--\ref{eq:quat2ypr_2}, taking
into account that it can take three different forms for the cases $\chi=90^\circ$,
$\chi=-90^\circ$ and $|\chi| \neq 90^\circ$.

\subsubsection{Implementation in MRPT}

This conversion can be achieved by means of an explicit transform constructor, as shown below:

\begin{lstlisting}
#include <mrpt/poses/CPose3DQuat.h>
#include <mrpt/poses/CPose3DQuatPDFGaussian.h>
using namespace mrpt::poses;
using namespace mrpt::math;

CPose3DQuat            p7_mean = ...
CMatrixDouble77        p7_cov  = ...
CPose3DQuatPDFGaussian p7(p7_mean,p7_cov);
...
CPose3DPDFGaussian     p6 = CPose3DPDFGaussian(p7);  // Explicit constructor
\end{lstlisting}

\section{3D+YPR to matrix }
\label{sect:ypr2mat}

\subsection{Transformation}

The transformation matrix associated to a 6D pose given in yaw-pitch-roll form has this structure:

\begin{equation}
\mathbf{P}(x,y,z,\phi,\chi,\psi)=\left(
  \begin{array}{ccc|c}
   & & & x \\
   & \mathbf{R}(\phi,\chi,\psi) & & y \\
   & & & z \\ \hline
   0 & 0 & 0& 1
  \end{array}
\right)
\label{eq:matP_ypr}
\end{equation}

\noindent where the $3 \times 3$ rotation matrix $\mathbf{R}$ can be easily derived
from the fact that each of the three individual rotations (yaw, pitch and roll) operate
consecutively one after the other, i.e. over the already modified axis.
This can be achieved by right-side multiplication of the individual rotation matrices:

\begin{eqnarray}
\mathbf{R}_z(\phi) &=&
\left(
\begin{array}{ccc}
\cos \phi & -\sin \phi & 0 \\
\sin \phi & \cos \phi & 0 \\
0 & 0 & 1
\end{array}
\right) \quad \mathrm{\text{Yaw rotates around Z}} \\
\mathbf{R}_y(\chi) &= &
\left(
\begin{array}{ccc}
\cos \chi & 0 & \sin \chi \\
0 & 1 & 0 \\
-\sin \chi & 0 & \cos \chi
\end{array}
\right) \quad \mathrm{\text{Pitch rotates around Y}} \\
\mathbf{R}_x(\psi) &=&
\left(
\begin{array}{ccc}
1 & 0 & 0 \\
0 & \cos \psi & -\sin \psi \\
0 & \sin \psi &  \cos \psi
\end{array}
\right) \quad \mathrm{\text{Roll rotates around X}}
\end{eqnarray}

\noindent thus, concatenating them in the proper order ($\mathbf{R}_x$, then $\mathbf{R}_y$, then $\mathbf{R}_z$)
we obtain the complete rotation matrix:

\begin{eqnarray}
\mathbf{R}(\phi,\chi,\psi) &=&  \mathbf{R}_z(\phi) \mathbf{R}_y(\chi) \mathbf{R}_x(\psi)
\label{eq:mat_ypr} \\
&=&
\left(
\begin{array}{ccc}
\cos \phi \cos \chi  & \cos \phi \sin \chi \sin \psi - \sin \phi \cos \psi   & \cos \phi \sin \chi \cos \psi + \sin \phi \sin \psi \\
\sin \phi \cos \chi  & \sin \phi \sin \chi \sin \psi + \cos \phi \cos \psi  &  \sin \phi \sin \chi \cos \psi - \cos \phi \sin \psi \\
-\sin \chi & \cos \chi \sin \psi  &  \cos \chi \cos \psi
\end{array}
\right) \nonumber
\end{eqnarray}

A transformation matrix $\mathbf{P}$ is always well-defined and does not suffer of degenerate cases, but its large
storage requirements ($4\times 4=16$ elements) makes more advisable to use other representations such
as 3D+YPR (3+3=6 elements) or 3D+Quat (3+4=7 elements) in many situations.
An important exception is the case when computation time is critical and the most common operation
is composing (or inverse composing) a pose with a 3D point, where matrices require about half the
computation time than the other methods. On the other hand, composing a pose with another pose is
a slightly more efficient operation to carry out with a 3D+Quat representation.

In any case, when dealing with uncertainties, transformation matrices are not a
reasonable choice due to the
quadratic cost of keeping their covariance matrices.
The most common representation of a 6D pose with uncertainty in the literature
are 3D+Quat forms (e.g. see \cite{davison2007mrt}), thus
we will not describe how to
obtain covariance matrices of a transformation matrix here.
Note however that Jacobians of matrices are sometimes handy
as intermediaries (see \S\ref{chap:mat_deriv} and \S\ref{ch:se3_optim}).

\subsubsection{Implementation in MRPT}

The transformation matrix of any yaw-pitch-roll-based 6D pose stored in a
\texttt{CPose3D} class can be obtained as follows:

\begin{lstlisting}
#include <mrpt/poses/CPose3D.h>
using namespace mrpt::math;
using namespace mrpt::poses;

CPose3D          p;
CMatrixDouble44  M = p.getHomogeneousMatrixVal();
\end{lstlisting}

\section{3D+Quat to matrix }
\label{sect:quat2mat}

\subsection{Transformation}

The transformation matrix associated to a 6D pose given as a 3D translation plus
a quaternion is simply given by:

\begin{equation}
\mathbf{P}(x,y,z,q_r,q_x,q_y,q_z)=\left(
  \begin{array}{ccc|c}
   q_r^2+q_x^2-q_y^2-q_z^2 	&  2(q_x q_y - q_r q_z)	&  	2(q_z q_x+q_r q_y)  & x \\
   2(q_x q_y+q_r q_z) 		& q_r^2-q_x^2+q_y^2-q_z^2 	& 2(q_y q_z-q_r q_x) 	& y \\
   2(q_z q_x-q_r q_y) & 2(q_y q_z+q_r q_x)  & q_r^2- q_x^2 - q_y^2 + q_z^2 & z \\ \hline
   0 & 0 & 0& 1
  \end{array}
\right)
\end{equation}

\subsubsection{Implementation in MRPT}

In this case the interface of \texttt{CPose3DQuat} is exactly identical to that
of the yaw-pitch-roll form, that is:

\begin{lstlisting}
#include <mrpt/poses/CPose3DQuat.h>
using namespace mrpt::math;
using namespace mrpt::poses;

CPose3DQuat      p;
CMatrixDouble44  M = p.getHomogeneousMatrixVal();
\end{lstlisting}

\section{Matrix to 3D+YPR   }
\label{sect:mat2ypr}

\subsection{Transformation}

If we consider the $4 \times 4$ transformation matrix for
a 6D pose in 3D+YPR form (see Eq.~(\ref{eq:matP_ypr}) and (\ref{eq:mat_ypr})):

\begin{equation*}
\begin{array}{l}
\mathbf{P}(x,y,z,\phi,\chi,\psi)  \\
=\left(
  \begin{array}{ccc|c}
   \cos \phi \cos \chi  & \cos \phi \sin \chi \sin \psi - \sin \phi \cos \psi   & \cos \phi \sin \chi \cos \psi + \sin \phi \sin \psi & x \\
   \sin \phi \cos \chi  & \sin \phi \sin \chi \sin \psi + \cos \phi \cos \psi  &  \sin \phi \sin \chi \cos \psi - \cos \phi \sin \psi & y \\
   -\sin \chi & \cos \chi \sin \psi  &  \cos \chi \cos \psi & z \\ \hline
   0 & 0 & 0& 1
  \end{array}
\right) \\
=\left(
  \begin{array}{ccc|c}
   p_{11} & p_{12}& p_{13}& p_{14} \\
   p_{21} & p_{22}& p_{23}& p_{24} \\
   p_{31} & p_{32}& p_{33}& p_{34} \\
 \hline
   \cancel{0} & \cancel{0} & \cancel{0} & \cancel{1}
  \end{array}
\right)
\end{array}
\end{equation*}

\noindent where we seek a closed-form expression for the following function:

\begin{eqnarray*}
\mathbf{p_{6}}(\mathbf{p_{12}}): \mathcal{R}^{3 \times 4} &\rightarrow& \mathcal{R}^6
\\
\left(
\begin{array}{ccc|c}
	p_{11} & p_{12}& p_{13}& p_{14} \\
	p_{21} & p_{22}& p_{23}& p_{24} \\
	p_{31} & p_{32}& p_{33}& p_{34} \\
\end{array}
\right)
 &\rightarrow&
\begin{bmatrix}
	x \\ y \\ z \\ \phi \\ \chi \\ \psi
\end{bmatrix}
\begin{array}{c}
	\\ \\ \\ (yaw) \\ (pitch) \\ (roll)
\end{array}
\end{eqnarray*}

\noindent it is obvious that the 3D translation part can be recovered by simply:

\begin{eqnarray*}
 \left\{
  \begin{array}{rcl}
    x &=& p_{14} \\
    y &=& p_{24} \\
    z &=& p_{34}
  \end{array}
  \right.
\end{eqnarray*}

Regarding the three angles yaw ($\phi$), pitch ($\chi$) and roll ($\psi$), they must be obtained
in two steps in order to properly handle the special cases
(refer to the gimbal lock problem in \S \ref{sect:ypr:gimbal}).
Firstly, pitch is obtained from:

\begin{eqnarray}
pitch: \quad  \chi &=& \mathrm{atan2} \left( -p_{31} , \sqrt{ p_{11}^2 + p_{21}^2 } \right)
\end{eqnarray}

Next, depending on whether we are in a degenerate case ($|\chi|=90^\circ$) or
not ($|\chi| \neq 90^\circ$), the following expressions must
be applied\footnote{At this point, special thanks go to
Pablo Moreno Olalla for his work deriving robust expressions from Eq.~(\ref{eq:mat_ypr})
that work for all the special cases.}:

\begin{eqnarray}
  \chi = -90^\circ \quad \longrightarrow \quad
   \left\{
  \begin{array}{rcl}
    yaw: \quad  \phi  &=& \mathrm{atan2}( -p_{23}, -p_{13} ) \\
    roll: \quad  \psi  &=& 0 \\
  \end{array}
   \right.
\\
  |\chi| \neq 90^\circ \quad \longrightarrow \quad
   \left\{
  \begin{array}{rcl}
    yaw: \quad  \phi   &=& \mathrm{atan2}(p_{21},p_{11}) \\
     roll: \quad  \psi  &=&  \mathrm{atan2}(p_{32},p_{33}) \\
  \end{array}
   \right.
\\
  \chi = 90^\circ \quad \longrightarrow \quad
   \left\{
  \begin{array}{rcl}
    yaw: \quad  \phi  &=& \mathrm{atan2}( p_{23}, p_{13} ) \\
     roll: \quad  \psi  &=& 0 \\
  \end{array}
   \right.
\end{eqnarray}

\subsubsection{Implementation in MRPT}

Given a matrix \texttt{M}, the \texttt{CPose3D} representation can be obtained via
an explicit transform constructor:

\begin{lstlisting}
#include <mrpt/poses/CPose3D.h>
using namespace mrpt::math;
using namespace mrpt::poses;

CMatrixDouble44  M;
...
CPose3D          p = CPose3D(M);
\end{lstlisting}

\subsection{Uncertainty}
\label{sect:jacob.p6_p12}

Let a Gaussian distribution over a SE(3) pose in matrix form be specified 
by a $\mathbf{\bar{p}_{12}}$ mean and let $cov(\mathbf{p_{12}})$ be its $12 \times 12$ covariance matrix (refer to \S\ref{sect:mat_deriv:not} for an explanation of where ``12'' comes from). We can estimate the $6 \times 6$  covariance matrix $cov(\mathbf{p_6})$ of the equivalent yaw-pitch-roll form by means of:

\begin{equation}
cov(\mathbf{p_{6}}) =
\frac{\partial \mathbf{p_{6}}(\mathbf{p_{12}}) }{\partial \mathbf{p_{12}}} ~
cov(\mathbf{p_{12}})  ~
\frac{\partial \mathbf{p_{6}}(\mathbf{p_{12}}) }{\partial \mathbf{p_{12}}}^\top
\end{equation}

\noindent where the Jacobian matrix has the following block structure:

\begin{equation}
\label{eq:jac_p6_p12}
\frac{\partial \mathbf{p_{6}}(\mathbf{p_{12}}) }{\partial \mathbf{p_{12}}} =
\left(
\begin{array}{cc}
\mathbf{0}_{3 \times 9} & \mathbf{I}_3 \\
\dfrac{\partial \{ \phi,\chi,\psi \} }{\partial vec(\mathbf{R})} & \mathbf{0}_{3 \times 3}
\end{array}
\right)_{6 \times 12}
\end{equation}

\noindent where $\mathbf{R}$ is the $3 \times 3$ SO(3) rotational part of the pose $\mathbf{\bar{p}_{12}}$ and the $vec(\cdot)$ operator (column major) is defined in \S\ref{sect:mat_deriv:ops}.

The remaining Jacobian block is defined as:

\begin{subeqnarray}
\frac{\partial \{ \phi,\chi,\psi \} }{\partial vec(\mathbf{R})} &=&
\left(\begin{array}{ccccccccc} 
	J_{11} & 0 & 0 & J_{14} & 0 & 0 & 0 & 0 & 0\\ 
	J_{21} & 0 & 0 & J_{24} & 0 & 0 & 0 & J_{28} & 0\\ 
	0 & 0 & 0 & 0 & 0 & 0 & 0 & J_{38} & J_{39}
\end{array}\right)_{3 \times 12}
\\
J_{11} &=& -\frac{p_{21}}{k} \\
J_{14} &=& \frac{p_{11}}{k} \\
J_{21} &=& \frac{p_{11}\,p_{32}}{\sqrt{k}\,\left(k+{p_{32}}^2\right)} \\
J_{24} &=& \frac{p_{21}\,p_{32}}{\sqrt{k}\,\left(k+{p_{32}}^2\right)} \\
J_{28} &=&  -\frac{\sqrt{k}}{k+{p_{32}}^2} \\
J_{38} &=& \frac{p_{33}}{{p_{32}}^2+{p_{33}}^2} \\
J_{39} &=& -\frac{p_{32}}{{p_{32}}^2+{p_{33}}^2} \\
k &=& {p_{11}}^2 + {p_{21}}^2
\end{subeqnarray}

\section{Matrix to 3D+Quat }

\subsection{Transformation}

A numerically stable method to convert a $3 \times 3$ rotation matrix into a quaternion is
described in \cite{bar2000new}, which includes creating a temporary $4 \times 4$ matrix and
computing the eigenvector corresponding to its largest eigenvalue.
However, an alternative, more efficient method which can be applied if we are sure about
the matrix being orthonormal is to simply convert it firstly to a yaw-pitch-roll representation
(see \S \ref{sect:mat2ypr}) and then convert it
to a quaternion representation (see \S \ref{sect:ypr2quat}).

\subsubsection{Implementation in MRPT}

Given a matrix \texttt{M}, the \texttt{CPose3DQuat} representation can be obtained via
an explicit transform constructor:

\begin{lstlisting}
#include <mrpt/poses/CPose3DQuat.h>
using namespace mrpt::math;
using namespace mrpt::poses;

CMatrixDouble44  M;
...
CPose3DQuat      p = CPose3DQuat(M);
\end{lstlisting}

\chapter{Composing a pose and a point}
\label{ch:comp_pose_pt}

This chapter reviews how to compute the global coordinates of a point $\mathbf{a}$
given a pose $\mathbf{p}$ and the point coordinates relative to that coordinate system $\mathbf{a'}$,
as illustrated in Figure~\ref{fig:1}, that is, the pose chaining $\mathbf{a} = \mathbf{p} \oplus \mathbf{a'}$.

\section{With poses in 3D+YPR form}

\subsection{Composition}

In this case the solution is to firstly compute the $4 \times 4$
transformation matrix of the pose using Eq.~\ref{eq:mat_ypr}, then proceed
as described in \S \ref{sect:comp_point:mat}.

\subsubsection{Implementation in MRPT}

A pose-point composition can be evaluated by means of:

\begin{lstlisting}
#include <mrpt/poses/CPose3D.h>
#include <mrpt/math/lightweight_geom_data.h>
using namespace mrpt::poses;
using namespace mrpt::math;

CPose3D   q;
TPoint3D  in_p, out_p;
...
q.composePoint(in_p, out_p);
\end{lstlisting}

\subsection{Uncertainty}

Given a Gaussian distribution over a 6D pose in 3D+YPR form with
mean ${\mathbf{\bar{p}_6}= (\bar x ~\bar y ~\bar z ~ \bar\phi ~ \bar\chi ~ \bar\psi)^\top}$ and being $cov(\mathbf{p_6})$ its $6 \times 6$ covariance matrix,
and being ${\mathbf{\bar{a'}} = (\bar a'_x ~\bar a'_y ~\bar a'_z )^\top }$ and $cov(\mathbf{a'})$ the mean and covariance of the 3D point
$\mathbf{a'}$, respectively, and assuming that both distributions are independent,
then the approximated covariance of the transformed point
$\mathbf{a} = \mathbf{f_{pr}} (\mathbf{p_6},\mathbf{a}) = \mathbf{p_6} \oplus \mathbf{a'}$ is given by:

\begin{equation}
cov(\mathbf{a}) =
\frac{\partial \mathbf{f_{pr}} (\mathbf{p_6},\mathbf{a})}{\partial \mathbf{p_6}} ~
cov(\mathbf{p_6})  ~
\frac{\partial \mathbf{f_{pr}} (\mathbf{p_6},\mathbf{a})}{\partial \mathbf{p_6}}^\top
+
\frac{\partial \mathbf{f_{pr}} (\mathbf{p_6},\mathbf{a})}{\partial \mathbf{a}} ~
cov(\mathbf{a'})  ~
\frac{\partial \mathbf{f_{pr}} (\mathbf{p_6},\mathbf{a})}{\partial \mathbf{a}}^\top
\end{equation}

The Jacobian matrices are:

\begin{eqnarray}
\left. \frac{\partial \mathbf{f_{pr}} (\mathbf{p_6},\mathbf{a})}{\partial \mathbf{p_6}} ~
\right|_{3 \times 6}
&=&
\left(
\begin{array}{c|ccc}
~            & j_{14} & j_{15} & j_{16} \\
\mathbf{I_3} & j_{24} & j_{25} & j_{26} \\
~            & j_{34} & j_{35} & j_{36}
\end{array}
\right)
\\
\left.\frac{\partial \mathbf{f_{pr}} (\mathbf{p_6},\mathbf{a})}{\partial \mathbf{a}}
\right|_{3 \times 3}
&=&
\mathbf{R}(\bar\phi,\bar\chi,\bar\psi) \quad\quad \text{See Eq.(\ref{eq:mat_ypr})}
\end{eqnarray}

\noindent with these entry values:

\begin{eqnarray*}
j_{14} &=&
-\bar a'_x \sin \bar\phi \cos \bar\chi+\bar a'_y (-\sin \bar\phi \sin \bar\chi \sin \bar\psi-\cos \bar\phi \cos \bar\psi)+\bar a'_z (-\sin \bar\phi \sin \bar\chi \cos
\bar\psi+\cos
\bar\phi \sin \bar\psi)
\\
j_{15} &=&
-\bar a'_x \cos \bar\phi \sin \bar\chi+\bar a'_y (\cos \bar\phi \cos \bar\chi \sin \bar\psi       )+\bar a'_z (\cos \bar\phi \cos \bar\chi \cos \bar\psi      )
\\
j_{16} &=&
\bar a'_y (\cos \bar\phi \sin \bar\chi \cos \bar\psi+\sin \bar\phi \sin \bar\psi)+\bar a'_z (-\cos \bar\phi \sin \bar\chi \sin \bar\psi+\sin \bar\phi \cos \bar\psi)
\\
j_{24} &=&
\bar a'_x \cos \bar\phi \cos \bar\chi+\bar a'_y (\cos \bar\phi \sin \bar\chi \sin \bar\psi-\sin \bar\phi \cos \bar\psi)+\bar a'_z (\cos \bar\phi \sin \bar\chi \cos
\bar\psi+\sin
\bar\phi \sin \bar\psi)
\\
j_{25} &=&
-\bar a'_x \sin \bar\phi \sin \bar\chi+\bar a'_y (\sin \bar\phi \cos \bar\chi \sin \bar\psi)      +\bar a'_z (\sin \bar\phi \cos \bar\chi \cos \bar\psi      )
\\
j_{26} &=&
\bar a'_y (\sin \bar\phi \sin \bar\chi \cos \bar\psi-\cos \bar\phi \sin \bar\psi)+\bar a'_z (-\sin \bar\phi \sin \bar\chi \sin \bar\psi-\cos \bar\phi \cos \bar\psi)
\\
j_{34} &=&
0
\\
j_{35} &=&
    -\bar a'_x \cos \bar\chi-\bar a'_y \sin \bar\chi \sin \bar\psi-\bar a'_z \sin \bar\chi \cos \bar\psi
\\
j_{36} &=&
\bar a'_y \cos \bar\chi \cos \bar\psi-\bar a'_z \cos \bar\chi \sin \bar\psi
\end{eqnarray*}

An approximate version of the Jacobian w.r.t. the pose has been proposed in
\cite{sibley2009rba} for the case of very small rotations.
It can be derived from the expression for
$\frac{\partial \mathbf{f_{pr}} (\mathbf{p_6},\mathbf{a})}{\partial \mathbf{p_6}}$
above by replacing all $\sin \alpha \approx 0$ and $\cos \alpha \approx 1$,
leading to:

\begin{eqnarray}
\left. \frac{\partial \mathbf{f_{pr}} (\mathbf{p_6},\mathbf{a})}{\partial \mathbf{p_6}} ~
\right|_{3 \times 6}
&\approx&
\left(
\begin{array}{c|ccc}
~            & -\bar a'_y & \bar a'_z & 0 \\
\mathbf{I_3} & \bar a'_x & 0 & -\bar a'_z \\
~            & 0 & -\bar a'_x & \bar a'_y \\
\end{array}
\right)
\quad\quad \text{(For small rotations only!!)}
\end{eqnarray}

\subsubsection{Implementation in MRPT}

There is not a direct method to implement a pose-point composition with uncertainty,
but the two required Jacobians can be obtained from the method \texttt{composePoint()}:

\begin{lstlisting}
#include <mrpt/poses/CPose3D.h>
using namespace mrpt::poses;
using namespace mrpt::math;

CPose3D  q;
CMatrixFixedNumeric<double,3,3>  df_dpoint;
CMatrixFixedNumeric<double,3,6>  df_dpose;
q.composePoint(lx,ly,lz,gx,gy,gz, &df_dpoint, &df_dpose);
\end{lstlisting}

\section{With poses in 3D+Quat form}

\subsection{Composition}

Given a pose described as $\mathbf{p_7} = [x ~ y ~ z ~  q_r ~ q_x ~ q_y ~ q_z ] ^ \top$,
we are interested in the coordinates of $\mathbf{a}=[a_x ~ a_y ~ a_z]^\top$ such as
$\mathbf{a} = \mathbf{p_7} \oplus \mathbf{a'}$ for some known input point
$\mathbf{a'} = [a'_x ~ a'_y ~ a'_z]^\top$.
The solution is given by:

\begin{equation}
\mathbf{a} = \mathbf{f_{qr}} (\mathbf{p},\mathbf{a'})
\end{equation}

\noindent where the function $\mathbf{f_{qr}}(\cdot)$ is defined as:

\begin{equation}
\mathbf{f_{qr}} (\mathbf{p},\mathbf{a'}) =
 \left(
\begin{array}{c}
 x + a'_x + 2 \left[-(q_y^2+ q_z^2) a'_x +(q_x q_y - q_r q_z) a'_y+(q_r q_y+q_x q_z) a'_z \right]  \\
 y + a'_y + 2 \left[(q_r q_z+  q_x q_y) a'_x-(q_x^2 +q_z^2) a'_y+(q_y q_z-q_r q_x) a'_z \right] \\
 z + a'_z + 2 \left[(q_x q_z-  q_r q_y) a'_x+(q_r q_x + q_y q_z) a'_y-(q_x^2+q_y^2) a'_z \right]  \\
\end{array}
\right)
\label{eq:quat_rot_point_func}
\end{equation}

\subsubsection{Implementation in MRPT}

A pose-point composition can be evaluated by means of:

\begin{lstlisting}
#include <mrpt/poses/CPose3D.h>
#include <mrpt/math/lightweight_geom_data.h>
using namespace mrpt::poses;
using namespace mrpt::math;

CPose3DQuat  q;
TPoint3D     in_p, out_p;
...
q.composePoint(in_p, out_p);
\end{lstlisting}

\subsection{Uncertainty}

Given a Gaussian distribution over a 7D pose in quaternion form with
mean ${\mathbf{\bar{p}_7}}$ and being $cov(\mathbf{p_7})$ its $7 \times 7$ covariance matrix,
and being ${\mathbf{\bar{a'}}}$ and $cov(\mathbf{a'})$ the mean and covariance of the 3D point
$\mathbf{a'}$, respectively, the approximated covariance of the transformed point
$\mathbf{a} = \mathbf{p_7} \oplus \mathbf{a'}$ is given by:

\begin{equation}
cov(\mathbf{a}) =
\frac{\partial \mathbf{f_{qr}} (\mathbf{p},\mathbf{a})}{\partial \mathbf{p}} ~
cov(\mathbf{p_7})  ~
\frac{\partial \mathbf{f_{qr}} (\mathbf{p},\mathbf{a})}{\partial \mathbf{p}}^\top
+
\frac{\partial \mathbf{f_{qr}} (\mathbf{p},\mathbf{a})}{\partial \mathbf{a}} ~
cov(\mathbf{a'})  ~
\frac{\partial \mathbf{f_{qr}} (\mathbf{p},\mathbf{a})}{\partial \mathbf{a}}^\top
\end{equation}

The Jacobian matrices are:

\small{
\begin{equation}
 \begin{array}{l}
\left.
\frac{\partial \mathbf{f_{qr}} (\mathbf{p},\mathbf{a})}{\partial \mathbf{p}}
\right|_{3 \times 7} =
  \left(
  \begin{array}{cccc}
    1  & 0 & 0 & ~ \\
    0  & 1 & 0 & \frac{\partial \mathbf{f_{qr}} (\mathbf{p},\mathbf{a})}{\partial [qr ~ qx ~ qy ~ qz]}  \\
    0  & 0 & 1 & ~ \\
  \end{array}
  \right)
 \end{array}
\label{eq:quat_rot_point_func_jacob1}
\end{equation}

\noindent with the auxiliary term $\frac{\partial \mathbf{f_{qr}} (\mathbf{p},\mathbf{a})}{\partial [qr ~ qx ~ qy ~ qz]} $
including the normalization Jacobian (see \S \ref{sect:quat:norm}):

\begin{eqnarray}
\frac{\partial \mathbf{f_{qr}} (\mathbf{p},\mathbf{a})}{\partial [qr ~ qx ~ qy ~ qz]}
  &=&
  2
  \left(
  \begin{array}{ccccccc}
    -q_z a_y +q_y a_z   & q_y a_y + q_z a_z   &   -2q_y a_x + q_x a_y +q_r a_z & -2q_z a_x - q_r a_y +q_x a_z \\
    q_z a_x - q_x a_z   & q_y a_x - 2q_x a_y -q_r a_z  & q_x a_x +q_z a_z   & q_r a_x - 2 q_z a_y +q_y a_z \\
    -q_y a_x + q_x a_y  & q_z a_x + q_r a_y - 2q_x a_z  & -q_r a_x + q_z a_y - 2 q_y a_z  & q_x a_x + q_y a_y \\
  \end{array}
  \right)
   \nonumber \\
  & \times  &
\frac{\partial (q_r',q_x',q_y',q_z')(q_r,q_x,q_y,q_z)}{\partial q_r,q_x,q_y,q_z}
\end{eqnarray}

The other Jacobian is given by:

\begin{equation}
 \begin{array}{l}
\left.
\frac{\partial \mathbf{f_{qr}} (\mathbf{p},\mathbf{a})}{\partial \mathbf{a}}
\right|_{3 \times 3} =
  2
  \left(
  \begin{array}{ccc}
    \frac{1}{2}-q_y^2-q_z^2   & q_x q_y - q_r q_z   & q_r q_y + q_x q_z \\
    q_r q_z + q_x q_y  & \frac{1}{2} - q_x^2-q_z^2  & q_y q_z - q_r q_x \\
    q_x q_z - q_r q_y  & q_r q_x + q_y q_z & \frac{1}{2} - q_x^2-q_y^2
  \end{array}
  \right)
\label{eq:quat_rot_point_func_jacob2}
 \end{array}
\end{equation}
}

\subsubsection{Implementation in MRPT}

There is not a direct method to implement a pose-point composition with uncertainty,
but the two required Jacobians can be obtained from the method \texttt{composePoint()}:

\begin{lstlisting}
#include <mrpt/poses/CPose3DQuat.h>
#include <mrpt/math/CMatrixFixedNumeric.h>
using namespace mrpt::poses;
using namespace mrpt::math;

CPose3DQuat  q;
CMatrixFixedNumeric<double,3,3>  df_dpoint;
CMatrixFixedNumeric<double,3,7>  df_dpose;
q.composePoint(lx,ly,lz,gx,gy,gz, &df_dpoint, &df_dpose);
\end{lstlisting}

\section{With poses in matrix form}
\label{sect:comp_point:mat}

Given a $4\times 4$ transformation matrix $\mathbf{M}$ corresponding to a 6D pose
$\mathbf{p}$ and a point in local coordinates
$\mathbf{a'} = [a'_x ~ a'_y ~ a'_z]$, the corresponding point in global coordinates
$\mathbf{a} = [a_x ~ a_y ~ a_z]$ can be computed easily as:

\begin{eqnarray}
\mathbf{a} &=& \mathbf{p} \oplus \mathbf{a'} \nonumber \\
\left(\begin{array}{c}
 a_x \\ a_y \\ a_z \\ 1
\end{array}\right)
&=&
\mathbf{M}
\left(\begin{array}{c}
 a'_x \\ a'_y \\ a'_z \\ 1
\end{array}\right)
\end{eqnarray}

\noindent where homogeneous coordinates (the column matrices) have been used for the 3D points
-- see also Eq.~(\ref{eq:x2_T_x1}.

\chapter{Points relative to a pose}
\label{ch:inv_pose_point}

In the next sections we will review how to compute the relative coordinates
of a point $\mathbf{a'}$ given a pose $\mathbf{p}$ and the point global coordinates
$\mathbf{a}$,
as illustrated in Figure~\ref{fig:1}, that is, $\mathbf{a'} = \mathbf{a} \ominus \mathbf{p}$.

\section{With poses in 3D+YPR form}

\subsection{Inverse transformation}

The relative coordinates of a point with respect to a pose in this parameterization can be computed
by first obtaining the matrix form of the pose \S\ref{sect:ypr2mat}, then using it as described in \S\ref{sect:inv_comp_mat}.

\subsubsection{Implementation in MRPT}

Given a 6D-pose as an object of type \texttt{CPose3D}, one can invoke its method
\texttt{inverseComposePoint()} which, in one of its signatures, reads:

\begin{lstlisting}
#include <mrpt/poses/CPose3D.h>
#include <mrpt/math/lightweight_geom_data.h>
using namespace mrpt::poses;
using namespace mrpt::math;

CPose3D    q;
TPoint3D   in_p, out_p;
...
q.inverseComposePoint(in_p, out_p);
\end{lstlisting}

\subsection{Uncertainty}

In this case it's preferred to transform the 3D pose to a 3D+Quat, then
perform the transformation as described in the following section.

\section{With poses in 3D+Quat form}
\label{sect:point_inv:quat}

\subsection{Inverse transformation}

Given a 7D-pose $\mathbf{p_7}=[x ~ y ~ z ~ qr ~ qx ~ qy ~ qz]^\top$ and a point in
global coordinates $\mathbf{a} = [a_x ~ a_y ~ a_z]^\top$, the point coordinates
relative to $\mathbf{p_7}$, that is, $\mathbf{a'} = \mathbf{a} \ominus \mathbf{p_7}$, are given by:

\begin{eqnarray}
\label{eq:fqri}
\mathbf{a'} = \mathbf{f_{qri}}( \mathbf{a}, \mathbf{p_7} ) =
\left(\begin{array}{c}
 (a_x-x) + 2 \left[-(q_y^2+ q_z^2) (a_x-x) +(q_x q_y + q_r q_z) (a_y-y)+(-q_r q_y+q_x q_z) (a_z-z) \right]  \\
 (a_y-y) + 2 \left[(-q_r q_z+  q_x q_y) (a_x-x)-(q_x^2 +q_z^2) (a_y-y)+(q_y q_z+q_r q_x) (a_z-z) \right] \\
 (a_z-z) + 2 \left[(q_x q_z+ q_r q_y) (a_x-x)+(-q_r q_x + q_y q_z) (a_y-y)-(q_x^2+q_y^2) (a_z-z) \right]  \\
\end{array}\right)
\end{eqnarray}

\subsubsection{Implementation in MRPT}

Given a 7D-pose as an object of type \texttt{CPose3DQuat}, one can invoke its method
\texttt{inverseComposePoint()} which, in one of its signatures, reads:

\begin{lstlisting}
#include <mrpt/poses/CPose3DQuat.h>
#include <mrpt/math/lightweight_geom_data.h>
using namespace mrpt::poses;
using namespace mrpt::math;

CPose3DQuat  q;
TPoint3D     in_p, out_p;
...
q.inverseComposePoint(in_p, out_p);
\end{lstlisting}

\subsection{Uncertainty}

Given a Gaussian distribution over a 7D pose in 3D+Quar form with
mean $\mathbf{\bar{p}_7}$ and being $cov(\mathbf{p_7})$ its $7 \times 7$ covariance matrix,
and assuming that a 3D point follows an (independent) Gaussian distribution
with mean $\mathbf{\bar{a}}$ and $3 \times 3$ covariance $cov(\mathbf{a})$,
we can estimate the covariance of the transformed local point $\mathbf{a'}$
as:

\begin{equation}
cov(\mathbf{a'}) =
\frac{\partial \mathbf{f_{qri}}( \mathbf{a}, \mathbf{p} )}{\partial \mathbf{p_7}}
cov(\mathbf{p_7})  ~
\frac{\partial \mathbf{f_{qri}}( \mathbf{a}, \mathbf{p} )}{\partial \mathbf{p_7}}^\top
+
\frac{\partial \mathbf{f_{qri}}( \mathbf{a}, \mathbf{p} )}{\partial \mathbf{a}}
cov(\mathbf{a})  ~
\frac{\partial \mathbf{f_{qri}}( \mathbf{a}, \mathbf{p} )}{\partial \mathbf{a}}^\top
\end{equation}

\noindent where the Jacobian matrices are given by:

\begin{eqnarray}
\frac{ \partial \mathbf{f_{qri}}( \mathbf{a}, \mathbf{p} )}{\partial \mathbf{a}} =
\left(\begin{array}{ccc}
1 - 2 (q_y^2 + q_z^2)   &    2 q_x q_y + 2 q_r q_z  &     - 2 q_r q_y + 2 q_x q_z \\
-2 q_r q_z + 2 q_x q_y &   1 - 2 (q_x^2 + q_z^2) &   2 q_y q_z + 2 q_r q_x \\
2 q_x q_z + 2 q_r q_y &   -2 q_r q_x + 2 q_y q_z & 1 - 2 (q_x^2 + q_y^2 ) \\
\end{array}\right)_{3 \times 3}
\end{eqnarray}

\noindent and, if we define
$\Delta x = (a_x - x)$, $\Delta y = (a_y - y)$ and $\Delta z = (a_z - z)$, we can
write the Jacobian with respect to the pose as:

\begin{eqnarray}
\label{eq:df_qri_p}
\frac{ \partial \mathbf{f_{qri}}( \mathbf{a}, \mathbf{p} )}{\partial \mathbf{p}} =
\left(\begin{array}{ccc|c}
2 q_y^2 + 2 q_z^2 - 1 &   -2 q_r q_z - 2 q_x q_y &  2 q_r q_y - 2 q_x q_z &  ~
\\
2 q_r q_z - 2 q_x q_y &  2 q_x^2 + 2 q_z^2 - 1   &  -2 q_r q_x - 2 q_y q_z &
  \frac{ \partial \mathbf{f_{qrir}}( \mathbf{a}, \mathbf{p} )}{\partial \mathbf{p}}
\\
-2 q_r q_y - 2 q_x q_z &  2 q_r q_x - 2 q_y q_z &  2 q_x^2 + 2 q_y^2 - 1 & ~
\\
\end{array}\right)_{3 \times 7}
\end{eqnarray}

\noindent with:

\begin{eqnarray}
\label{eq:jacob.f_qrir_p}
\frac{ \partial \mathbf{f_{qrir}}( \mathbf{a}, \mathbf{p} )}{\partial \mathbf{p}}
&=&
2
\left(\begin{array}{cccc}
-q_y \Delta z + q_z \Delta y &                  q_y \Delta y + q_z \Delta z &
q_x \Delta y - 2 q_y \Delta x - q_r \Delta z &  q_x \Delta z + q_r \Delta y - 2q_z \Delta x
\\
q_x \Delta z - q_z \Delta x &  q_y \Delta x - 2 q_x \Delta y + q_r \Delta z &
q_x \Delta x + q_z \Delta z &  -q_r \Delta x - 2 q_z \Delta y + q_y \Delta z
\\
q_y \Delta x - q_x \Delta y &  q_z \Delta x - q_r \Delta y - 2 q_x \Delta z &
q_z \Delta y + q_r \Delta x - 2 q_y \Delta z &  q_x \Delta x + q_y \Delta y
\\
\end{array}\right) \nonumber \\
& ~ & \cdot \frac{\partial (q_r',q_x',q_y',q_z')(q_r,q_x,q_y,q_z)}{\partial q_r,q_x,q_y,q_z}
\end{eqnarray}

\noindent where the second term in the product is the Jacobian of the quaternion
normalization (see \S \ref{sect:quat:norm}).

\subsubsection{Implementation in MRPT}

As in the previous case, here we it can be also employed the method
\texttt{inverseComposePoint()} which if provided the optional output parameters,
will return the desired Jacobians:

\begin{lstlisting}
#include <mrpt/poses/CPose3DQuat.h>
#include <mrpt/math/lightweight_geom_data.h>
using namespace mrpt::poses;
using namespace mrpt::math;

CPose3DQuat  q;
TPoint3D     g, l;
CMatrixFixedNumeric<double,3,3>  dfi_dpoint;
CMatrixFixedNumeric<double,3,7>  dfi_dpose;
...
q.inverseComposePoint(
  g.x,g.y,g.z, // Input  (global coords)
  l.x,l.y,l.z, // Output (local coords)
  &dfi_dpoint, // 3x3 Jacobian
  &dfi_dpose   // 3x7 Jacobian
  );
\end{lstlisting}

\section{With poses as matrices}
\label{sect:inv_comp_mat}

Given a $4\times 4$ transformation matrix $\mathbf{M}$ corresponding to a 6D pose
$\mathbf{p}$ and a point in global coordinates
$\mathbf{a} = [a_x ~ a_y ~ a_z]$, the corresponding point in local coordinates
$\mathbf{a'} = [a'_x ~ a'_y ~ a'_z]$ is given by:

\begin{eqnarray}
\mathbf{a'} &=& \mathbf{a} \ominus \mathbf{p} \nonumber \\
\left(\begin{array}{c}
 a'_x \\ a'_y \\ a'_z \\ 1
\end{array}\right)
&=&
\mathbf{M}^{-1}
\left(\begin{array}{c}
 a_x \\ a_y \\ a_z \\ 1
\end{array}\right)
\end{eqnarray}

\noindent where homogeneous coordinates (the column matrices) have been used for the 3D points.
An efficient way to compute the inverse of a homogeneous matrix is
described in \S \ref{sect:inverse:mat}

\section{Relation with pose-point direct composition}

There is an interesting result that naturally arises from the matrix form
explained in the previous section.
By definition, we have:

\begin{eqnarray}
 \mathbf{a} = \mathbf{p} \oplus \mathbf{a'}
\leftrightarrow
 \mathbf{a'} = \mathbf{a} \ominus \mathbf{p}
\label{eq:comp_invcomp}
\end{eqnarray}

Then, starting with $\mathbf{a} = \mathbf{p} \oplus \mathbf{a'}$
and using the matrix form, we can proceed as follows:

\begin{eqnarray*}
 \mathbf{a} &=& \mathbf{p} \oplus \mathbf{a'}    \\
 \mathbf{A} &=& \mathbf{P}  \mathbf{A'}  \quad \text{(Representation as matrices)}  \\
 \mathbf{P}^{-1} \mathbf{A} &=& \mathbf{P}^{-1} \mathbf{P}  \mathbf{A'} \\
 \mathbf{P}^{-1} \mathbf{A} &=& \mathbf{A'}  \\
 (\ominus \mathbf{p}) \oplus \mathbf{a} &=& \mathbf{a'}  \quad \text{(Back to $\oplus$/$\ominus$ notation)} \\
 (\ominus \mathbf{p}) \oplus \mathbf{a} &=& \mathbf{a} \ominus \mathbf{p}  \quad \text{(Using Eq.~\ref{eq:comp_invcomp})}
\end{eqnarray*}

\noindent where $(\ominus \mathbf{p})$ stands for the inverse of a pose $\mathbf{p}$.
Thus, the result is that any inverse pose composition can be transformed into a normal
pose composition, by switching the order of the two arguments
($\mathbf{a}$ and $\mathbf{p}$ in this case) and inverting the latter.
Note that the inverse of a pose is a topic discussed in \S \ref{sect:inverse}.

\chapter{Composition of two poses}
\label{ch:pose_pose_comp}

Next sections are devoted to computing the composed pose $\mathbf{p}$ resulting
from a concatenation of two 6D poses $\mathbf{p_1}$ and $\mathbf{p_2}$,
that is, $\mathbf{p} = \mathbf{p_1} \oplus \mathbf{p_2}$.
An example of this operation was shown in Figure~\ref{fig:comp_2poses}.

\section{With poses in 3D+YPR form}

\subsection{Pose composition}

There is not simple
equation for pose composition for poses described as triplets
of yaw-pitch-roll angles, thus it is recommended to transform them into
either 3D+Quad or matrix form
(see, \S \ref{sect:ypr2quat} and \S \ref{sect:ypr2mat}, respectively),
then compose them as described in the following sections and finally
convert the result back into 3D+YPR form.

\subsubsection{Implementation in MRPT}

Pose composition for 3D+YPR poses is implemented via overloading
of the ``\texttt{+}'' C++ operator
(using matrix representation to perform the intermediary computations),
such as composing can be simply
writen down as:

\begin{lstlisting}
#include <mrpt/poses/CPose3D.h>
using namespace mrpt::poses;

CPose3D  p1,p2;
...
CPose3D  p = p1 + p2;    // Pose composition
\end{lstlisting}

\subsection{Uncertainty}

Let $\mathcal{N}(\mathbf{\bar{p}_6^1}, cov(\mathbf{p_6^1}))$ and
$\mathcal{N}(\mathbf{\bar{p}_6^2}, cov(\mathbf{p_6^2}))$ represent
two independent Gaussian distributions over a pair of 6D
poses in 3D+YPR form.
Note that superscript indexes have been employed
for notation convenience (they do \emph{not} denote exponentiation!).

Then, the probability distribution of their composition
$\mathbf{p_6^R} = \mathbf{p_6^1} \oplus \mathbf{p_6^2}$
can be approximated via linear error propagation by considering
a mean value of:

\begin{eqnarray}
\mathbf{\bar{p}_6^R}
= \mathbf{f_{pc}}(\mathbf{\bar{p}_6^1}, \mathbf{\bar{p}_6^2})
= \mathbf{\bar{p}_6^1} \oplus \mathbf{\bar{p}_6^2}
\end{eqnarray}

\noindent and a covariance matrix given by:

\begin{eqnarray}
cov(\mathbf{p_6^R}) &=&
\left. \frac{\partial \mathbf{f_{pc}}(\mathbf{p},\mathbf{q}) }{\partial \mathbf{p}} \right|_{\overset{ \mathbf{p}=\mathbf{p_6^1} }{ \mathbf{q}=\mathbf{p_6^2} }}
cov(\mathbf{p_6^1})
\left. \frac{\partial \mathbf{f_{pc}}(\mathbf{p},\mathbf{q}) }{\partial \mathbf{p}} \right|_{\overset{ \mathbf{p}=\mathbf{p_6^1} }{ \mathbf{q}=\mathbf{p_6^2} }}^\top
\nonumber \\ &+&
\left. \frac{\partial \mathbf{f_{pc}}(\mathbf{p},\mathbf{q}) }{\partial \mathbf{q}} \right|_{\overset{ \mathbf{p}=\mathbf{p_6^1} }{ \mathbf{q}=\mathbf{p_6^2} }}
cov(\mathbf{p_6^2})
\left. \frac{\partial \mathbf{f_{pc}}(\mathbf{p},\mathbf{q}) }{\partial \mathbf{q}} \right|_{\overset{ \mathbf{p}=\mathbf{p_6^1} }{ \mathbf{q}=\mathbf{p_6^2} }}^\top
\end{eqnarray}

The problematic part is obtaining a closed form expression for the
Jacobians
$\frac{\partial \mathbf{f_{pc}}(\mathbf{p},\mathbf{q}) }{\partial \mathbf{p}}$
and
$\frac{\partial \mathbf{f_{pc}}(\mathbf{p},\mathbf{q}) }{\partial \mathbf{q}}$
since, as mentioned in the previous section, there is not a simple expression
for the function $\mathbf{f_{pc}}(\cdot,\cdot)$ that maps pairs of yaw-pitch-roll angles
to the corresponding triplet of their composition.

However, a solution can be found following this path: first,
the 3D+YPR poses $\mathbf{p_6^i}$ will be converted to
3D+Quat form $\mathbf{p_7^i}$, which are then
composed such as $\mathbf{p_7^R} = \mathbf{p_6^1} \oplus \mathbf{p_6^2}$,
and finally that pose is converted back to 3D+YPR form to obtain
$\mathbf{p_6^R}$.

The chain rule can be applied to this sequence of transformations, leading to:

\begin{eqnarray}
 \left.
\frac{\partial \mathbf{f_{pc}}(\mathbf{p},\mathbf{q}) }{\partial \mathbf{p}} \right|
 _{\overset{ \mathbf{p}=\mathbf{p_6^1} }{ \mathbf{q}=\mathbf{p_6^2} }}
&=&
 \left.
\frac{\partial \mathbf{p_6} (\mathbf{p_7} ) }{\partial \mathbf{p_7}}
\right|
 _{ \mathbf{p_7}=\mathbf{p_7^R} }
 \left.
\frac{\partial \mathbf{f_{qc}} (\mathbf{p},\mathbf{q} ) }{\partial \mathbf{p}}
\right|
 _{\overset{ \mathbf{p}=\mathbf{p_7^1} }{ \mathbf{q}=\mathbf{p_7^2} }}
 \left.
\frac{\partial \mathbf{p_7} (\mathbf{p_6} ) }{\partial \mathbf{p_6}}
\right|
 _{ \mathbf{p_6}=\mathbf{p_6^1} }
\\
 \left.
\frac{\partial \mathbf{f_{pc}}(\mathbf{p},\mathbf{q}) }{\partial \mathbf{q}} \right|
 _{\overset{ \mathbf{p}=\mathbf{p_6^1} }{ \mathbf{q}=\mathbf{p_6^2} }}
&=&
 \left.
\frac{\partial \mathbf{p_6} (\mathbf{p_7} ) }{\partial \mathbf{p_7}}
\right|
 _{ \mathbf{p_7}=\mathbf{p_7^R} }
 \left.
\frac{\partial \mathbf{f_{qc}} (\mathbf{p},\mathbf{q} ) }{\partial \mathbf{q}}
\right|
 _{\overset{ \mathbf{p}=\mathbf{p_7^1} }{ \mathbf{q}=\mathbf{p_7^2} }}
 \left.
\frac{\partial \mathbf{p_7} (\mathbf{p_6} ) }{\partial \mathbf{p_6}}
\right|
 _{ \mathbf{p_6}=\mathbf{p_6^2} }
\end{eqnarray}

\noindent where the three chained Jacobians are described in
Eq.(\ref{eq:jac_p6_p7}), Eq.(\ref{eq:jacob_f_qc}) and Eq.(\ref{eq:jac_p7_p6}), respectively.

\subsubsection{Implementation in MRPT}

The composition is easily performed via an overloaded ``+'' operator, as can be seen in this code:

\begin{lstlisting}
#include <mrpt/poses/CPose3DPDFGaussian.h>
using namespace mrpt::poses;

CPose3DPDFGaussian p6a( p6_mean_a, p6_cov_a );
CPose3DPDFGaussian p6b( p6_mean_b, p6_cov_b );
...

CPose3DPDFGaussian p6 = p6a + p6b; // Pose composition (both mean and covariance)
\end{lstlisting}

\section{With poses in 3D+Quat form}

\subsection{Pose composition}

Given two poses
$\mathbf{p_1} = [x_1 ~ y_1 ~ z_1 ~  q_{r1} ~ q_{x1} ~ q_{y1} ~ q_{z1} ] ^ \top$
and
$\mathbf{p_2} = [x_2 ~ y_2 ~ z_2 ~  q_{r2} ~ q_{x2} ~ q_{y2} ~ q_{z2} ] ^ \top$,
we are interested in their composition $\mathbf{p}=\mathbf{p_1} \oplus \mathbf{p_2}$.

Operating, this pose can be found to be:

\begin{equation}
\mathbf{p} =
\left(\begin{array}{c}
 x \\ y \\ z \\ q_r \\ q_x \\ q_y \\ q_z
\end{array}\right)
= \mathbf{f_{qn}}\left(  \mathbf{f_{qc}}(\mathbf{p_1},\mathbf{p_2}) \right)
=
\mathbf{f_{qn}}
\left(\begin{array}{c}
  \mathbf{f_{qr}} (\mathbf{p_1}, [ x_2 ~ y_2 ~ z_2]^\top ) \\
q_{r1} q_{r2} - q_{x1} q_{x2} - q_{y1} q_{y2} - q_{z1} q_{z2} \\
q_{r1} q_{x2} + q_{r2} q_{x1} + q_{y1} q_{z2} - q_{y2} q_{z1} \\
q_{r1} q_{y2} + q_{r2} q_{y1} + q_{z1} q_{x2} - q_{z2} q_{x1} \\
q_{r1} q_{z2} + q_{r2} q_{z1} + q_{x1} q_{y2} - q_{x2} q_{y1}
\end{array}\right)
\end{equation}

\noindent with the function $\mathbf{f_{qr}}(\cdot)$ already defined in Eq.~\ref{eq:quat_rot_point_func}
and $\mathbf{f_{qn}}$ being the quaternion normalization function, discussed in
\S \ref{sect:quat:norm}.

\subsubsection{Implementation in MRPT}

Pose composition for 3D+Quat poses is implemented via overloading
of the ``\texttt{+}'' operator, such as composing can be simply
writen down as:

\begin{lstlisting}
#include <mrpt/poses/CPose3DQuat.h>
using namespace mrpt::poses;

CPose3DQuat  p1,p2;
...
CPose3DQuat  p = p1 + p2;    // Pose composition
\end{lstlisting}

\subsection{Uncertainty}

Let $\mathcal{N}(\mathbf{\bar{p}_1}, cov(\mathbf{p_1}))$ and
$\mathcal{N}(\mathbf{\bar{p}_2}, cov(\mathbf{p_2}))$ represent
two independent Gaussian distributions over a pair of 6D
poses in quaternion form.
Then, the probability distribution of their composition
$\mathbf{p} = \mathbf{p_1} \oplus \mathbf{p_2}$
can be approximated via linear error propagation by considering
a mean value of:

\begin{eqnarray}
\mathbf{\bar{p}} = \mathbf{\bar{p}_1} \oplus \mathbf{\bar{p}_2}
\end{eqnarray}

\noindent and a covariance matrix given by:

\begin{eqnarray}
cov(\mathbf{p}) &=&
\left. \frac{\partial \mathbf{f_{qn}} }{\partial \mathbf{p}} \right|_{\mathbf{p}=\mathbf{p_1}}
\frac{\partial \mathbf{f_{qc}}(\mathbf{p_1},\mathbf{p_2}) }{\partial \mathbf{p_1}}
cov(\mathbf{p_1})
\frac{\partial \mathbf{f_{qc}}(\mathbf{p_1},\mathbf{p_2}) }{\partial \mathbf{p_1}}^\top
\left. \frac{\partial \mathbf{f_{qn}} }{\partial \mathbf{p}} \right|_{\mathbf{p}=\mathbf{p_1}} ^\top
\nonumber \\ &+&
\left. \frac{\partial \mathbf{f_{qn}} }{\partial \mathbf{p}} \right|_{\mathbf{p}=\mathbf{p_2}}
\frac{\partial \mathbf{f_{qc}}(\mathbf{p_1},\mathbf{p_2}) }{\partial \mathbf{p_2}}
cov(\mathbf{p_2})
\frac{\partial \mathbf{f_{qc}}(\mathbf{p_1},\mathbf{p_2}) }{\partial \mathbf{p_2}}^\top
\left. \frac{\partial \mathbf{f_{qn}} }{\partial \mathbf{p}} \right|_{\mathbf{p}=\mathbf{p_2}}^\top
\end{eqnarray}

The Jacobians of the pose composition function $\mathbf{f_{qc}}(\cdot)$ are given by:

\begin{eqnarray}
\label{eq:jacob_f_qc}
\left.
\frac{\partial \mathbf{f_{qc}}(\mathbf{p_1},\mathbf{p_2}) }{\partial \mathbf{p_1}}
\right|_{7 \times 7}
=
\left(
  \begin{array}{c}
     \left. \frac{\partial \mathbf{f_{qr}} (\mathbf{p_1}, [ x_2 ~ y_2 ~ z_2]^\top ) }{\partial \mathbf{p_1}} \right|_{3\times 7} \\ \hline
  \begin{array}{ccccccc}
    &  &  			& q_{r2} &-q_{x2} &-q_{y2} &-q_{z2}  \\
    & & \mathbf{0}_{4\times 3}  	& q_{x2} & q_{r2} & q_{z2} &-q_{y2}  \\
    & &  			& q_{y2} &-q_{z2} & q_{r2} & q_{x2}  \\
    & &  			& q_{z2} & q_{y2} &-q_{x2} & q_{r2}
  \end{array}
  \end{array}
\right)
\\
\left.
\frac{\partial \mathbf{f_{qc}}(\mathbf{p_1},\mathbf{p_2}) }{\partial \mathbf{p_2}}
\right|_{7 \times 7}
=
\left(
  \begin{array}{ccc|cccc}
     & \left.
	\frac{\partial \mathbf{f_{qr}} (\mathbf{p_1}, [ x_2 ~ y_2 ~ z_2]^\top ) }{\partial [ x_2 ~ y_2 ~ z_2]^\top} \right|_{3\times 3}
      &  & & \mathbf{0}_{3 \times 4}  \\ \hline
   & & 				& q_{r1} &-q_{x1} &-q_{y1} &-q_{z1}  \\
   & \mathbf{0}_{4\times 3} &	& q_{x1} & q_{r1} &-q_{z1} & q_{y1}  \\
   & & 				& q_{y1} & q_{z1} & q_{r1} &-q_{x1}  \\
   & & 				& q_{z1} &-q_{y1} & q_{x1} & q_{r1}
  \end{array}
\right)
\end{eqnarray}

Note that the partial Jacobians used in these expressions were already defined
in Eq.~(\ref{eq:quat_rot_point_func_jacob1})-(\ref{eq:quat_rot_point_func_jacob2}),
and that the Jacobian of the normalization function $\mathbf{f_{qn}}$ is described
in \S \ref{sect:quat:norm}.

\subsubsection{Implementation in MRPT}

The composition is easily performed via an overloaded ``+'' operator:

\begin{lstlisting}
#include <mrpt/poses/CPose3DQuatPDFGaussian.h>
using namespace mrpt::poses;

CPose3DQuatPDFGaussian     p7a( p7_mean_a, p7_cov_a );
CPose3DQuatPDFGaussian     p7b( p7_mean_b, p7_cov_b );
...

CPose3DQuatPDFGaussian p7 = p7a + p7b; // Pose composition (both mean and covariance)
\end{lstlisting}

\section{With poses in matrix form}
\label{sect:comp_poses:mat}

\subsection{Pose composition}

Given a pair of $4\times 4$ transformation matrices
$\mathbf{M_1}$ and $\mathbf{M_2}$ corresponding to two 6D poses
$\mathbf{p_1}$ and $\mathbf{p_2}$, we can compute the
matrix $\mathbf{M}$ for their composition $\mathbf{p} = \mathbf{p_1} \oplus \mathbf{p_2}$
simply as:

\begin{equation}
\mathbf{M} =  \mathbf{M_1}  \mathbf{M_2}
\end{equation}

\subsubsection{Implementation in MRPT}

In this case, operate just like with ordinary matrices:

\begin{lstlisting}
#include <mrpt/math/lightweight_geom_data.h>
using namespace mrpt::math;

CMatrixDouble44  M1, M2;
...
CMatrixDouble44  M = M1 * M2;  // Matrix multiplication
\end{lstlisting}

\chapter{Inverse of a pose}
\label{sect:inverse}

Given a pose $\mathbf{p}$, we define its \emph{inverse}
(denoted as $\ominus \mathbf{p}$)
as that pose that, composed with the former, gives the null element
in $\mathbf{SE}(3)$.
In practice, it is useful to visualize the inverse of a pose as
how the origin of coordinates ''is seen'', from that pose.

\section{For a 3D+YPR pose}

In this case it's preferred to transform the 3D pose to either a 3D+Quat or a matrix form,
invert the pose in that form (as described in the next sections) and convert back to 3D+YPR.

\subsubsection{Implementation in MRPT}

Obtaining the inverse of a 6D-pose of type \texttt{CPose3D} is implemented
with the unary \texttt{-} operator which internally uses the cached $4 \times 4$
transformation matrix within \texttt{CPose3D} objects:

\begin{lstlisting}
#include <mrpt/poses/CPose3D.h>
using namespace mrpt::poses;

CPose3D   q;
CPose3D   q_inv = -q;
\end{lstlisting}

\section{For a 3D+Quat pose}
\label{sect:inverse:quat}

\subsection{Inverse}

The inverse of a pose $\mathbf{p_7} = [x ~ y ~ z ~  q_r ~ q_x ~ q_y ~ q_z ] ^ \top$
comprises two parts which can be computed separately.
If we denote this inverse as $\mathbf{p^\star_7} = [x^\star ~ y^\star ~ z^\star ~  q^\star_r ~ q^\star_x ~ q^\star_y ~ q^\star_z ] ^ \top$,
its rotational part is simply the conjugate quaternion of the original pose, while the 3D translational
part must be computed as the relative position of the origin $[0 ~ 0 ~ 0]^\top$ as seen
from the pose $\mathbf{p_7}$, that is:

\begin{eqnarray}
\mathbf{p^\star_7} =
\left(
\begin{array}{c}
x^\star \\ y^\star \\ z^\star \\  q^\star_r \\ q^\star_x \\ q^\star_y \\ q^\star_z
\end{array}
\right)
=
\mathbf{f_{qi}}( \mathbf{p_7} )
=
\left(
\begin{array}{c}
\mathbf{f_{qri}}( [0 ~ 0 ~ 0]^\top, \mathbf{p_7} )
\\  q_r \\ -q_x \\ -q_y \\ -q_z
\end{array}
\right)
\end{eqnarray}

\noindent where $\mathbf{f_{qri}}(\mathbf{a},\mathbf{p})$  was defined in Eq.~(\ref{eq:fqri}).

\subsubsection{Implementation in MRPT}

Obtaining the inverse of a 7D-pose of type \texttt{CPose3DQuat} is implemented
with the unary \texttt{-} operator:

\begin{lstlisting}
#include <mrpt/poses/CPose3DQuat.h>
using namespace mrpt::poses;

CPose3DQuat   q;
CPose3DQuat   q_inv = -q;
\end{lstlisting}

\subsection{Uncertainty}

Let $\mathcal{N}(\mathbf{\bar{q}}, cov(\mathbf{q}))$ represent
the Gaussian distributions of a 7D-pose $\mathbf{q}$ in 3D+Quat form.
Then, the probability distribution of the inverse pose
$\mathbf{q_i} = \ominus \mathbf{q_i} $ can be approximated via
linear error propagation by considering a mean value of:

\begin{eqnarray}
\mathbf{\bar{q}_i} = \ominus \mathbf{\bar{q}}
\end{eqnarray}

\noindent and a covariance matrix:

\begin{eqnarray}
cov(\mathbf{q_i}) &=&
\frac{\partial \mathbf{f_{qi}} }{\partial \mathbf{q}}
cov(\mathbf{q})
\frac{\partial \mathbf{f_{qi}} }{\partial \mathbf{q}} ^\top
\end{eqnarray}

\noindent with the Jacobian:

\begin{subeqnarray}
\label{eq:jacob.inverse.quat}
\frac{\partial \mathbf{f_{qi}} }{\partial \mathbf{q}}
&=&
\left(
\begin{array}{c}
\frac{\partial \mathbf{f_{qri}}([0~0~0]^\top,\mathbf{q}) }{\partial \mathbf{q}} \\
\hline
 \begin{array}{c|c}
   \mathbf{0}_{4 \times 3} & 
   \mathbf{D}
 \end{array}
\end{array}
\right)
\\
\mathbf{D}
&=& 
\left(
\begin{array}{rrrr}
	1 & 0 & 0 & 0 \\
	0 & -1 & 0 & 0 \\
	0 & 0 & -1 & 0 \\
	0 & 0 & 0 & -1
\end{array}
\right)
\frac{\partial (q_r',q_x',q_y',q_z')(q_r,q_x,q_y,q_z)}{\partial q_r,q_x,q_y,q_z}
\end{subeqnarray}

\noindent where the sub-Jacobian on the top has been already defined in Eq.~(\ref{eq:df_qri_p}) and the normalization Jacobian 
is defined \S \ref{sect:quat:norm}.

\subsubsection{Implementation in MRPT}

The Gaussian distrution of an inverse 3D+Quat pose can be computed simply by:

\begin{lstlisting}
#include <mrpt/poses/CPose3DQuatPDFGaussian.h>
using namespace mrpt::poses;

CPose3DQuatPDFGaussian  p1 = ...
CPose3DQuatPDFGaussian  p1_inv = -p1;
\end{lstlisting}

\section{For a transformation matrix}
\label{sect:inverse:mat}

From the description of inverse pose at the begining of this chapter, and given
that the null element in $\mathbf{SE}(3)$ in matrix form is the identity $\mathbf{I}_4$,
it's clear that the inverse of pose defined by a matrix $\mathbf{M}$ is simply $\mathbf{M}^{-1}$,
since $\mathbf{M}^{-1}\mathbf{M}=\mathbf{I}$.

The inverse of a homogeneous matrix can be computed very efficiently by
simply transposing its $3 \times 3$ rotation part (which actually requires \emph{just 3 swaps})
and using the following expressions for the fourth column (the translation):

\begin{eqnarray}
M^{-1} &=&
\left(
  \begin{array}{ccc|c}
   \mathbf{i} & \mathbf{j} & \mathbf{k} & \mathbf{t} \\
\hline
   0 & 0 & 0 & 1
  \end{array}
\right) ^{-1}
=
\left(
  \begin{array}{ccc|c}
   i_1 & j_1 & k_1 & x \\
   i_2 & j_2 & k_2 & y \\
   i_3 & j_3 & k_3 & z \\
\hline
   0 & 0 & 0 & 1
  \end{array}
\right) ^{-1}
=
\left(
  \begin{array}{ccc|c}
   i_1 & i_2 & i_3   & -\mathbf{i} \cdot \mathbf{t} \\
   j_1 & j_2 & j_3   & -\mathbf{j} \cdot \mathbf{t} \\
   k_1 & k_2 & k_3   & -\mathbf{k} \cdot \mathbf{t} \\
\hline
   0 & 0 & 0 & 1
  \end{array}
\right)
\end{eqnarray}

\noindent where $\mathbf{a} \cdot \mathbf{b}$ stands for the dot product.
See also \S\ref{sect:mat_deriv:exp} for derivatives of this transformation,
under the form of matrix derivatives.

\chapter{Derivatives of pose transformation matrices}
\label{chap:mat_deriv}

\section{Operators}
\label{sect:mat_deriv:ops}

The following operators are extremely useful when dealing with derivatives of matrices:

\begin{itemize}
\item{The $vec$ operator. It stacks all the columns of an $M \times N$ matrix to form a $MN\times 1$ vector.
Example:
\begin{equation}
  vec\left( \left[
    \begin{array}{ccc}
      1 & 2 & 3 \\
      4 & 5 & 6
    \end{array}
 \right] \right) =
\left(
\begin{array}{c}
1 \\ 4 \\ 2 \\ 5 \\ 3 \\ 6
\end{array}
\right)
\end{equation}
}
\item{The \textbf{Kronecker operator}, or matrix direct product.
Denoted as $\mathbf{A} \otimes \mathbf{B}$ for any two matrices
$\mathbf{A}$ and $\mathbf{B}$ of dimensions $M_A \times N_A$
and $M_B \times N_B$, respectively,
it gives a tensor product of the matrices as an $M_AM_B \times N_A N_B$ matrix.
That is,
\begin{equation}
 \mathbf{A} \otimes \mathbf{B}
 =
\left(
\begin{array}{cccc}
a_{11} \mathbf{B}  & a_{12} \mathbf{B} & a_{13} \mathbf{B}  & ... \\
a_{21} \mathbf{B}  & a_{22} \mathbf{B} & a_{23} \mathbf{B}  & ... \\
 & ... &
\end{array}
\right)
\end{equation}
}
\item{The \textbf{transpose permutation matrix}. Denoted as $\mathbf{T_{M,N}}$,
these are simple permutation matrices of size $MN \times MN$ containing all 0s
but for just one 1 at each column or row,
such as for any $M \times N$ matrix $\mathbf{A}$ it holds:
\begin{equation}
\mathbf{T_{N,M}} vec(\mathbf{A}) = vec(\mathbf{A}^\top)
\end{equation}
}
\item{The \textbf{hat (wedge) operator} $(\cdot)^\wedge$, maps a $3\times 1$ vector to its corresponding skew-symmetric matrix:
	\begin{equation}
	\label{eq:skew}
	\bm{\omega} = \left[ \begin{array}{c} x \\ y \\z \end{array} \right]
	\quad \quad
	{\bm{\omega}}^\wedge
	=
	\left(
	\begin{array}{ccc}
	0 & -z & y  \\
	z & 0 & -x  \\
	-y & x & 0
	\end{array}
	\right)
	\end{equation}
}
\item{The \textbf{vee operator} $(\cdot)^{\vee}$, is the inverse of the hat map:
	\begin{equation}
	\label{eq:inv.skew}
	\left(
	\begin{array}{ccc}
	0 & -z & y  \\
	z & 0 & -x  \\
	-y & x & 0
	\end{array}
	\right)
	^{\vee}
	=
	\left[ \begin{array}{c} x \\ y \\z \end{array} \right]
	\end{equation}
	\noindent, such that $({\bm{\omega}}^\wedge)^\vee = \bm{\omega}$.
}
\end{itemize}

\section{On the notation}
\label{sect:mat_deriv:not}

Previous chapters have discussed three popular ways of representing 6D poses,
namely, 3D+YPR, 3D+Quat and $4\times 4$ transformation matrices.
In the following we will be only interested in the matrix form, which will be
described here once again to stress the relevant facts for this chapter.

A pose (rigid transformation) in three-dimensional Euclidean space can be uniquely determined by means
of a $4 \times 4$ matrix with this structure:

\begin{equation}
\label{eq:T_Rt}
 \mathbf{T} =
\left(
\begin{array}{c|c}
  \mathbf{R} & \mathbf{t} \\
\hline
  \mathbf{0}_{1\times 3} & 1
\end{array}
\right)
\end{equation}

\noindent where $\mathbf{R} \in \mathbf{SO}(3)$
is a proper rotation matrix (see \S\ref{sect:basic}) and $\mathbf{t}=[t_x ~ t_y ~ t_z]^\top \in \mathbb{R}^3$ is a translation vector.
In general, any invertible $4 \times 4$ matrix belongs to the
general linear group $\mathbf{GL}(4,\mathbb{R})$, but
matrices in the form above belongs to $\mathbf{SE}(3)$,
which actually is the manifold $\mathbf{SO}(3) \times \mathbb{R}^3$
embedded in the more general $\mathbf{GL}(4,\mathbb{R})$.
The point here is to notice that the manifold has a dimensionality of 12:
9 coordinates for the $3\times 3$ matrix plus other 3 for the translation vector.

Since we will be interested here in expressions involving \emph{derivatives} of functions of poses,
we need to define a clear notation for what a derivative of a matrix actually means.
As an example, consider an arbitrary function, say,
the map of pairs of poses $p_1$, $p_2$ to their composition $p_1 \oplus p_2$, that is,
$f_\oplus: \mathbf{SE}(3) \times \mathbf{SE}(3) \mapsto \mathbf{SE}(3)$.
Then, what does the expression

\begin{equation}
\label{eq:deriv_sample1}
\frac{\partial f_\oplus(p_1,p_2)}{\partial p_1}
\end{equation}

\noindent means?
If $p_i$ were scalars, the expression would be a standard 1-dimensional derivative.
If they were vectors, the expression would become a Jacobian matrix.
But they are \emph{poses}, thus some kind of convention on how a pose is \emph{parameterized}
must be made explicit to understand such an expression.

As also considered in other works, e.g. \cite{strasdat2010scale}, poses will be treated
as matrices.
When dealing with derivatives of matrices it is convention to implicitly assume
that all the involved matrices are actually expanded with the $vec$ operator (see \S\ref{sect:mat_deriv:ops}),
meaning that derivatives of matrices become standard Jacobians.
However, for matrices describing rigid motions we will only expand
the top $3 \times 4$ submatrix; the fourth row of \ref{eq:T_Rt} can be discarded since it is fixed.

To sum up: poses appearing in a derivative expressions are replaced by their $4 \times 4$
matrices, but when expanding them with the $vec$ operator, the last row is discarded.
Poses become 12-vectors.
Although this implies a clear over-parameterization of an entity with 6 DOFs, it turns
out that many important operations become linear with this representation, enabling us to
obtain exact derivatives in an efficient way.

Recovering the example in Eq.\ref{eq:deriv_sample1}, if we denote the
transformation matrix associated to $p_i$ as $\mathbf{T}_i$, we have:

\begin{equation}
\label{eq:deriv_sample2}
\frac{\partial f_\oplus(p_1,p_2)}{\partial p_1} =
\left.
\frac{\partial f_\oplus(\mathbf{T}_1,\mathbf{T}_2)}{\partial \mathbf{T}_1}
\right|_{12 \times 12}
\end{equation}

It is instructive to explicitly unroll at least one such expression.
Using the standard matrix element subscript notation, i.e:

\begin{equation}
\label{eq:deriv_sample2b}
\mathbf{M} =
\left(
\begin{array}{cccc}
 m_{11} & m_{12} & m_{13} & m_{14} ~~  \\
 m_{21} & m_{22} & m_{23} & m_{24}  \\
 m_{31} & m_{32} & m_{33} & m_{34}  \\
 \cancelto{0}{m_{41}} & \cancelto{0}{m_{42}} & \cancelto{0}{m_{43}} & \cancelto{1}{m_{44}}
\end{array}
\right)
\end{equation}

\noindent and denoting the resulting matrix from $f_\oplus(p_1,p_2)$ as $\mathbf{F}$:

\begin{eqnarray}
\label{eq:deriv_sample3}
\frac{\partial f_\oplus(p,q)}{\partial p} =
\frac{\partial \mathbf{F} (\mathbf{P},\mathbf{Q})}{\partial \mathbf{P}}
&=&
\frac{\partial vec( \mathbf{F} (\mathbf{P},\mathbf{Q})) }{\partial vec(\mathbf{P}) }
\\
&=&
\frac{\partial [
f_{11} f_{21} f_{31}
f_{12} f_{22} ...
f_{33} f_{14} f_{24} f_{34}] }
{\partial [
p_{11} p_{21} p_{31}
p_{12} p_{22} ...
p_{33} p_{14} p_{24} p_{34}] }
=
\left(
\begin{array}{cccc}
 \frac{\partial f_{11}}{\partial  p_{11} } &  \frac{\partial f_{11}}{\partial  p_{21} } & ...  & \frac{\partial f_{11}}{\partial  p_{34} } \\
  ... & ... & ... & ... \\
 \frac{\partial f_{34}}{\partial  p_{11} } &  \frac{\partial f_{34}}{\partial  p_{21} } & ...  & \frac{\partial f_{34}}{\partial  p_{34} }
\end{array}
\right)_{12 \times 12}
\end{eqnarray}

\section{Useful expressions}
\label{sect:mat_deriv:exp}

Once defined the notation, we can give the following list of
useful expressions which may arise when working with derivatives of transformations,
as when dealing with optimization problems -- see \S\ref{chap:se3_lie:deriv}.

\subsection{Pose-pose composition}
\label{sect:jacob_pose_pose_comp}

Let $f_\oplus: \mathbf{SE}(3) \times \mathbf{SE}(3) \mapsto \mathbf{SE}(3)$ denote
the pose composition operation,
such as $f_\oplus(A,B) = A \oplus B$
(refer to \S\ref{sect:basic} and \S\ref{ch:pose_pose_comp}).
Then we can take derivatives of $f_\oplus(A,B)$ w.r.t. both involved poses $A$ and $B$.

If we denote the $4 \times 4$ transformation matrix associated to a pose $X$ as:

\begin{equation}
 \mathbf{T_X} =
\left(
\begin{array}{c|c}
  \mathbf{R_X} & \mathbf{t_X} \\
\hline
  \mathbf{0}_{1\times 3} & 1
\end{array}
\right)
\end{equation}

\noindent the matrix multiplication $\mathbf{T_A}\mathbf{T_B}$ can be
expanded element by element and, rearranging terms, it can be easily shown that:

\begin{eqnarray}
\label{eq:oplus.ab.wrt.a}
\frac{\partial f_\oplus(A,B) }{\partial A } &=&
\frac{\partial \mathbf{T_A}\mathbf{T_B} }{\partial \mathbf{T_A} } = \mathbf{T_B}^\top \otimes \mathbf{I_3}
\quad \quad \quad
\text{(a $12 \times 12$ Jacobian)}
\\
\label{eq:oplus.ab.wrt.b}
\frac{\partial f_\oplus(A,B) }{\partial B } &=&
\frac{\partial \mathbf{T_A}\mathbf{T_B} }{\partial \mathbf{T_B} } =
\mathbf{I_4} \otimes   \mathbf{R_A}
\quad \quad \quad
\text{(a $12 \times 12$ Jacobian)}
\end{eqnarray}

These Jacobians are provided in MRPT via \texttt{mrpt::poses::Lie::SE<3>}, methods \texttt{jacob\_dAB\_dA()} and \texttt{jacob\_dAB\_dB()}, respectively.

\subsection{Pose-point composition}

Let $g_\oplus: \mathbf{SE}(3) \times \mathbb{R}^3 \mapsto \mathbb{R}^3$ denote
the pose-point composition operation
such as $g_\oplus(A,p) = A \oplus p$
(refer to \S\ref{ch:comp_pose_pt}).
Then we can take derivatives of $g_\oplus(A,p)$ w.r.t. either the pose $A$ or the point $p$.

We obtain in this case:

\begin{eqnarray}
\label{eq:jac_dAp_p}
\frac{\partial g_\oplus(A,p) }{\partial p } &=&
\frac{\partial \mathbf{T_A}\mathbf{p} }{\partial \mathbf{p} } =
\frac{\partial (\mathbf{R_A}\mathbf{p} + \mathbf{t_A})}{\partial \mathbf{p} } =
\mathbf{R_A}
\quad \quad \quad
\text{(a $3 \times 3$ Jacobian)}
\\
\label{eq:jac_dAp_A}
\frac{\partial g_\oplus(A,p) }{\partial A } &=&
\frac{\partial \mathbf{T_A}\mathbf{p} }{\partial \mathbf{T_A} } =
\left( \mathbf{p}^\top ~~ 1 \right) \otimes \mathbf{I_3}
\quad \quad \quad
\text{(a $3 \times 12$ Jacobian)}
\end{eqnarray}

\subsection{Inverse of a pose}

The inverse of a pose $A$ is given by the inverse of its associated matrix $\mathbf{T_A}$,
which always exists and has a closed form expression (see \S\ref{sect:inverse:mat}).
Its derivative can be shown to be:

\begin{eqnarray}
\label{eq:jac_dInv_A}
\frac{\partial \left( \mathbf{T_A} ^{-1} \right) }{\partial \mathbf{T_A} } &=&
\left(
\begin{array}{cc}
 \mathbf{T_{3,3}} & \mathbf{0}_{9 \times 3} \\
 \mathbf{I_3} \otimes (-\mathbf{t_A}^\top) & - \mathbf{R_A}^\top
\end{array}
\right)
\quad \quad \quad
\text{(a $12 \times 12$ Jacobian)}
\end{eqnarray}

Remember that $\mathbf{T_{3,3}}$ stands for a transpose permutation matrix (of size $9 \times 9$ in this case),
as defined in \S\ref{sect:mat_deriv:ops}.

\subsection{Inverse pose-point composition}

Employing the above defined Jacobians and the standard chain rule for derivatives one can
obtain arbitrarily complex Jacobians. As an example, it will derived here
the derivative of pose-point inverse composition, that is,
given a pose $A$ and a point $p$, obtaining $p \ominus A$, or $\mathbf{A}^{-1} \mathbf{p}$
(see \S\ref{ch:inv_pose_point}).

Operating:

\begin{eqnarray}
\label{eq:jac_dp_min_p}
\frac{\partial \left( \mathbf{T_A} ^{-1} \mathbf{p} \right) }{\partial \mathbf{p} }
&\overset{\text{Eq.(\ref{eq:jac_dAp_p})}}{=}&
(\mathbf{R_A})^{-1}
=\mathbf{R_A}^\top
\quad \quad \quad
\text{(a $3 \times 3$ Jacobian)}
\\
\frac{\partial \left( \mathbf{T_A} ^{-1} \mathbf{p} \right) }{\partial \mathbf{T_A} }
& \overset{\text{Chain rule}}{=}&
\frac{\partial \left( \mathbf{T_A} ^{-1} \mathbf{p} \right) }{\partial (\mathbf{T_A}^{-1}) }
\frac{\partial \left( \mathbf{T_A} ^{-1} \right) }{\partial \mathbf{T_A} }
\overset{ \scriptsize{ \begin{array}{c} \text{Eq.(\ref{eq:jac_dAp_A}) \&} \\  \text{Eq.(\ref{eq:jac_dInv_A})} \end{array} } }
{=}
\left[
\left( \mathbf{p}^\top ~~ 1 \right) \otimes \mathbf{I_3}
\right]
\left(
\begin{array}{cc}
 \mathbf{T_{3,3}} & \mathbf{0}_{3 \times 9} \\
 \mathbf{I_3} \otimes (-\mathbf{t_A}^\top) & - \mathbf{R_A}^\top
\end{array}
\right)
\\
&=&
\label{eq:jac_dp_min_A}
\left(
\begin{array}{c|c}
 \mathbf{I_3} \otimes \left( (\mathbf{p}-\mathbf{t_A})^\top \right)
 &
 -\mathbf{R_A}^\top
\end{array}
\right)
\quad \quad \quad
\text{(a $3 \times 12$ Jacobian)}
\end{eqnarray}

\chapter{Concepts on Lie groups}
\label{chap:lie_intro}

\section{Definitions}
\label{sect:lie_defs}

Before addressing the practical applications of looking at rigid motions
as a Lie group,
we need to provide several mathematical definitions which are fundamental
to understand the subsequent discussion
(for example, what a Lie group actually is!).
A more in-deep treatment of some of the topics covered in this chapter
can be found in \cite{gallier2001geometric,varadarajan1974lie}.

\subsection{Mathematical group}
\label{sect:defs:group}

A group $G$ is a structure consisting of a
finite or infinite set of elements plus
some binary operation (the \emph{group operation}),
which for any two group elements $A,B \in G$
is denoted as the multiplication $A B$.

A group is said to be a group \emph{under} some given operation if
it fulfills the following conditions:
\begin{enumerate}
 \item \textbf{Closure.} The group operation is a function $G \times G \mapsto G$,
that is, for any $A,B \in G$, we have $A B \in G$.
 \item \textbf{Associativity.} For $A,B,C \in G$, $(AB)C = A(BC)$.
 \item \textbf{Identity element.} There must exists an identity element $I \in G$,
such as $IA=AI=A$ for any $A \in G$.
 \item \textbf{Inverse.} For any $A \in G$ there must exist an inverse element $A^{-1}$
such as $A A^{-1} = A^{-1} A = I$.
\end{enumerate}

Examples of simple groups are:
\begin{itemize}
 \item The integer numbers $\mathbb{Z}$, under the operation of addition.
 \item The sets of invertible $N \times N$ matrices $\mathbf{GL}(N,\mathbb{R})$,
or the 3D special orthogonal group $\mathbf{SO}(3)$ (recall \S\ref{sect:basic}) are groups
under the operation of standard matrix multiplication.
\end{itemize}

\subsection{Manifold}

An $N$-dimensional manifold $M$ is a topological space where
every point $p \in M$ is endowed with \emph{local} Euclidean structure.
Another way of saying it: the neighborhood of every point $p$
is homeomorphic\footnote{A
function that maps from $M$ to $\mathbb{R}^N$ is homeomorphic if it is
a bicontinuous function, that is, both $f(\cdot)$ and its inverse $f(\cdot)^{-1}$
are continuous.} to $\mathbb{R}^N$.

\begin{figure}
\centering
\includegraphics[width=0.40\textwidth]{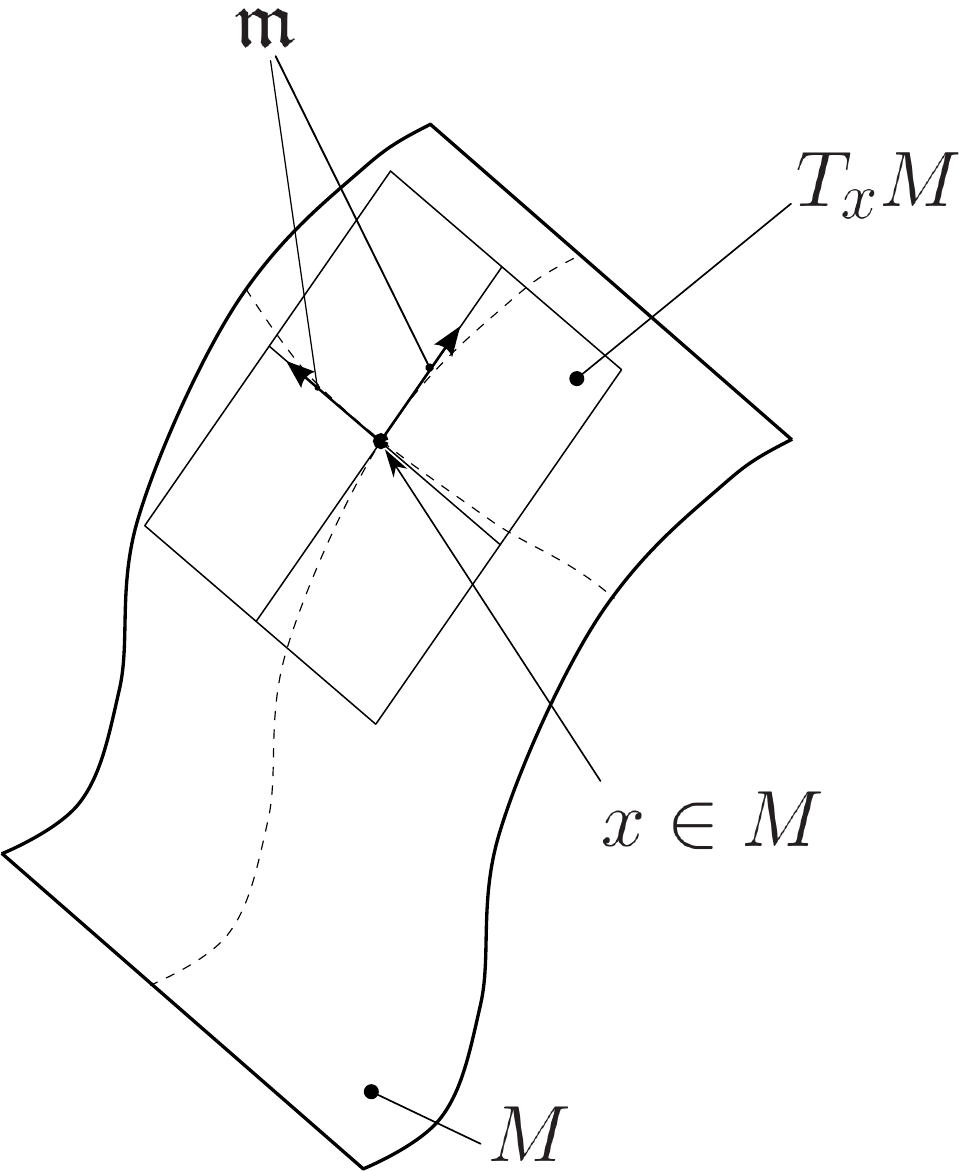}
\caption{An illustration of the elements introduced in the text:
a sample 2-dimensional manifold $M$ (embedded in 3D-space),
a point on it $x \in M$, the tangent space at $x$,
denoted $T_x M$ and the algebra $\mathfrak{m}$, the vectorial base of that space.
}
\label{fig:manifold}
\end{figure}

From an intuitive point of view, it means that, in an infinitely small vicinity of
a point $p$ the space looks ``flat''. A good way to visualize it is to think
of the surface of the Earth, a manifold of dimension 2 (we can move in two perpendicular
directions, North-South and East-West). Although it is curved, at a given point
it looks ``flat'', or a $\mathbb{R}^2$ Euclidean space (refer to Fig.~\ref{fig:manifold}).

\subsection{Smooth manifolds embedded in $\mathbb{R}^N$}
\label{sect:smooth_manif}

A $D$-dimensional manifold is a smooth manifold embedded in the
$\mathbb{R}^N$ space ($N \geq D$) if every point $p \in M$ is
contained by $U \subseteq M$, defined by some function:

\begin{eqnarray}
 \varphi: & \Omega & \mapsto U \\
              & \mathbb{R}^N &  \mapsto M
\end{eqnarray}

\noindent where $\Omega$ is an open subset of $\mathbb{R}^N$
which contains the origin of that space (i.e. $\mathbf{0}_N$).

Additionally, the function $\varphi$ must fulfill:
\begin{enumerate}
 \item Being a homeomorphism (i.e. $\varphi(\cdot)$ and $\varphi(\cdot)^{-1}$ are continue).
 \item Being smooth ($C^{\infty}$).
 \item Its derivative at the origin $\varphi'(\mathbf{0}_N)$ must be injective.
\end{enumerate}

The function $\varphi()$ is a \emph{local parameterization} of M
centered at the point $p$, where:

\begin{eqnarray}
 \varphi(\mathbf{0}_N) = p \quad  \quad  \quad  , p \in M
\end{eqnarray}

The inverse function:

\begin{eqnarray}
 \varphi^{-1}: & U & \mapsto \Omega  \\
              & M  &  \mapsto \mathbb{R}^N
\end{eqnarray}

\noindent is called a \emph{local chart} of M, since provides a ``flattened''
representation of an area of the manifold.

\subsection{Tangent space of a manifold}
\label{sect:tang_space}

A $D$-dimensional manifold $M$ embedded in $\mathbb{R}^N$
(with $N \geq D$) has associated an $N$-dimensional tangent space
for every point $p \in M$.
This space is denoted as $T_x M$ and in non-singular points
has a dimensionality of $D$ (identical to that of the manifold).
See Fig.~\ref{fig:manifold} for an illustration of this concept.

Informally, a tangent space can be visualized as the vector space
of the derivatives at $p$ of all possible smooth curves that pass
through $p$, e.g. $T_x M$ contains all the possible
``velocity'' vectors of a particle at $p$ and constrained to $M$.

\subsection{Lie group}
\label{sect:lie_group_def}

A Lie group is a (non-empty) subset $G$ of $\mathbb{R}^N$ that fulfills:
\begin{enumerate}
 \item $G$ is a group (see \S\ref{sect:defs:group}).
 \item $G$ is a manifold in $\mathbb{R}^N$ (see \S\ref{sect:smooth_manif}).
 \item Both, the group product operation ($\cdot: G \mapsto G$)
and its inverse (${}^{-1}: G \mapsto G$) are smooth functions.
\end{enumerate}

\subsection{Linear Lie groups (or \emph{matrix groups})}
\label{sect:lie:linear}

Let the set of all $N \times N$ matrices (invertible or not)
be denoted as $\mathbf{M}(N,\mathbb{R})$.
We also define the Lie bracket operator $[\cdot,\cdot]$ such as
$[A,B]=AB-BA$ for any $A,B \in \mathbf{M}(N,\mathbb{R})$.

Then, a theorem from Von Newman and Cartan reads (\cite{gallier2001geometric}, p.397):

\begin{mytheorem}
A closed subgroup $G$ of $\mathbf{GL}(N,\mathbb{R})$
is a linear Lie group (thus, a smooth manifold in $\mathbb{R}^{N^2}$).
Also, the set $\mathfrak{g}$:

\begin{equation}
  \mathfrak{g} = \{ \mathbf{X} \in \mathbf{M}(N,\mathbb{R}) | e^{t\mathbf{X}} \in G,  \forall t \in \mathbb{R}  \}
\end{equation}

\noindent is a vector space equal to $T_I G$ (the tangent space of $G$
at the identity entity $I$), and $\mathfrak{g}$ is closed under the Lie bracket.
\end{mytheorem}

It must be noted that, for any square matrix $\mathbf{M}$,
the exponential map $e^{\mathbf{M}}$ is well defined and coincides with the matrix
exponentiation, which in general has this (always convergent) power series form:

\begin{equation}
\label{eq:exp_map_matrices}
e^{\mathbf{M}} = \sum_{k=0}^\infty \frac{1}{k!} \mathbf{M}^k
\end{equation}

For the purposes of this report, the interesting result of the theorem
above is that the group $\mathbf{SO}(3)$ (proper rotations in $\mathbb{R}^3$)
can be also viewed now as a linear Lie group, since it is a
subgroup of $\mathbf{GL}(3,\mathbb{R})$.
Regarding the group of rigid transformations $\mathbf{SE}(3)$,
since it is isomorphic to a subset of $\mathbf{GL}(4,\mathbb{R})$
(any pose in $\mathbf{SE}(3)$ can be represented as a $4\times 4$ matrix),
we find out that it is also a linear Lie group \cite{gallier2001geometric}.

\subsection{Lie algebra}

A Lie algebra\footnote{For our purposes, an algebra means a vector space $A$ plus
a bilinear multiplication function: $A \times A \mapsto A$.}
is an algebra $\mathfrak{m}$ together with a Lie bracket operator
$[\cdot,\cdot]: \mathfrak{m} \times \mathfrak{m} \mapsto \mathfrak{m}$
such as for any elements $a,b,c \in \mathfrak{m}$
it holds:

\begin{eqnarray}
 & [a,b] = -[b,a]   & \quad\quad\quad \text{(Anti-commutativity)} \\
 & \left[ c, \left[a,b \right] \right] =
\left[ \left[c,a\right] , b \right] +
\left[ a,\left[c,b\right] \right]  &  \quad\quad\quad \text{(Jacobi identity)}
\end{eqnarray}

It follows that $[a,a]=0$ for any $a \in \mathfrak{m}$.

An important fact is that the Lie algebra $\mathfrak{m}$
associated to a Lie group $M$ happens to be the tangent
space at the identity element $\mathbf{I}$, that is:

\begin{equation}
 \mathfrak{m} = T_\mathbf{I} M   \quad\quad\quad\text{(For $M$ being a Lie group)}
\end{equation}

\subsection{Exponential and logarithm maps of a Lie group}
\label{sect:exp_ln}

Associated to a Lie group $M$ and its
Lie algebra $\mathfrak{m}$ there are
two important functions:

\begin{itemize}
\item{\textbf{The exponential map}, which maps elements from
the algebra to the manifold and determines the local structure
of the manifold:
\begin{equation}
 \exp: \mathfrak{m}  \mapsto M
\end{equation}
}
\item{\textbf{The logarithm map}, which maps elements from
the manifold to the algebra:
\begin{equation}
 \log: M  \mapsto \mathfrak{m}
\end{equation}
}
\end{itemize}

The next chapter will describe these functions for
the cases of interest in this report.

\chapter{$\mathbf{SE}(3)$ as a Lie group}
\label{chap:se3_lie}

\section{Properties}

For the sake of clarity, we repeat here the description of
the group of rigid transformations in $\mathbb{R}^3$ already given
in \S\ref{sect:mat_deriv:not}.
This group of transformations is denoted as $\mathbf{SE}(3)$,
and its members are the set of $4 \times 4$ matrices with this structure:

\begin{equation}
\label{eq:T_Rt_bis}
 \mathbf{T} =
\left(
\begin{array}{c|c}
  \mathbf{R} & \mathbf{t} \\
\hline
  \mathbf{0}_{1\times 3} & 1
\end{array}
\right)
\end{equation}

\noindent with $\mathbf{R} \in \mathbf{SO}(3)$,
$\mathbf{t}=[t_x ~ t_y ~ t_z]^\top \in \mathbb{R}^3$
and group product the standard matrix product.

~\\

Some facts on this group (see for example, \cite{gallier2001geometric}, \S14.6):

\begin{itemize}
 \item $\mathbf{SE}(3)$ is a 6-dimensional manifold (i.e. has 6 degrees of freedom). Three correspond to the 3D translation vector and the other three to the rotation.
 \item  $\mathbf{SE}(3)$ is isomorphic to a subset of $\mathbf{GL}(4,\mathbb{R})$.
 \item Since $\mathbf{SE}(3)$ is embedded in the more general $\mathbf{GL}(4,\mathbb{R})$,
         from \S\ref{sect:lie:linear} we have that it is also a Lie group.
 \item  $\mathbf{SE}(3)$ is diffeomorphic to $\mathbf{SO}(3) \times \mathbb{R}^3$ as a manifold,
  where each element is described by $3\cdot 3+3=12$ coordinates (see \S\ref{sect:mat_deriv:not}).
 \item  $\mathbf{SE}(3)$ is \emph{not} isomorphic to $\mathbf{SO}(3) \times \mathbb{R}^3$ as a group,
since the group multiplications of both groups are different.
It is said that $\mathbf{SE}(3)$ is a \emph{semidirect product} of
the groups $\mathbf{SO}(3)$ and $\mathbb{R}^3$.
\end{itemize}

\section{Lie algebra of $\mathbf{SO}(3)$}

Since $\mathbf{SE}(3)$ has the manifold structure of the product
$\mathbf{SO}(3) \times \mathbb{R}^3$, it makes sense to define first
the properties of $\mathbf{SO}(3)$, which is also a Lie group
(by the way, $\mathbb{R}^N$ can be also considered a Lie group for any $N \geq 1$).

The group $\mathbf{SO}(3)$ has an associated Lie algebra $\mathfrak{so}(3)$,
whose base are three skew symmetric matrices, each corresponding to
infinitesimal rotations along each axis:

\begin{eqnarray}
\mathfrak{so}(3) &=& \{  \mathbf{G^{\mathfrak{so}(3)}_i} \}_{i=1,2,3}  \\
\mathbf{G^{\mathfrak{so}(3)}_1} &=& \hatop{\mathbf{e}}_1
=
\left(
\begin{array}{ccc}
0 & 0 & 0  \\
0 & 0 & -1  \\
0 & 1 & 0
\end{array}
\right) 
\quad\quad
\mathbf{e}_1 = \left[ \begin{array}{c} 1 \\ 0 \\0 \end{array} \right]
\\
\mathbf{G^{\mathfrak{so}(3)}_2} &=& \hatop{\mathbf{e}}_2
=
\left(
\begin{array}{ccc}
0 & 0 & 1  \\
0 & 0 & 0  \\
-1 & 0 & 0
\end{array}
\right)
\quad\quad
\mathbf{e}_2 = \left[ \begin{array}{c} 0 \\ 1 \\0 \end{array} \right]
\\
\mathbf{G^{\mathfrak{so}(3)}_3} &=& \hatop{\mathbf{e}}_3
=
\left(
\begin{array}{ccc}
0 & -1 & 0  \\
1 & 0 & 0  \\
0 & 0 & 0
\end{array}
\right)
\quad\quad
\mathbf{e}_3 = \left[ \begin{array}{c} 0 \\ 0 \\1 \end{array} \right]
\\
\end{eqnarray}

Notice that this means that an arbitrary element in $\mathfrak{so}(3)$
has three coordinates (each coordinate multiplies a generator
matrix $\{  \mathbf{G^{\mathfrak{so}(3)}_1}, \mathbf{G^{\mathfrak{so}(3)}_2}, \mathbf{G^{\mathfrak{so}(3)}_3} \}$) so it can
be represented as a vector in $\mathbb{R}^3$.

We have used above the so-called "hat" and "vee" operators (see \S\ref{sect:mat_deriv:ops}).

\section{Lie algebra of $\mathbf{SE}(3)$}

The group $\mathbf{SE}(3)$ has an associated Lie algebra $\mathfrak{se}(3)$,
whose base are these six $4 \times 4$ matrices, each corresponding to
either infinitesimal rotations or infinitesimal translations along each axis:

\begin{eqnarray}
\mathfrak{se}(3) &=& \{  \mathbf{G^{\mathfrak{se}(3)}_i} \}_{i=1...6}  \\
\mathbf{G^{\mathfrak{se}(3)}_{\{1,2,3\}} } &=&
\left(
\begin{array}{c|c}
 \mathbf{G^{\mathfrak{so}(3)}_{\{1,2,3\}} } & \begin{array}{c} 0 \\0 \\0 \end{array} \\
\hline    0 & 0
\end{array}
\right)
\\
\mathbf{G^{\mathfrak{se}(3)}_4 } &=&
\left(
\begin{array}{c|c}
 \mathbf{0}_{3\times 3} & \begin{array}{c} 1 \\0 \\0 \end{array} \\
\hline    0 & 0
\end{array}
\right)
\\
\mathbf{G^{\mathfrak{se}(3)}_5 } &=&
\left(
\begin{array}{c|c}
 \mathbf{0}_{3\times 3} & \begin{array}{c} 0 \\1 \\0 \end{array} \\
\hline    0 & 0
\end{array}
\right)
\\
\mathbf{G^{\mathfrak{se}(3)}_6 } &=&
\left(
\begin{array}{c|c}
 \mathbf{0}_{3\times 3} & \begin{array}{c} 0 \\0 \\1 \end{array} \\
\hline    0 & 0
\end{array}
\right)
\end{eqnarray}

Recall that this means that an arbitrary element in $\mathfrak{se}(3)$
has six coordinates (each coordinate multiplies a generator matrix)
so it can be represented as a vector in $\mathbb{R}^6$.
In this consists what is called the ``linearization'' of the manifold
$\mathbf{SE}(3)$.

\section{Exponential and logarithm maps}
\label{sect:se3.exp.log}

As defined in \S\ref{sect:exp_ln}, the exponential and logarithm maps
transform elements between Lie groups and their corresponding
Lie algebras.
In this report we sometimes denote the $\exp$ and $\log$ functions
as operating on vectors and returning vectors, respectively,
of the corresponding dimensions (3 for $\mathbf{SO}(3)$, 6 for $\mathbf{SE}(3)$).
Those vectors are the coordinates in the
vector spaces of matrices defined by the corresponding Lie algebras.

\subsection{For $\mathbf{SO}(3)$}
\label{eq:exp.log.so3}

\subsubsection{Exponential map}

\Paragraph{Axis-angle to Matrix}

The map:

\begin{subeqnarray}
  \exp: \mathfrak{so}(3) & \mapsto & \mathbf{SO}(3) \\
           \W & \mapsto & \mathbf{R}_{3\times 3}
\end{subeqnarray}

\noindent is well-defined, surjective, and
corresponds to the matrix exponentiation (see Eq.~(\ref{eq:exp_map_matrices})),
which has the closed-form solution:
the  Rodrigues' formula from 1840 \cite{altafini2000cas}, that is

\begin{equation}
\label{eq:rodrigues}
  e^ { \W } \equiv  \mathrm{matexp}(\hatop{\W} ) =
\mathbf{I_3}
+ \frac{\sin \theta}{\theta} \hatop{\W}
+ \frac{1- \cos \theta}{\theta^2} (\hatop{\W})^2
\end{equation}

\noindent where
the angle $\theta = |\W|$ and
$\hatop{\W}$ is the skew symmetric matrix (see the definition of 
the hat operator in Eq.(\ref{eq:skew}))
generated by the 3-vector $\W$. 

We can define a unit vector, representing the axis of rotation, as 
$\mathbf{n} = \frac{\W}{|\W|} = (n_1, n_2, n_3)^\top$ with respect to a
fixed Cartesian coordinate system, and the angle of rotation $\theta =
|\W|$ around this axis. One can show that the Rodrigues' formula (eq.
\ref{eq:rodrigues}) for the rotation matrix $\mathbf{R}(\mathbf{n},
\theta)$ representing rotation around axis $\mathbf{n}$ about the
angle $\theta$ in the coordinate form can be written as:

\begin{eqnarray}
\label{eq:rodrigues_coordinate}
\mathbf{R}(\mathbf{n}, \theta) = 
\begin{pmatrix}
\cos \theta + n_1^2 (1 - \cos \theta) & n_1 n_2 (1 - \cos \theta) -
n_3 \sin \theta & n_1 n_3 (1 - \cos \theta) + n_2 \sin \theta\\
n_1 n_2 (1 - \cos \theta) + n_3 \sin \theta & \cos \theta + n_2^2 (1 -
\cos \theta) & n_2 n_3 (1 - \cos \theta) - n_1 \sin \theta \\
n_1 n_3 (1 - \cos \theta) - n_2 \sin \theta & n_2 n_3 (1 - \cos 
\theta) + n_1 \sin \theta & \cos \theta + n_3^2 (1 - \cos \theta)
\end{pmatrix}
\end{eqnarray}

It is also useful to derive the following representation of rotation 
matrix:
\begin{eqnarray}
\label{eq:rotation_factorization}
\mathbf{R}(\mathbf{n}, \theta) = P \mathbf{R}(z, \theta) P^{-1}
\end{eqnarray}

\noindent where $P$ is an orthogonal matrix, i.e. $P^{-1} = P^\top$, 
and $\mathbf{R}(z, \theta)$ is a standard rotation matrix around $z$-
axis about angle $\theta$:

\begin{equation}
\label{eq:axis.angle.PR}
P = 
\begin{pmatrix}
\frac{n_3 n_1}{\sqrt{n_1^2 + n_2^2}} & \frac{-n_2}{\sqrt{n_1^2 + 
n_2^2}} & n_1\\\\
\frac{n_3 n_2}{\sqrt{n_1^2 + n_2^2}} & \frac{n_1}{\sqrt{n_1^2 + 
n_2^2}} & n_2 \\\\
-\sqrt{n_1^2 + n_2^2} 			 & 0	 & n_3
\end{pmatrix},
\quad
\mathbf{R}(z, \theta) = 
\begin{pmatrix}
\cos \theta & -\sin \theta & 0 \\
\sin \theta & \cos \theta & 0 \\
0 & 0 & 1 
\end{pmatrix}
\end{equation}

\Paragraph{Axis-angle to Quaternion}
The exponential map can be also directly mapped as a unit quaternion $
(q_r ~ q_x ~ q_y ~ q_z)^\top$ as follows \cite{grassia1998practical}:

\begin{subeqnarray}
\label{eq:exp.map.so3.quat}
	\exp: \mathfrak{so}(3) & \mapsto & \mathbf{SO}(3) \\
	\W & \mapsto & \mathbf{SU}(2) \\
	~ \nonumber \\
	e_q^ { \W } &=& \left\{ 
	\begin{array}{ll}
		(1,0,0,0)^\top  & \text{ , if }  \W = (0,0,0)^\top \\
		\left( \cos\dfrac{|\W|}{2}, \dfrac{\sin\dfrac{|\W|}{2}}{|\W|} ~ \W \right)^\top &  \text{ , otherwise}
	\end{array}
	\right.
\end{subeqnarray}

\subsubsection{Logarithm map}
\label{sect:log_map_so3}

\Paragraph{Matrix to Axis-angle}

The map:

\begin{subeqnarray}
  \log: \mathbf{SO}(3) & \mapsto & \mathfrak{so}(3)   \\
           \mathbf{R}_{3\times 3} & \mapsto &  \W
\end{subeqnarray}

\noindent is well-defined for rotation angles $\theta \in (0, \pi)$, 
surjective, and is the inverse of the $\exp$ function defined above. 
From Rodrigues' formula (eq. \ref{eq:rodrigues_coordinate}) or from 
rotation matrix factorization and trace properties (eq. 
\ref{eq:rotation_factorization}) it follows that:
\begin{eqnarray}
\label{eq:rotation_angle}
\cos \theta &=& \frac{1}{2}(tr(\mathbf{R})-1)  \\
\sin \theta &=& (1 - \cos^2 \theta)^{1/2} = \frac{1}{2}\sqrt{(3 - 
tr(\mathbf{R}))(1 + tr(\mathbf{R}))} \nonumber
\end{eqnarray}

\noindent where $\sin \theta \geq 0$ is a consequence of the 
convention for the range of the rotation angle, $\theta \in [0, \pi]$.

If $\sin \theta \neq 0$, i.e. $\theta \neq \{0, \pi\}$, from eq. 
\ref{eq:rodrigues_coordinate}:
\begin{eqnarray}
\label{eq:rodrigues_ln}
\log(\mathbf{R}) &=& \frac{\theta}{2\sin \theta} \left( \mathbf{R} - 
\mathbf{R}^\top
\right), \quad tr(\mathbf{R}) \neq \{-1, 3\} \nonumber \\
\label{eq:log_so3}
 \W = \left[ \log(\mathbf{R}) \right]^\vee &=& \frac{\theta}{2 \sin 
 \theta} (R_{32} - R_{23}, R_{13} - R_{31}, R_{21} - R_{12})^\top 
\end{eqnarray}

If $\sin \theta = 0$, then $\theta = 0$ or $\theta = \pi$. In both 
cases (from eq. \ref{eq:rodrigues_coordinate}) $R_{ij} = R_{ji}$, and 
$\W$ can not be determined by eq. \ref{eq:rodrigues_ln}. However, the 
angle $\theta$ is derived from eq. \ref{eq:rotation_angle}, and 
inserting it to the Rodrigues' formula \ref{eq:rodrigues}:

\begin{eqnarray}
\label{eq:rodrigues_ln_special}
\W \text{ is undetermined} \hspace{0.2\textwidth} \text{ if } \theta 
&=& 0, \nonumber \\
\frac{\W}{|\W|} = \mathbf{n} = \left(\epsilon_1 \sqrt{\frac{1}{2}(1 + 
R_{11})}, \epsilon_2 \sqrt{\frac{1}{2}(1 + R_{22})}, \epsilon_3 
\sqrt{\frac{1}{2}(1 + R_{33})} \right)^\top, \quad \text{if } \theta 
&=& \pi
\end{eqnarray}

\noindent where the individual signs $\epsilon_i = \pm 1$ (if $n_i 
\neq 0$) are determined up to an overall sign (since $\mathbf{R}
(\mathbf{n}, \pi) = \mathbf{R}(\mathbf{n}, -\pi)$) via the following 
relation:

\begin{equation}
\label{eq:rodrigues_ln_special_sign}
\epsilon_i \epsilon_j = \frac{R_{ij}}{\sqrt{(1 + R_{ii})(1 + 
R_{jj})}}, \text{ for } i \neq j, R_{ii} \neq -1, R_{jj} \neq -1 
\end{equation}

There is an alternative approach for the case $\theta = \pi$, which 
determines the axis of rotation $\mathbf{n}$ for the angles $\theta 
\approx \pi$ without numerical issues. We define matrix:

\begin{equation}
\label{eq:matrix_for_axis_around_pi}
S \equiv \mathbf{R} + \mathbf{R}^\top + (1 - tr\mathbf{R}) 
\mathbf{I_3}
\end{equation}

\noindent Then the Rodrigues' equation in coordinate form (eq. 
\ref{eq:rodrigues}) yields:

\begin{equation}
\label{eq:axis_around_pi}
n_j n_k = \frac{S_{jk}}{3 - tr(\mathbf{R})}, \qquad tr(\mathbf{R}) \neq 3
\end{equation}

\noindent To determine $\mathbf{n}$ up to an overall sign, we simply 
set $j = k$ in eq. (\ref{eq:axis_around_pi}), which fixes the value of 
$n_j^2$. If $\sin \theta \neq 0$, the overall sign of $\mathbf{n}$ is 
determined by eq. (\ref{eq:rodrigues_ln}). If $\sin \theta = 0$ then 
there are two cases. For $\theta = 0$ (corresponding to the identity 
rotation), $S = 0$ and the rotation axis $\mathbf{n}$ is undefined. 
For $\theta = \pi$, the ambiguity in the overall sign of $\mathbf{n}$ 
is immaterial, since $\mathbf{R}(\mathbf{n}, \pi) = \mathbf{R}
(\mathbf{n}, -\pi)$.

\Paragraph{Quaternion to Axis-angle}

The logarithm map can be also directly given from a unit quaternion $\mathbf{q} = (q_r ~ q_x ~ q_y ~ q_z)^\top = (q_r, \mathbf{q}_v^\top)^\top$ as follows \cite{grassia1998practical}:

\begin{subeqnarray}
	\log: \mathbf{SO}(3) & \mapsto & \mathfrak{so}(3)   \\
	\mathbf{SU}(2) & \mapsto &  \W \\
    \W  &=& \dfrac{2 \arccos(q_r)}{|\mathbf{q}_v|} ~ \mathbf{q}_v
\end{subeqnarray}

\subsection{For $\mathbf{SE}(3)$}
\label{eq:exp.log.se3}

\subsubsection{Exponential map}
\label{sect:se3_exp}

Let

\begin{equation}
\label{eq:vector_in_se3}
\mathbf{v}= \left( \begin{array}{c} \mathbf{t} \\ \W \end{array} \right)
\end{equation}

\noindent denote the 6-vector of coordinates in
the Lie algebra $\mathfrak{se}(3)$,
comprising
two separate 3-vectors: $\W$, the vector that determine
the rotation, and $\mathbf{t}$ which determines the translation.
Furthermore, we define the $4 \times 4$ matrix:

\begin{equation}
 \mathbf{A}(\mathbf{v})=
\left(
\begin{array}{cc}
 \hatop{\W}  & \mathbf{t} \\
 0 & 0
\end{array}
\right)
\end{equation}

Then, the map:

\begin{equation}
  \exp: \mathfrak{se}(3) \mapsto \mathbf{SE}(3)
\end{equation}

\noindent is well-defined, surjective, and has the closed form:

\begin{eqnarray}
\label{eq:se3.exp.map}
  e^ { \mathbf{v} } \equiv  e^ { \mathbf{A}(\mathbf{v}) } =
\left(
\begin{array}{cc}
  e^{\hatop{\W}} & \mathbf{V} \mathbf{t} \\
   0 & 1
\end{array}
\right)
\\
\label{eq:V_exp}
\mathbf{V} = \mathbf{I_3}
+ \frac{1-\cos \theta}{\theta^2} \hatop{\W}
+ \frac{\theta- \sin \theta}{\theta^3} (\hatop{\W})^2 
\end{eqnarray}

\noindent with $\theta = |\W|$ and $e^{\hatop{\W}}$
defined in Eq.(\ref{eq:rodrigues})
and $\hatop{\W}$ using the hat operator introduced in \S\ref{sect:mat_deriv:ops}.

\subsubsection{Logarithm map}

The map:

\begin{eqnarray}
  \log: \mathbf{SE}(3) &\mapsto& \mathfrak{se}(3) \\
  \mathbf{A}(\mathbf{v})  & \mapsto & \mathbf{v}
\nonumber
\end{eqnarray}

\noindent is well-defined and can be computed as \cite{wang2008nps}:

\begin{eqnarray}
\label{eq:rodrigues_ln_se3}
\mathbf{v} &=&
\left(
\begin{array}{c}
 \mathbf{t}' \\ \hline \W
\end{array}
\right)
=
\left(
\begin{array}{c}
 x'\\y'\\z' \\ \hline \omega_1 \\ \omega_2 \\ \omega_3
\end{array}
\right)
\nonumber \\
\W &=& \left[ \log \mathbf{R} \right]^\vee  \quad \quad \text{(see Eq.~\ref{eq:log_so3})}
\\
\mathbf{t}' &=& \mathbf{V}^{-1} \mathbf{t} \quad \quad \text{(with $\mathbf{V}$ in Eq.~\ref{eq:V_exp})}
\end{eqnarray}

\noindent where $\mathbf{R}$ and $\mathbf{t}$ are the $3 \times 3$ rotation matrix
and translational part of the $\mathbf{SE}(3)$ pose.
Note that $\mathbf{V}^{-1}$ has a closed-form expression \cite{gallier2003computing}:

\begin{equation}
\mathbf{V}^{-1} = \mathbf{I}_3 
- \dfrac{1}{2} \hatop{\W} + 
\dfrac{\left(
 1 - \dfrac{ \theta \cos(\theta/2)}{2 \sin(\theta / 2)}
\right) 
}{\theta^2}  (\hatop{\W})^2
\end{equation}

\subsubsection{Pseudo-exponential map}
\label{sect:se3_pseudo-exp}

Let $\mathbf{v}$ be a 6-vector of coordinates in
the Lie algebra $\mathfrak{se}(3)$, per Eq.~\ref{eq:vector_in_se3}, 
comprising a vector that determines the rotation ($\W$)
and another one for the translation ($\mathbf{t}$).

We can define the ``pseudo-exponential'' of $\mathbf{v}$
by leaving the translation part intact, and evaluating the 
matrix exponential for the SO(3) part only, that is:

\begin{equation}
\label{eq:pseudo-exp-se3}
\text{pseudo-exp}(\mathbf{v}) = 
\left(
\begin{array}{cc}
  e^{\hatop{\W}} & \mathbf{t} \\
   0_{1 \times 3} & 1
\end{array}
\right)
\end{equation}

The interest in this modified version of the exponential map 
is that it leads to Jacobians that are more efficient to evaluate
than those of the real matrix exponential.
Note that this defines a valid retraction on SE(3), as long as the corresponding ``pseudo-logarithm'' 
is also used to map SE(3) poses to local tangent space coordinates.

Compare Eq.~\ref{eq:pseudo-exp-se3} to Eq.~\ref{eq:se3.exp.map} to see why this leads to simpler Jacobians.

\subsubsection{Pseudo-logarithm map}
\label{sect:se3_pseudo-log}

Given a SE(3) pose $\mathbf{T}$: 

\begin{equation}
\mathbf{T} =
\left(
\begin{array}{c|c}
 \mathbf{R}_{3 \times 3}  & \mathbf{d_{t}}_{3 \times 1}  \\
\hline
  0 ~ 0 ~ 0 & 1
\end{array}
\right)
\end{equation}

\noindent we can compute the ``pseudo-logarithm'' of $\mathbf{T}$
by taking the regular matrix logarithm to the rotational part ($3 \times 3$), and leaving the translation vector intact, that is: 

\begin{equation}
\label{eq:pseudo-log-se3}
\left. \text{pseudo-log}(\mathbf{T})^\vee \right|_{6 \times 1} = 
\left(
\begin{array}{c}
\mathbf{d_{t}}_{3 \times 1} \\
\log(\mathbf{R})^\vee
\end{array}
\right)
\end{equation}

\subsection{Implementation in MRPT}

The class \texttt{mrpt::poses::CPose3D} implements both the exponential and logarithm
maps for both $\mathbf{SO}(3)$ and $\mathbf{SE}(3)$
up to MRPT version 1.5.x. Since MRPT 2.0, the pseudo-exponential and pseudo-logarithm 
maps are available in the namespace 
\texttt{mrpt::poses::Lie::SE<n>}, with n=2 or 3:

\begin{lstlisting}
#include <mrpt/poses/CPose3D.h>
#include <mrpt/poses/CPose2D.h>
#include <mrpt/poses/Lie/SE.h>

mrpt::poses::Lie::SE<3>::tangent_vector v;
//...
mrpt::poses::CPose3D p = mrpt::poses::Lie::SE<3>::exp(v);

mrpt::poses::CPose3D p;
//...
mrpt::poses::Lie::SE<3>::tangent_vector v = mrpt::poses::Lie::SE<3>::log(p);
\end{lstlisting}

\chapter{Optimization problems on $\mathbf{SE}(3)$}
\label{ch:se3_optim}

Now that the main concepts needed to handle $\mathbf{SE}(3)$
as a manifold have been established in
the previous chapters \S\S\ref{chap:mat_deriv}--\ref{chap:se3_lie},
in this chapter we focus on the ultimate goal of all that theoretical
dissertation: being able to solve practical numerical
problems that involve
estimating $\mathbf{SE}(3)$ poses.
It is noteworthy that this problem was already addressed back in 1982 in \cite{gabay1982mdf} for the general case of differential manifolds.

\section{Optimization solutions are made for flat Euclidean spaces}

Gradient descent, Gauss-Newton, Levenberg-Marquart and the family of Kalman
filters are all
invaluable methods which, at their core, perform exactly the same operation:
iteratively improving a state vector $\mathbf{x}$ so that it minimizes
a sum of square errors between some prediction and a vector of observed data $\mathbf{z}$
\footnote{In fact, the widely used Extended Kalman filter does not iterate,
but it can be seen as doing just one Gauss-Newton iteration \cite{bell1993ikf}.}.

It does not matter for our purposes which method is employed to solve a problem.
All the relevant information is that, at some stage of the optimization
it is used a prediction (or system model) function
$\mathbf{f}(\mathbf{x})$.
The goal is always to minimize the squared error from
this prediction to the observation, that is,
to minimize:

\begin{equation}
S(\mathbf{x}) = (\mathbf{f}(\mathbf{x})-\mathbf{z})^\top (\mathbf{f}(\mathbf{x})-\mathbf{z})
 = | \mathbf{f}(\mathbf{x})-\mathbf{z} |^2
\end{equation}

To achieve this, $\mathbf{x}$ is updated iteratively by means of small increments:

\begin{equation}
\label{eq:optim1}
 \mathbf{x}  \leftarrow \mathbf{x} + \DEL
\end{equation}

Increments $\DEL$ are obtained
(in all the methods mentioned above)
by solving the equation:

\begin{equation}
\label{eq:optim2}
\left. \frac{\partial S(\mathbf{x} + \DEL )}{\partial \DEL }
\right|_{\DEL=0}
= 0
\end{equation}

\noindent since a null derivative means a minimum in the error function $S(\cdot)$.
Notice how the Jacobian is evaluated at $\DEL=0$, that is, at the vicinity of
the present estimation $\mathbf{x}$.
Typically, the steps Eq.(\ref{eq:optim2}) and Eq.(\ref{eq:optim1}) are
iterated until convergence or for a fixed number of iterations.

At this point, it must be raised the problem of employing any of
these methods when $\mathbf{SE}(3)$ poses
are part of the state vector $\mathbf{x}$ being estimated:
all these optimization methods are designed to work on flat Euclidean spaces, i.e. on $\mathbb{R}^N$.
If we wanted to optimize a state vector that contains (one or more) poses,
we would have to store it, as a vector, in one of the parameterizations explained in this report, namely:

\begin{enumerate}
 \item A 3D+YPR -- each pose comprises 6 elements in $\mathbf{x}$.
 \item A 3D+Quat -- each pose comprises 7 elements in $\mathbf{x}$.
 \item A full $4\times 4$ matrix -- each pose comprises 16 elements in $\mathbf{x}$.
 \item The top $3\times 4$ submatrix -- each pose comprises 12 elements in $\mathbf{x}$.
\end{enumerate}

None of them are an ideal solution, and some are a really bad idea:

\begin{enumerate}
 \item The first case achieves minimum storage requirements (6 elements for a 6D pose),
but there might not exist closed-form Jacobians for all possible pose-pose chained operations,
and also the update rule $\mathbf{x}  \leftarrow \mathbf{x} + \DEL$ means that the three
angles may go out of their valid ranges (need to renormalize the state vector after each update).
Furthermore, there exists the problem of gimbal lock (\S \ref{sect:ypr:gimbal}) where one
DOF is lost. When there are free DOFs, an optimization method may try to move along the
degenerated set of solutions and get stuck.

\item In the second case, Jacobians are always well-defined, but there is one extra DOF, which has
the above-mentioned problems.

\item In the third and fourth cases, Jacobians are always well-defined but there are
even more extra DOFs, making the problem even worse. The storage requirements are also an
important drawback.
\end{enumerate}

To sum up: storing poses in a state vector and trying to optimize
them is not a good idea.
In spite of the fact that
the 3D+Quat parameterization is bad to a lesser degree,
still being usable
(in fact, it led to good results in computer vision \cite{davison2007mrt}),
a more robust and general approach is described in the next section.

\section{An elegant solution: to optimize on the manifold}
\label{sect:opti_manif}

Although the idea is not new at all (see \cite{gabay1982mdf}),
carrying out optimization directly on the manifold while
keeping a 3D-YPR or 3D-Quat parameterization in the
state vector is a solution which is gaining popularity
in the robotics and computer vision community in recent years
(e.g. \cite{hertzberg2008fsm,strasdat2010scale}).

Following the notation of \cite{hertzberg2008fsm,hertzberg2013integrating}, the only changes required
to the optimization method are
to replace the expressions on the left column by their counterparts on the right (the so-called ``boxplus'' notation $\boxplus$):

\begin{eqnarray}
 \DEL^\star \leftarrow
\left. \frac{\partial S(\mathbf{x} + \DEL )}{\partial \DEL }
\right|_{\DEL=0}  = 0
 &
\quad \quad \Longrightarrow \quad \quad
&
 \E^\star \leftarrow
\left. \frac{\partial S(\mathbf{x} \boxplus {\E} )}{\partial \E }
\right|_{\E=0}  = 0
\\
 \mathbf{x} \leftarrow \mathbf{x} + \DEL^\star
 &
\quad \quad \Longrightarrow \quad \quad
&
 \mathbf{x} \leftarrow \mathbf{x} \boxplus \E^\star
\end{eqnarray}

\noindent where $\mathbf{x} \in M$ is the state vector of the problem, which lies
on some $N$-dimensional manifold $M$ (a Lie group, actually),
$\E \in \mathbb{R}^N$ is the increment in the linearization of the manifold
around $\mathbf{x}$ (using $M$'s Lie algebra as a vector base),
and the ``boxplus'' operator $\boxplus: M \times \mathbb{R}^N \mapsto M$
is a generalization of the normal addition operator $+$ for Euclidean spaces. 

There are two possible ways to implement $\boxplus$, both of them perfectly valid:
Let $\mathbf{x}, \mathbf{x}' \in M$ be elements of the manifold of the problem $M$,
and $\E \in \mathbb{R}^N$ an increment in its linearized approximation.
Then:

\begin{eqnarray}
 \mathbf{x}' = \mathbf{x} \boxplus \E  & \Longrightarrow & \mathbf{x}' = \mathbf{x} e^\E 
\end{eqnarray}

\noindent $\mathbf{x} e^\E$ being the ``product'' as defined by
the manifold group operation, and $e^\E$ being the exponential map of the Lie group $M$
(\S\ref{chap:se3_lie}).
It is important to highlight that the topological structure of
$\mathbf{x}$ may be the product of many elemental topological substructures
(e.g. storing two 3D points and three $\mathbf{SE}(3)$ poses would give a
$\mathbb{R}^3 \times \mathbb{R}^3 \times \mathbf{SE}(3) \times \mathbf{SE}(3) \times \mathbf{SE}(3)$
structure). Therefore, if the estimated vector contains parts in Euclidean space, the
group product falls back to common addition
(as it would be in the original optimization method).

In GraphSLAM problems, the ``boxminus'' operator ($\boxminus$) is also required:

\begin{eqnarray}
\label{eq:boxminus}
 \mathbf{y} \boxminus \mathbf{x} = \log( \mathbf{x}^{-1} \mathbf{y} )
\end{eqnarray}

\section{Useful manifold derivatives}
\label{chap:se3_lie:deriv}

Below follow some Jacobians that usually appear in optimization problems
when using the on-manifold optimization approach described in the previous section.
The formulas below plus the chain rule of Jacobians will be probably enough
to obtain ready-to-use expressions for a large number of optimization problems
in robotics and computer vision.

Before reading this section, make sure of taking a look at the notation conventions
for matrix derivatives explained in \S\ref{chap:mat_deriv}
(e.g. where does the dimensionality of 12 comes from?).

\subsection{Jacobian of the SE(3) exponential generator}
\label{sect:jacob_se3_gen}

This is the most basic Jacobian, since the term $e^\E$ appears in all
the on-manifold optimization problems.
Note that the derivative is taken at $\E=0$ for the reasons explained
in the previous section.
These are ones of the few genuinely new Jacobians in this chapter.
Most of what follows then is obtained by combining several Jacobians,
as those in \S\ref{chap:mat_deriv}, via the chain rule.

\subsubsection{SO(3) in matrix form}

Taking derivatives of the exponential map (see Eq.(\ref{eq:rodrigues})) at the Lie algebra coordinates $\E=0$ we obtain:

\begin{equation}
\left. \frac{\partial e^\W}{\partial \W} \right|_{\W=0}
\equiv 
\left. \frac{\partial vec(e^\W)}{\partial \W} \right|_{\W=0}
=
\left(
\begin{array}{c}
 -\hatop{\mathbf{e}}_1 \\
-\hatop{\mathbf{e}}_2 \\
-\hatop{\mathbf{e}}_3
\end{array}
\right)
\quad\quad\quad \text{(A $9 \times 3$ Jacobian)}
\end{equation}

\noindent with $\mathbf{e_1}=[1 ~ 0 ~ 0]^\top$,
$\mathbf{e_2}=[0 ~ 1 ~ 0]^\top$ and
$\mathbf{e_3}=[0 ~ 0 ~ 1]^\top$. The dimensionality ``9'' comes from the 
vector-stacked view (the $vec(\cdot)$ operator) of the rotation matrix. 

\subsubsection{SO(3) in quaternion form}

We need to take derivatives of the exponential map in quaternion form in Eq.(\ref{eq:exp.map.so3.quat}). 
For convenience, we will express $e_q^\W(\W)$ in Eq.(\ref{eq:exp.map.so3.quat}) as a function of 
$\W$ and $\theta=|\W|$ such that $e_q^\W(\W,\theta)$, which will result in simpler (factorized) Jacobian 
expression than that of direct approach:

\begin{subeqnarray}
\left. \frac{\partial e_q^\W}{\partial \W} \right|_{\W=0}
&=& 
\frac{\partial e_q^\W(\W,\theta)}{\partial \{ \omega_x,\omega_y,\omega_z, \theta \} }
\frac{\partial \{ \omega_x,\omega_y,\omega_z, \theta \} }{\partial \{ \omega_x,\omega_y,\omega_z \} }
\quad\quad\quad \text{(A $4 \times 3$ Jacobian)}
\\
&=& 
\left(\begin{array}{cccc} 
	0 & 0 & 0 & -\frac{\sin\left(\frac{|\W|}{2}\right)}{2}
	\\ 
	\frac{\sin\left(\frac{|\W|}{2}\right)}{|\W|} & 0 & 0 &  \omega_x \left( \frac{\cos\left(\frac{|\W|}{2}\right)}{2\,|\W|}-\frac{\sin\left(\frac{|\W|}{2}\right)}{{|\W|}^2}\right)
	\\ 
	0 & \frac{\sin\left(\frac{|\W|}{2}\right)}{|\W|} & 0 & \omega_y \left( \frac{\cos\left(\frac{|\W|}{2}\right)}{2\,|\W|}-\frac{\sin\left(\frac{|\W|}{2}\right)}{{|\W|}^2} \right)
	\\
	0 & 0 & \frac{\sin\left(\frac{|\W|}{2}\right)}{|\W|} & \omega_z \left( \frac{\cos\left(\frac{|\W|}{2}\right)}{2\,|\W|}-\frac{\sin\left(\frac{|\W|}{2}\right)}{{|\W|}^2} \right)
\end{array}\right)_{(4\times 4)}
\left(\begin{array}{ccc} 
	& & \\
	& \mathbf{I}_3 & \\
	& & \\
	\hline	
	\frac{\omega_x}{|\W|} & \frac{\omega_y}{|\W|} & \frac{\omega_z}{|\W|} \end{array}\right)
_{(4\times 3)}
\end{subeqnarray}

In the Sophus C++ library \cite{sophus}, this Jacobian is available as the method \texttt{SO3::Dx\_exp\_x(omega)}, 
with a slight variable reordering, i.e. in Sophus, quaternions are stored as $(q_x,q_y,q_z,q_r)$ instead of $(q_r,q_x,q_y,q_z)$.

\subsubsection{SE(3) in matrix form}

Taking derivatives of the exponential map (see Eq.(\ref{sect:se3_exp})) at the Lie algebra coordinates $\E=0$ we obtain:

\begin{equation}
\left. \frac{\partial e^\E}{\partial \E} \right|_{\E=0}
\equiv
\left. \frac{\partial vec(e^\E)}{\partial \E} \right|_{\E=0}
=
\left(
\begin{array}{cc}
 \mathbf{0}_{3\times 3}  & -\hatop{\mathbf{e}}_1 \\
 \mathbf{0}_{3\times 3}  & -\hatop{\mathbf{e}}_2 \\
 \mathbf{0}_{3\times 3}  & -\hatop{\mathbf{e}}_3 \\
 \mathbf{I}_{3}  & \mathbf{0}_{3\times 3} \\
\end{array}
\right)
\quad\quad\quad \text{(A $12 \times 6$ Jacobian)}
\end{equation}

\noindent with $\mathbf{e_1}=[1 ~ 0 ~ 0]^\top$,
$\mathbf{e_2}=[0 ~ 1 ~ 0]^\top$ and
$\mathbf{e_3}=[0 ~ 0 ~ 1]^\top$.
Notice that the resulting Jacobian is for the
ordering convention of $\mathfrak{se}(3)$ coordinates shown in Eq.(\ref{eq:vector_in_se3}), i.e. $\E = (d_x, d_y, d_z, \W^\top)^\top$.


\subsection{Jacobian of the SO(3) logarithm}
\label{sect:eq:jacob_dLnROT_dROT}

This Jacobian will end up appearing wherever we take derivatives of a function which at some point takes
as argument a rotation matrix ($3\times 3$) and computes the vee operator of its logarithm map \S\ref{sect:log_map_so3},
e.g. while optimizing pose graphs in Graph-SLAM with the 
``boxminus'' operator (see Eq.~\ref{eq:boxminus}).

Given an input rotation matrix $\mathbf{R}$:

\begin{equation}
\mathbf{R} = \left(
\begin{array}{ccc}
 R_{11} & R_{12} & R_{13} \\
 R_{21} & R_{22} & R_{23} \\
 R_{31} & R_{32} & R_{33} \\
\end{array}
\right)
\nonumber
\end{equation}

\noindent it can be shown that:

\begin{eqnarray}
\label{eq:dLnRot_wrt_R}
\left.
 \frac{\partial \log(\mathbf{R})^\vee }{\partial \mathbf{R}}
\right|_{3\times 9}
 =
\left\{
\begin{array}{ll}
  \left(
  \begin{array}{ccc|ccc|ccc}
   0  & 0 &  0  & 0 & 0 & \frac{1}{2} & 0 & -\frac{1}{2} & 0   \\
   0  & 0 &  -\frac{1}{2}  & 0 & 0 & 0 & \frac{1}{2} & 0 & 0   \\
   0  & \frac{1}{2} &  0 & -\frac{1}{2} & 0 & 0 & 0 & 0 & 0
  \end{array}
  \right)
 &
\text{, if $\cos\theta > 0.999999...$} \\
~\\
  \left(
  \begin{array}{ccc|ccc|ccc}
   a_1  & 0 &  0  & 0 & a_1 & b & 0 & -b & a_1   \\
   a_2  & 0 & -b  & 0 & a_2 & 0 & b & 0 & a_2   \\
   a_3  & b &  0 & -b & a_3 & 0 & 0 & 0 & a_3
  \end{array}
  \right)
 &
\text{, otherwise}
\end{array}
\right.
\end{eqnarray}

\noindent where the order of the 9 components is assumed to be column-major ($R_{11},R_{21},...$) and:

\begin{eqnarray}
\cos \theta &=& \frac{tr(\mathbf{R})-1}{2}  \nonumber \\
\sin \theta &=& \sqrt{1-\cos^2 \theta}  \nonumber \\
\left[
\begin{array}{c}
 a_1 \\ a_2 \\ a_3
\end{array}
\right]
 &=& \left[ \mathbf{R} - \mathbf{R}^\top \right]^\vee
\frac{\theta \cos\theta  -\sin \theta  }{4 \sin^3 \theta}
 = \left[
\begin{array}{c}
 R_{32} - R_{23} \\
 R_{13} - R_{31} \\
 R_{21} - R_{12} \\
\end{array}
 \right]
\frac{\theta \cos\theta  -\sin \theta  }{4 \sin^3 \theta}
\nonumber \\
b &=& \frac{\theta}{2 \sin \theta}
\nonumber
\end{eqnarray}

\subsection{Jacobian of $D\boxplus \varepsilon = e^\varepsilon \oplus D$ (left-multiply option)}
\label{sect:jacob_eD}

Let $\mathbf{D} \in \mathbf{SE}(3)$ be a pose with associated transformation matrix:

\begin{equation}
\mathbf{T}(\mathbf{D}) =
\left(
\begin{array}{ccc|c}
 \mathbf{d_{c1}}  & \mathbf{d_{c2}}  & \mathbf{d_{c3}}  & \mathbf{d_{t}}  \\
\hline
  0 & 0 & 0 & 1
\end{array}
\right)
\end{equation}

Following the convention of left-composition for the infinitesimal
pose $e^\E$ described in \S\ref{sect:opti_manif},
we are interested in the derivative of $e^\E \oplus \mathbf{D}$ w.r.t $\E$:

\begin{eqnarray}
\left. \frac{\partial e^\E \mathbf{D}}{\partial \E} \right|_{\E = 0}
&=&
\left. \frac{\partial \mathbf{A} \mathbf{D}}{\partial \mathbf{A} } \right|_{\mathbf{A}=\mathbf{I_4} = e^\E}
\left. \frac{\partial e^\E}{\partial \E} \right|_{\E=0}
\\
\label{eq:jacob_eD.2}
&=&
\left[ \mathbf{T}(\mathbf{D})^\top \otimes  \mathbf{I}_3 \right]
\left. \frac{\partial e^\E}{\partial \E} \right|_{\E=0}
\\
&=&
\left(
\begin{array}{cc}
 \mathbf{0}_{3\times 3}  & -\hatop{\mathbf{d}}_{c1} \\
 \mathbf{0}_{3\times 3}  & -\hatop{\mathbf{d}}_{c2} \\
 \mathbf{0}_{3\times 3}  & -\hatop{\mathbf{d}}_{c3} \\
 \mathbf{I}_{3}  & -\hatop{\mathbf{d}}_{t} \\
\end{array}
\right)
\quad\quad\quad \text{(A $12 \times 6$ Jacobian)}
\end{eqnarray}

Note: This Jacobian is implemented in MRPT in \texttt{CPose3D::jacob\_dexpeD\_de()}.

\subsection{Jacobian of $D\boxplus \varepsilon = D \oplus e^\varepsilon$  (right-multiply option)}
\label{sect:jacob_De}

Let $\mathbf{D} \in \mathbf{SE}(3)$ be a pose with associated transformation matrix:

\begin{equation}
\mathbf{T}(\mathbf{D}) =
\left(
\begin{array}{ccc|c}
 \mathbf{d_{c1}}  & \mathbf{d_{c2}}  & \mathbf{d_{c3}}  & \mathbf{d_{t}}  \\
\hline
  0 & 0 & 0 & 1
\end{array}
\right)
=
\left(
\begin{array}{c|c}
 \mathbf{R}(D) & \mathbf{d_{t}} \\
\hline
     0 & 1
\end{array}
\right)
\end{equation}

We are here interested in the derivative of $\mathbf{D} \oplus e^\E$ w.r.t $\E$,
which can be obtained from the results of \S\ref{sect:jacob_pose_pose_comp} and \S\ref{sect:jacob_se3_gen}):

\begin{eqnarray}
\left. \frac{\partial \mathbf{D e^\E}}{\partial \E} \right|_{\E = 0}
&=&
\left. \frac{\partial \mathbf{A} \mathbf{B}}{\partial \mathbf{B} } \right|_{\mathbf{A}=\mathbf{D}, \mathbf{B}=\mathbf{I_4}}
\left. \frac{\partial e^\E}{\partial \E} \right|_{\E=0}
\\
&=&
\left[ \mathbf{I}_4 \otimes \mathbf{R}(\mathbf{D}) \right]
\left(
\begin{array}{cc}
 \mathbf{0}_{3\times 3}  & -\hatop{\mathbf{e}}_1 \\
 \mathbf{0}_{3\times 3}  & -\hatop{\mathbf{e}}_2 \\
 \mathbf{0}_{3\times 3}  & -\hatop{\mathbf{e}}_3 \\
 \mathbf{I}_{3}  & \mathbf{0}_{3\times 3} \\
\end{array}
\right)
\\
\label{eq:jacob.d_eps_wrt_eps}
&=&
\left(
\begin{array}{c|ccc}
  ~  &  0_{3\times 1}  & -\mathbf{d_{c3}} & \mathbf{d_{c2}} \\
  0_{9\times 3}
     & \mathbf{d_{c3}} & 0_{3\times 1} & -\mathbf{d_{c1}} \\
  ~  &  -\mathbf{d_{c2}} & \mathbf{d_{c1}} & 0_{3\times 1} \\
\hline
  \mathbf{R}(\mathbf{D})   & ~ & 0_{3 \times 3} &
\end{array}
\right)
\quad\quad \text{(A $12 \times 6$ Jacobian)}
\end{eqnarray}

Note: This Jacobian is implemented in MRPT in \texttt{CPose3D::jacob\_dDexpe\_de()}.

\subsection{Jacobian of $e^\varepsilon \oplus D \oplus p$}
\label{sect:jacob_eDp}

This is the composition of a pose $\mathbf{D}$ with a point $\mathbf{p}$,
an operation needed, for example, in Bundle Adjustment implementations \cite{triggs2000bundle}
(with the convention of points relative to the camera being $D \oplus p$, that is,
$D$ being the inverse of the actual camera position).

Let $\mathbf{p} \in \mathbb{R}^3$ be a 3D point, and
$\mathbf{D} \in \mathbf{SE}(3)$ be a pose with associated transformation matrix:

\begin{equation}
\mathbf{T}(\mathbf{D}) =
\left(
\begin{array}{ccc|c}
 d_{11} & d_{12} & d_{13} & d_{tx}   \\
 d_{21} & d_{22} & d_{23} & d_{ty}   \\
 d_{31} & d_{32} & d_{33} & d_{tz}   \\
\hline
  0 & 0 & 0 & 1
\end{array}
\right)
=
\left(
\begin{array}{ccc|c}
 \mathbf{d_{c1}}  & \mathbf{d_{c2}}  & \mathbf{d_{c3}}  & \mathbf{d_{t}}  \\
\hline
  0 & 0 & 0 & 1
\end{array}
\right)
=
\left(
\begin{array}{ccc|c}
   & \mathbf{R_D}  &  & \mathbf{d_{t}}  \\
\hline
  0 & 0 & 0 & 1
\end{array}
\right)
\end{equation}

We are interested in the derivative of $e^\E \oplus \mathbf{D} \oplus p$ w.r.t $\E$:

\begin{eqnarray}
\left. \frac{\partial (e^\E \mathbf{D}) \oplus \mathbf{p} }{\partial \E} \right|_{\E = 0}
&=&
\left. \frac{\partial \mathbf{A} \oplus \mathbf{p}}{\partial \mathbf{A} } \right|_{\mathbf{A}=e^\E \mathbf{D} = \mathbf{D}}
\left. \frac{\partial e^\E \mathbf{D}}{\partial \E} \right|_{\E=0}
\\
\text{\scriptsize{(Using Eq.(\ref{eq:jac_dAp_A}) \& \S\ref{sect:jacob_eD}} )} &=&
\left(\left( \mathbf{p}^\top ~~ 1 \right) \otimes \mathbf{I_3} \right)
\left(
\begin{array}{cc}
 \mathbf{0}_{3\times 3}  & -\hatop{\mathbf{d}}_{c1} \\
 \mathbf{0}_{3\times 3}  & -\hatop{\mathbf{d}}_{c2} \\
 \mathbf{0}_{3\times 3}  & -\hatop{\mathbf{d}}_{c3} \\
 \mathbf{I}_{3}  & -\hatop{\mathbf{d}}_{t} \\
\end{array}
\right)
\\
\label{eq:jacob_eDp_e}
&=&
\left(
\begin{array}{cc}
 \mathbf{I}_3   & - \hatop{ \left[ \mathbf{D} \oplus \mathbf{p} \right]}
\end{array}
\right)
\quad\quad\quad \text{(A $3 \times 6$ Jacobian)}
\end{eqnarray}

\subsection{Jacobian of $p \ominus (e^\varepsilon \oplus D) $}
\label{sect:jacob_eDp_inv}

This is the relative position of a point $\mathbf{p}$ relative to a pose $\mathbf{D}$,
an operation needed, for example, in Bundle Adjustment implementations \cite{triggs2000bundle}
(with the convention of points relative to the camera being $p \ominus D$, that is,
$D$ being the real position of the cameras).

Let $\mathbf{p} = [p_x ~ p_y ~ p_z]^\top \in \mathbb{R}^3$ be a 3D point, and
$\mathbf{D} \in \mathbf{SE}(3)$ be a pose with associated transformation matrix:

\begin{equation}
\mathbf{T}(\mathbf{D}) =
\left(
\begin{array}{ccc|c}
 d_{11} & d_{12} & d_{13} & d_{tx}   \\
 d_{21} & d_{22} & d_{23} & d_{ty}   \\
 d_{31} & d_{32} & d_{33} & d_{tz}   \\
\hline
  0 & 0 & 0 & 1
\end{array}
\right)
=
\left(
\begin{array}{ccc|c}
 \mathbf{d_{c1}}  & \mathbf{d_{c2}}  & \mathbf{d_{c3}}  & \mathbf{d_{t}}  \\
\hline
  0 & 0 & 0 & 1
\end{array}
\right)
=
\left(
\begin{array}{ccc|c}
   & \mathbf{R_D}  &  & \mathbf{d_{t}}  \\
\hline
  0 & 0 & 0 & 1
\end{array}
\right)
\end{equation}

We are interested in the derivative of $p \ominus (e^\E \oplus \mathbf{D})$ w.r.t $\E$:

\begin{eqnarray}
\label{eq:jacob_p_min_eD_e}
\left. \frac{\partial \mathbf{p} \ominus (e^\E \mathbf{D}) }{\partial \E} \right|_{\E = 0}
&=&
\left. \frac{\partial \mathbf{p} \ominus \mathbf{A} }{\partial \mathbf{A} } \right|_{\mathbf{A}=e^\E \mathbf{D} = \mathbf{D}}
\left. \frac{\partial e^\E \mathbf{D}}{\partial \E} \right|_{\E=0}
\nonumber \\
&=&
\left(
\begin{array}{c|c}
 \mathbf{I_3} \otimes \left( (\mathbf{p}-\mathbf{d_t})^\top \right)
 &
 -\mathbf{R_D}^\top
\end{array}
\right)
\left(
\begin{array}{cc}
 \mathbf{0}_{3\times 3}  & -\hatop{\mathbf{d}}_{c1} \\
 \mathbf{0}_{3\times 3}  & -\hatop{\mathbf{d}}_{c2} \\
 \mathbf{0}_{3\times 3}  & -\hatop{\mathbf{d}}_{c3} \\
 \mathbf{I}_{3}  & -\hatop{\mathbf{d}}_{t} \\
\end{array}
\right)
\quad\quad \text{\scriptsize{(Using Eq.(\ref{eq:jac_dp_min_A})  \& \S\ref{sect:jacob_eD}} )}
\nonumber \\
&=&
\left(
\begin{array}{cc}
   -\mathbf{R_D}^\top &
  \begin{array}{ccc}
  d_{21} p_z - d_{31} p_y & - d_{11} p_z + d_{31} p_x & d_{11} p_y - d_{21} p_x \\
  d_{22} p_z - d_{32} p_y & - d_{12} p_z + d_{32} p_x & d_{12} p_y - d_{22} p_x \\
  d_{23} p_z - d_{33} p_y & - d_{13} p_z + d_{33} p_x & d_{13} p_y - d_{23} p_x \\
  \end{array}
\end{array}
\right) \\
 && \quad\quad \quad\quad\quad \text{(A $3 \times 6$ Jacobian)} \nonumber
\end{eqnarray}


\subsection{Jacobian of $A \oplus e^\varepsilon \oplus D$}

Let $\A,\D \in \mathbf{SE}(3)$ be two poses, such as
$\D$ is defined as in the previous section, and $\mathbf{R}(\A)$ is
the $3\times 3$ rotation matrix associated to $\A$.

When optimizing a pose $\D$ which belongs to a sequence of chained poses
($\A \oplus \D$), we will need to evaluate:

\begin{eqnarray}
\label{eq:jacob.AeD}
\left. \frac{\partial \A e^\E \D}{\partial \E} \right|_{\E = 0}
&=&
\left. \frac{\partial \A \B}{\partial \B} \right|_{\B= e^0 \D = \D}
\left. \frac{\partial e^\E \mathbf{D}}{\partial \E} \right|_{\E = 0}
\\
&=&
\left[ \I_4 \otimes \mathbf{R}(\A) \right]
\left. \frac{\partial e^\E \mathbf{D}}{\partial \E} \right|_{\E = 0}
\\
&=&
\left(
\begin{array}{cc}
 \mathbf{0}_{3\times 3}  & -\mathbf{R}(\A) \hatop{\mathbf{d}}_{c1} \\
 \mathbf{0}_{3\times 3}  & -\mathbf{R}(\A) \hatop{\mathbf{d}}_{c2} \\
 \mathbf{0}_{3\times 3}  & -\mathbf{R}(\A) \hatop{\mathbf{d}}_{c3} \\
 \mathbf{R}(\A)          & -\mathbf{R}(\A) \hatop{\mathbf{d}}_{t} \\
\end{array}
\right)
\quad\quad\quad \text{(A $12 \times 6$ Jacobian)}
\end{eqnarray}

\subsection{Jacobian of $A \oplus e^\varepsilon \oplus D \oplus p$}
\label{eq:jacob_A_e_D_p}

This expression may appear in computer-vision problems,
such as in relative bundle-adjustment \cite{sibley2009rba}.
Let $\mathbf{p} \in \mathbb{R}^3$ be a 3D point
and $\A,\D \in \mathbf{SE}(3)$ be two poses, such as
$\mathbf{R}(\A)$ is the $3\times 3$ rotation matrix associated to $\A$ and
the rows and columns of $\D$ referred to as:

\begin{equation}
\mathbf{T}(\mathbf{D}) =
\left(
\begin{array}{ccc|c}
 \mathbf{d_{c1}}  & \mathbf{d_{c2}}  & \mathbf{d_{c3}}  & \mathbf{d_{t}}  \\
\hline
  0 & 0 & 0 & 1
\end{array}
\right)
=
\left(
\begin{array}{c|c}
    \mathbf{d_{r1}}^\top  & d_{tx}  \\
    \mathbf{d_{r2}}^\top  & d_{ty} \\
    \mathbf{d_{r3}}^\top  & d_{tz} \\
\hline
  \begin{array}{ccc}
    0  & 0  & 0
  \end{array}
    & 1
\end{array}
\right)
\end{equation}

Then, the Jacobian of the chained poses-point composition w.r.t.
the increment in the pose $\D$ (on the manifold) is:

\begin{eqnarray}
\left. \frac{\partial \A e^\E \D \mathbf{p}}{\partial \E} \right|_{\E = 0}
&=&
\mathbf{R}(\A)
\left(
\begin{array}{c|c}
  \I_3 &
    \begin{array}{ccc}
      0  & \mathbf{p} \cdot \mathbf{d_{r3}} + d_{tz}  & -(\mathbf{p} \cdot \mathbf{d_{r2}} + d_{ty}) \\
      -(\mathbf{p} \cdot \mathbf{d_{r3}} + d_{tz})  & 0 & \mathbf{p} \cdot \mathbf{d_{r1}} + d_{tx} \\
      \mathbf{p} \cdot \mathbf{d_{r2}} + d_{ty}  & -(\mathbf{p} \cdot \mathbf{d_{r1}} + d_{tx}) & 0
    \end{array}
\end{array}
\right)
\\
&&
\quad \quad \quad \quad \quad \quad \quad \quad \quad \quad \quad \quad
\quad \quad \quad \quad \quad \quad \quad \quad \quad \quad \quad \quad
\text{(A $3 \times 6$ Jacobian)} \nonumber
\end{eqnarray}

\noindent where $\mathbf{a} \cdot \mathbf{b}$ stands for the scalar product of vectors.
Note that for both $\A$ and $\D$ being very close to the identity in $\mathbf{SE}(3)$,
the following approximation can be used:

\begin{eqnarray}
\left. \frac{\partial \A e^\E \D \mathbf{p}}{\partial \E} \right|_{\E = 0}
& \approx &
\left(
\begin{array}{cc}
  \I_3 &
  - \hatop{\left[ \mathbf{p} + \mathbf{d_t} \right]}
\end{array}
\right)
\quad \quad \quad
\text{(A $3 \times 6$ Jacobian)} \nonumber
\end{eqnarray}

\subsection{Jacobian of $p \ominus (A \oplus e^\varepsilon \oplus D)$}
\label{eq:jacob_pmA_e_D}

This expression may also appear in computer-vision problems,
such as in relative bundle-adjustment \cite{sibley2009rba}.
Let $\mathbf{p} \in \mathbb{R}^3$ be a 3D point
and $\A,\D \in \mathbf{SE}(3)$ be two poses, such as
$\mathbf{R}(\A)$ is the $3\times 3$ rotation matrix associated to $\A$,
the rows and columns of $\D$ are referred to as in the previous section,
and:

\begin{equation}
\mathbf{T}(\A) \mathbf{T}(\D)
=
\left(
\begin{array}{c|c}
 \mathbf{R}(\A\D) & \mathbf{t_{AD}} \\
\hline
 \begin{array}{ccc} 0 & 0 & 0 \end{array} & 1
\end{array}
\right)
\end{equation}

Then, the Jacobian of interest can be obtained by chaining Eq.~\ref{eq:jacob.AeD} and Eq.~\ref{eq:jac_dp_min_A}:

\begin{eqnarray}
\left. \frac{\partial (\A e^\E \D)^{-1} \mathbf{p}}{\partial \E} \right|_{\E = 0}
&=&
\left[
\begin{array}{cc}
 \I_3 \otimes (\mathbf{p}-\mathbf{t_{AD}})^\top  & -\mathbf{R}(\A\D)^\top
\end{array}
\right]
\left(
\begin{array}{cc}
 \mathbf{0}_{3\times 3}  & -\mathbf{R}(\A) \hatop{\mathbf{d}}_{c1} \\
 \mathbf{0}_{3\times 3}  & -\mathbf{R}(\A) \hatop{\mathbf{d}}_{c2} \\
 \mathbf{0}_{3\times 3}  & -\mathbf{R}(\A) \hatop{\mathbf{d}}_{c3} \\
 \mathbf{R}(\A)        & -\mathbf{R}(\A) \hatop{\mathbf{d}}_{t} \\
\end{array}
\right)
~ \text{(A $3 \times 6$ Jacobian)} \nonumber
\\
~ 
\end{eqnarray}


\subsection{Jacobian of $((P_2 \oplus e^{\varepsilon 2}) \ominus (P_1 \oplus e^{\varepsilon 1})) \ominus D$}
\label{sect:jacob.DinvP2invP2}

While solving Graph-SLAM problems in SE(3), one needs to optimize the global poses $P_1$ and $P_2$ given a measurement $D$ of the relative pose or $P_2$ with respect to $P_1$, i.e.  $D=P_2 \ominus P_1$ or $\D = \Pone^{-1} \Ptwo$.
The corresponding error function to be minimized can be written as $(P_2 \ominus P_1) \ominus D$ or $\D^{-1} \Pone^{-1} \Ptwo$.
Therefore, we need the Jacobians of the latter expression with respect to 
manifold increments of $P_1$ and $P_2$. 

Normally, we take the logarithm of that error and then apply the vee operator to it 
to retrieve a 6-vector describing the error. 
Using the chain rule of Jacobians, we have:

\begin{eqnarray}
\left. \frac{\partial \log( \D^{-1} (\Pone e^{\E_1} )^{-1} \Ptwo )^\vee}{\partial \E_1} \right|_{\E_1 = 0}
&=&
\nonumber \\
\left. \frac{\partial \log( \D^{-1} e^{-\E_1} \Pone^{-1} \Ptwo )^\vee}{\partial \E_1} \right|_{\E_1 = 0}
&=&
\underbrace{
\left.
 \frac{\partial \log(\mathbf{T})^\vee }{\partial \mathbf{T}}
\right|_{T=\D^{-1} \Pone^{-1} \Ptwo}
}_{\text{See Eq.~\ref{eq:dLnSE3_wrt_SE3}}}
\underbrace{
\left. \frac{\partial f_\oplus(A,B) }{\partial A } \right|_{\begin{subarray}{l} A=\D^{-1} \\ B=\Pone^{-1}\Ptwo \end{subarray}}
}_{\text{See Eq.~\ref{eq:oplus.ab.wrt.a}}}
%
\underbrace{
\left(
\left. - \frac{\partial \D^{-1} e^{\E_1} }{\partial \E_1} \right|_{\E_1 = 0}
\right)
}_{\text{See Eq.~\ref{eq:jacob.d_eps_wrt_eps}}}
\nonumber
\\
\end{eqnarray}

\noindent and:

\begin{eqnarray}
\left. \frac{\partial \log( \D^{-1} \Pone ^{-1} \Ptwo e^{\E_2} )^\vee}{\partial \E_2} \right|_{\E_2 = 0}
&=&
\underbrace{
\left.
 \frac{\partial \log(\mathbf{T})^\vee }{\partial \mathbf{T}}
\right|_{T=\D^{-1} \Pone^{-1} \Ptwo}
}_{\text{See Eq.~\ref{eq:dLnSE3_wrt_SE3}}}
\underbrace{
\left. \frac{\partial A e^{\E_2} }{\partial \E_2} \right|_{\begin{subarray}{l} \E_2=0\\ A= \D^{-1} \Pone ^{-1} \Ptwo\end{subarray}}
}_{\text{See Eq.~\ref{eq:jacob.d_eps_wrt_eps}}}
\nonumber
\\
\end{eqnarray}

Both Jacobians above are $6 \times 6$.

\subsection{Jacobian of the SE(3) pseudo-logarithm}
\label{sect:eq:jacob_dLnSE3_dSE3}

Given the definition of SE(3) pseudo-logarithm
in \S\ref{sect:se3_pseudo-log}, 
this Jacobian can then be defined as simply:

\begin{equation}
\label{eq:dLnSE3_wrt_SE3}
\left.
 \frac{\partial \text{pseudo-log}(\mathbf{T})^\vee }{\partial \mathbf{T}}
\right|_{6 \times 12}
=
\left(
\begin{array}{cc}
 \mathbf{0}_{3 \times 9} & \mathbf{I}_3 \\
 \dfrac{\partial \log(\mathbf{R})^\vee }{\partial \mathbf{R}} & \mathbf{0}_{3 \times 3}
\end{array}
\right)
\end{equation}

\noindent where the Jacobian in Eq.~\ref{eq:dLnRot_wrt_R} has been used.

\appendix

\chapter{Applications to computer vision}
\label{ch:apx:cv}

This appendix provides some useful expressions related to (and making use of)
the Jacobian derived in chapters \S\S\ref{chap:mat_deriv}--\ref{ch:se3_optim}
which are useful
in computer vision applications.

\section{Projective model of an ideal pinhole camera -- $h(\mathbf{p})$}

Given a point $\mathbf{p} \in \mathbb{R}^3$ relative to a
projective camera, with the following convention for the axes of the
camera:

\begin{figure}[h!]
\centering
\includegraphics[width=0.50\textwidth]{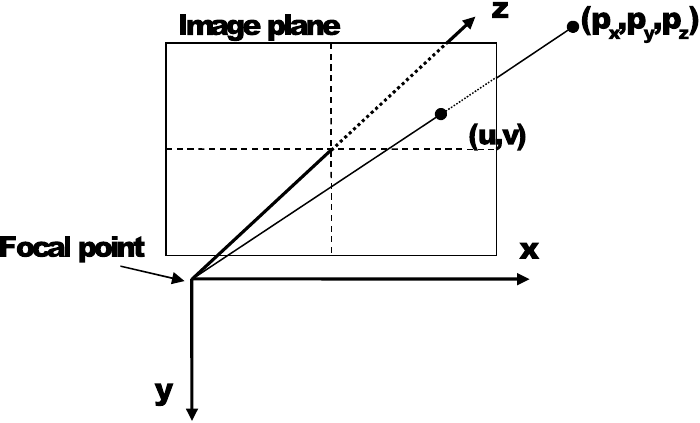}
\caption{The convention used in this report on the axes of a pinhole projective camera.}
\label{fig:pinhole}
\end{figure}

\noindent and given the $3 \times 3$ matrix of intrinsic camera parameters:

\begin{equation}
\mathbf{M}=
\left(
\begin{array}{ccc}
 f_x & 0 & c_x \\
  0  & f_y & c_y \\
 0 & 0 & 1 \\
\end{array}
\right)
\longrightarrow
\left\{
\begin{array}{l}
 f_x\text{: Focal distance, in 'x' pixel units.} \\
 f_y\text{: Focal distance, in 'y' pixel units.} \\
 c_x\text{: Image central point (x, in pixel units).} \\
 c_y\text{: Image central point (y, in pixel units).} \\
\end{array}
\right.
\end{equation}

\noindent then, the pixel coordinates $(u,v)$ of the projection of the
3D point $\mathbf{p}=[p_x ~ p_y ~ p_z]^\top$
is given (\emph{without} distortions) by the function $h: \mathbb{R}^3 \mapsto \mathbb{R}^2$,
with the well known expression:

\begin{equation}
h(\mathbf{p}) =
h\left(
\begin{array}{c}
 p_x \\ p_y \\p_z
\end{array}
\right) =
\left(
\begin{array}{c}
 c_x + f_x  \frac{p_x}{p_z} \\
 c_y + f_y  \frac{p_y}{p_z} \\
\end{array}
\right)
\end{equation}

In a number of computer vision problems we will need the Jacobian of this
projection function by the coordinates of the point w.r.t. the camera, which
is straightforward to obtain:

\begin{equation}
\label{eq:jacob_pinhole}
\frac{\partial h(\mathbf{p})}{\partial \mathbf{p}} =
\left(
\begin{array}{ccc}
 f_x/p_z  &    0    &  -f_x p_x/p_z^2   \\
 0 & f_y/p_z  &  -f_y p_y/ p_z^2
\end{array}
\right)
\end{equation}

\section{Projection of a point: $e^\varepsilon \oplus \mathbf{A} \oplus \mathbf{p}$}

Given a pose $\mathbf{A} \in \mathbf{SE}(3)$
(with rotation matrix denoted as $\mathbf{R_A}$)
and a point
$\mathbf{p} \in \mathbb{R}^3$
relative to that pose, we want here to derive the Jacobians of
the projection of $e^\E \oplus \A \oplus \mathbf{p}$ on a pinhole camera, that is,
of the expression $h(e^\E \oplus \A \oplus \mathbf{p})$.
Recall that $e^\E$ means the $\mathbf{SE}(3)$ Lie group exponentiation
of an auxiliary variable $\E$ which represents a small increment around $\A$ in the manifold.

Let $\mathbf{g}=[g_x ~ g_y ~ g_z]^\top$ denote $\mathbf{A} \oplus \mathbf{p}$.
Applying the chain rule of Jacobians and employing Eq.~(\ref{eq:jac_dAp_p}),
Eq.~(\ref{eq:jacob_eDp_e})
and  Eq.~(\ref{eq:jacob_pinhole})
we arrive at:

\begin{eqnarray}
\frac{\partial h(e^\E \oplus \A \oplus \mathbf{p})}{\partial \mathbf{p}}
&=&
\left. \frac{\partial h(\mathbf{p'})}{\partial \mathbf{p'}} \right|_{ \mathbf{p'} = \A \oplus \mathbf{p} = \mathbf{g} }
\frac{\partial e^\E \oplus \A \oplus \mathbf{p} }{\partial \mathbf{p}} \\
&=&
\left. \frac{\partial h(\mathbf{p'})}{\partial \mathbf{p'}} \right|_{ \mathbf{p'} = \A \oplus \mathbf{p} = \mathbf{g} }
\frac{\partial \A \oplus \mathbf{p} }{\partial \mathbf{p}} \\
&=&
\left(
\begin{array}{ccc}
 f_x / g_z   &    0    &  -f_x g_x / g_z^2   \\
 0 & f_y / g_z  &  -f_y g_y / g_z^2
\end{array}
\right)
\mathbf{R_A}
\quad\quad \text{(A $2 \times 3$ Jacobian)}
\end{eqnarray}

\noindent and:

\begin{eqnarray}
\frac{\partial h(e^\E \oplus \A \oplus \mathbf{p})}{\partial \mathbf{\E}}
&=&
\left. \frac{\partial h(\mathbf{p'})}{\partial \mathbf{p'}} \right|_{ \mathbf{p'} = \A \oplus \mathbf{p} = \mathbf{g} }
\frac{\partial e^\E \oplus \A \oplus \mathbf{p} }{\partial \E} \\
&=&
\left(
\begin{array}{ccc}
 f_x / g_z   &    0    &  -f_x g_x / g_z^2   \\
 0 & f_y / g_z  &  -f_y g_y / g_z^2
\end{array}
\right)
\left(
\begin{array}{cc}
 \mathbf{I}_3   & - \hatop{\left[ \mathbf{g} \right]}
\end{array}
\right) \\
&=&
\left(
\begin{array}{cccccc}
 \frac{f_x}{g_z}  &
 0 &
 -f_x\frac{g_x}{g_z^2} &
 -f_x \frac{g_x g_y}{g_z^2} &
  f_x  (1 + \frac{g_x^2}{g_z^2} )  &
 -f_x \frac{g_y}{g_z}
\\
 0 &
 \frac{f_y}{g_z}  &
 -f_y\frac{g_y}{g_z^2}
 &
  -f_y (1 + \frac{g_y^2}{g_z^2}) &
  f_y \frac{g_x g_y}{g_z^2} &
  f_y \frac{g_x}{g_z}
\end{array}
\right)
\quad\quad \text{(A $2 \times 6$ Jacobian)} \nonumber
\end{eqnarray}

\newpage

\section{Projection of a point: $\mathbf{p} \ominus (e^\varepsilon \oplus \mathbf{A})$}

The previous section Jacobians are applicable to optimization problems
where the convention is to estimate the inverse camera poses
(that is, the point to project, w.r.t. the camera, is $\A \oplus \mathbf{p}$).
In this section we address the alternative case of poses being the actual
camera positions
(that is, the point to project, w.r.t. the camera, is $\mathbf{p} \ominus \A$).

The expression we want to obtain the Jacobians of is in this case:
$h(\mathbf{p} \ominus (e^\E \oplus \A))$.
Using Eq.~(\ref{eq:jac_dp_min_p}),
and  Eq.~(\ref{eq:jacob_pinhole}),
and denoting $\mathbf{l}=[l_x ~ l_y ~ l_z]^\top=\mathbf{p} \ominus \A$,
we arrive at:

\begin{eqnarray}
\frac{\partial h(\mathbf{p} \ominus (e^\E \oplus \A) )}{\partial \mathbf{p}}
&=&
\left. \frac{\partial h(\mathbf{p'})}{\partial \mathbf{p'}} \right|_{ \mathbf{p'} = \mathbf{p} \ominus \A = \mathbf{l} }
\frac{\partial \mathbf{p} \ominus (e^\E \oplus \A) }{\partial \mathbf{p}} \\
&=&
\left. \frac{\partial h(\mathbf{p'})}{\partial \mathbf{p'}} \right|_{ \mathbf{p'} = \mathbf{p} \ominus \A = \mathbf{l} }
\frac{\partial \mathbf{p} \ominus \A }{\partial \mathbf{p}} \\
&=&
\left(
\begin{array}{ccc}
 f_x / l_z   &    0    &  -f_x l_x / l_z^2   \\
 0 & f_y / l_z  &  -f_y l_y / l_z^2
\end{array}
\right)
\mathbf{R_A^\top}
\quad\quad \text{(A $3 \times 3$ Jacobian)}
\end{eqnarray}

\noindent and:

\begin{eqnarray}
\frac{\partial h(\mathbf{p} \ominus (e^\E \oplus \A) )}{\partial \E}
&=&
\left. \frac{\partial h(\mathbf{p'})}{\partial \mathbf{p'}} \right|_{ \mathbf{p'} = \mathbf{p} \ominus \A }
\frac{\partial \mathbf{p} \ominus (e^\E \oplus \A) }{\partial \E} \\
&=&
\left(
\begin{array}{ccc}
 f_x / l_z   &    0    &  -f_x l_x / l_z^2   \\
 0 & f_y / l_z  &  -f_y l_y / l_z^2
\end{array}
\right)
\frac{\partial \mathbf{p} \ominus (e^\E \oplus \A) }{\partial \E}
\end{eqnarray}

\noindent with this last term given by Eq.~(\ref{eq:jacob_p_min_eD_e}).


\chapter{Expressions for SE(2) GraphSLAM}
\label{ch:apx:se2}

Poses in 2D, $SE(2)=\mathbb{R}^2 \times SO(2)$, have a much simpler structure than 
their three-dimensional counterparts, $SE(3)=\mathbb{R}^3 \times SO(3)$, therefore 
it is in order aiming at simpler, more efficient, expressions for solving SLAM problems in 2D.
This section explains the formulas used for SE(2) graph-SLAM within 
the MRPT framework.

\section{SE(2) definition}

A pose (rigid transformation) in two-dimensional Euclidean space can be uniquely determined by means
of a $3 \times 3$ homogeneous matrix with this structure:

\begin{equation}
\label{eq:se2:T_Rt}
 \mathbf{T}_2 =
\left(
\begin{array}{c|c}
  \mathbf{R}_2 & \mathbf{t} \\
\hline
  \mathbf{0}_{1\times 2} & 1
\end{array}
\right)
=
\left(
\begin{array}{cc|c}
\cos \phi & -\sin \phi & x \\
\sin \phi & \cos \phi & y \\
\hline
0 & 0 & 1
\end{array}
\right)
\end{equation}

\noindent where the three degrees of freedom of the 2D transformation are the $(x,y)$ translation
and the rotation of $\phi$ radians.

The $\mathbf{R}_2$ belongs to the group $SO(2)$, and $\mathbf{T}_2$ to $SE(2)$.

\section{Manifold local coordinates and retraction}

Just like we defined the exponential and logarithm map for SE(3) in \S\ref{sect:se3.exp.log},
we can define similar operations for SE(2).

Rigorously, the exponential and logarithm maps for SE(2) 
are defined as shown in \S\ref{sect:se2.exp}--\ref{sect:se2.exp} below, 
but in practice the simpler pseudo maps in \S\ref{sect:se2.ps-exp}--\ref{sect:se2.ps-log}
are more efficient to evaluate and work as a valid retraction and local coordinate map, respectively.
Therefore, the latter will be used in subsequent sections.

\subsection{SE(2) exponential map}
\label{sect:se2.exp}

Let

\begin{equation}
\label{eq:vector_in_se2}
\mathbf{v}= \left( \begin{array}{c} \mathbf{t}' \\ \phi \end{array} \right)
= \left( \begin{array}{c} x' \\ y' \\ \phi \end{array} \right) \in \mathfrak{se}(2)
\end{equation}

\noindent denote the 3-vector of local coordinates in the Lie algebra $\mathfrak{se}(2)$,
comprising a 2-vector $\mathbf{t}'$ for a translation, 
which is different than the actual plain SE(2) translation $\mathbf{t}=(x,y)$,
and a rotation $\phi$.
Next we define the $3 \times 3$ matrix:

\begin{equation}
 \mathbf{A}(\mathbf{v})=
\left(
\begin{array}{c|c}
 \hatop{[\phi]}  & \mathbf{t}' \\
 \hline
  0 & 0
\end{array}
\right)
=
\left(
\begin{array}{cc|c}
 0 & -\phi & x' \\
 \phi & 0 & y' \\
 \hline
 0 & 0 & 0
\end{array}
\right)
\end{equation}

Then, the map:

\begin{equation}
  \exp: \mathfrak{se}(2) \mapsto \mathbf{SE}(2)
\end{equation}

\noindent is well-defined, surjective, and has the closed form:

\begin{eqnarray}
  e^ { \mathbf{v} } \equiv  e^ { \mathbf{A}(\mathbf{v}) } =
\left(
\begin{array}{cc}
  e^{\hatop{[\phi]}} & \mathbf{V}_2 \mathbf{t}' \\
   0 & 1
\end{array}
\right)
\\
\label{eq:se2:V}
\mathbf{V}_2 = \mathbf{I_2}
+ \frac{1-\cos \phi}{\phi^2} \hatop{[\phi]}
+ \frac{\phi- \sin \phi}{\phi^3} (\hatop{[\phi]})^2 
\end{eqnarray}

\noindent with $e^{\hatop{[\phi]}}$ the matrix exponential 
and $\hatop{[\phi]}=\left( \begin{array}{cc} 0 & -\phi \\ \phi & 0\end{array}\right)$.

\subsection{SE(2) logarithm map}
\label{sect:se2.log}

The map:

\begin{eqnarray}
  \log: \mathbf{SE}(2) &\mapsto& \mathfrak{se}(2) \\
  \mathbf{A}(\mathbf{v})  & \mapsto & \mathbf{v}
\nonumber
\end{eqnarray}

\noindent is well-defined and can be computed as:

\begin{eqnarray}
\mathbf{v} &=&
\left(
\begin{array}{c}
 \mathbf{t}' \\ \phi
\end{array}
\right)
=
\left(
\begin{array}{c}
 x'\\y' \\ \phi
\end{array}
\right)
\nonumber \\
\mathbf{t} &=& \mathbf{V}_2^{-1} \mathbf{t}' \quad \quad \text{(with $\mathbf{V}_2$ in Eq.~\ref{eq:se2:V})}
\end{eqnarray}

Note that $\mathbf{V}_2^{-1}$ has a closed-form expression:

\begin{equation}
\mathbf{V}_2^{-1} = \mathbf{I}_2 
- \dfrac{1}{2} \hatop{[\phi]} + 
\dfrac{\left(
 1 - \dfrac{ \phi \cos(\phi/2)}{2 \sin(\phi / 2)}
\right) 
}{\phi^2}  (\hatop{[\phi]})^2
\end{equation}

\subsection{SE(2) pseudo-exponential map}
\label{sect:se2.ps-exp}

Since rotations in SE(2) are only parameterized by one scalar ($\phi$), 
it becomes more convenient to use a 3-vector to model local coordinates 
in the tangent space to the manifold, and to directly use $\mathbf{t}'=\mathbf{t}$
(see sections above). 
Therefore: 

\begin{eqnarray}
\label{eq:se2.pseudo-exp}
  \text{pseudo-exp}: \mathfrak{se}(2) &\mapsto& \mathbf{SE}(2) \\ 
  \left( \begin{array}{c} x' \\ y' \\ \phi \end{array} \right)
  &=&
  \left( \begin{array}{c} x \\ y \\ \phi \end{array} \right)
\end{eqnarray}

\noindent such that the Jacobian of the pseudo-exponential map 
becomes the identity:

\begin{equation}
	\frac{\partial \text{pseudo-exp}(\mathbf{v})}{\partial \mathbf{v}}
\equiv 
\mathbf{I}_3
\end{equation}

\subsection{SE(2) pseudo-logarithm map}
\label{sect:se2.ps-log}

As a consequence of the equations above, we define:

\begin{eqnarray}
\label{eq:se2.pseudo-log}
  \text{pseudo-log}: \mathbf{SE}(2)&\mapsto& \mathfrak{se}(2) \\ 
  \left( \begin{array}{c} x \\ y \\ \phi \end{array} \right)
  &=&
  \left( \begin{array}{c} x' \\ y' \\ \phi \end{array} \right)
\end{eqnarray}

\noindent whose Jacobian is also the identity:

\begin{equation}
\frac{\partial \text{pseudo-log}(\mathbf{T}_2)}{\partial \{x,y, \phi\}} = 
\mathbf{I}_3
\end{equation}

\subsection{SE(2) Jacobian of $D\boxplus \varepsilon = D \oplus e^\varepsilon$  (right-multiply option)}
\label{sect:se2.jacob_De}

Let $\mathbf{D} \in \mathbf{SE}(2)$ be the 2D pose $(x_D,y_D,\phi_D)$, 
and $\E=(\E_x ~ \E_y ~ \E_{\phi})^\top$ an increment on $\mathfrak{se}(2)$. 
We are interested in the derivative of $\mathbf{D} \oplus e^\E$ w.r.t $\E$, 
which can be shown to be (expanding the multiplication of the corresponding matrices):

\begin{eqnarray}
\label{eq:se2.dDexpe_de}
\left. \frac{\partial \mathbf{D e^\E}}{\partial \E} \right|_{\E = 0}
=
\left(
\begin{array}{ccc}
\cos \phi_D & -\sin \phi_D & 0 \\
\sin \phi_D & \cos \phi_D & 0 \\
0 & 0 & 1
\end{array}
\right)_{3 \times 3}
\end{eqnarray}

\subsection{Jacobians for SE(2) pose composition $A \oplus B$}
\label{sect:se2.jacob_AB}

Let $\mathbf{A}, \mathbf{B} \in \mathbf{SE}(2)$ be the 2D poses 
$(x_A,y_A,\phi_A)$, 
and
$(x_B,y_B,\phi_B)$, respectively. 
We are interested in the derivatives of the composed pose $A \oplus B$ w.r.t both poses. 
By expanding the matrix products it is easy to show that: 

\begin{eqnarray}
\label{eq:se2.dAB_dA}
\frac{\partial f_\oplus(A,B)}{\partial \A}
=
\left(
\begin{array}{ccc}
1 & 0 & -x_B \sin \phi_A - y_B \cos \phi_A \\
0 & 1 &  x_B \cos \phi_A - y_B \sin \phi_A \\
0 & 0 & 1
\end{array}
\right)_{3 \times 3}
\end{eqnarray}

\noindent and:

\begin{eqnarray}
\label{eq:se2.dAB_dB}
\frac{\partial f_\oplus(A,B)}{\partial \B}
=
\left(
\begin{array}{ccc}
\cos \phi_A & -\sin \phi_A & 0  \\
\sin \phi_A & \cos \phi_A & 0 \\
0 & 0 & 1
\end{array}
\right)_{3 \times 3}
\end{eqnarray}

\subsection{SE(2) Jacobian of $((P_2 \oplus e^{\varepsilon 2}) \ominus (P_1 \oplus e^{\varepsilon 1})) \ominus D$}
\label{sect:se2:jacob.DinvP2invP2}

While solving Graph-SLAM in SE(2), we need to optimize the global poses $P_1$ and $P_2$ given a measurement $D$ of the relative pose or $P_2$ with respect to $P_1$, i.e.  $D=P_2 \ominus P_1$ or $\D = \Pone^{-1} \Ptwo$.
The corresponding error function to be minimized can be written as $(P_2 \ominus P_1) \ominus D$ or $\D^{-1} \Pone^{-1} \Ptwo$.
Therefore, we need the Jacobians of the latter expression with respect to 
manifold increments of $P_1$ and $P_2$. 
For SE(2), we will assume that the error vector is the pseudo-logarithm of the pose mismatch above.

Using the chain rule of Jacobians, we have:

\begin{eqnarray}
\left. \frac{\partial \log( \D^{-1} (\Pone e^{\E_1} )^{-1} \Ptwo )^\vee}{\partial \E_1} \right|_{\E_1 = 0}
&=&
\nonumber \\
\left. \frac{\partial \log( \D^{-1} e^{-\E_1} \Pone^{-1} \Ptwo )^\vee}{\partial \E_1} \right|_{\E_1 = 0}
&=&
\cancelto{\mathbf{I}_3}{
\left.
 \frac{\partial \log(\mathbf{T_2})^\vee }{\partial \mathbf{T}}
\right|_{T=\D^{-1} \Pone^{-1} \Ptwo}
}
\quad
\underbrace{
\left. \frac{\partial f_\oplus(A,B) }{\partial A } \right|_{\begin{subarray}{l} A=\D^{-1} \\ B=\Pone^{-1}\Ptwo \end{subarray}}
}_{\text{See Eq.~\ref{eq:se2.dAB_dA}}}
\quad
\underbrace{
\left(
\left. - \frac{\partial \D^{-1} e^{\E_1} }{\partial \E_1} \right|_{\E_1 = 0}
\right)
}_{\text{See Eq.~\ref{eq:se2.dDexpe_de}}}
\nonumber
\\
\end{eqnarray}

\noindent and:

\begin{eqnarray}
\left. \frac{\partial \log( \D^{-1} \Pone ^{-1} \Ptwo e^{\E_2} )^\vee}{\partial \E_2} \right|_{\E_2 = 0}
&=&
\cancelto{\mathbf{I}_3}{
\left.
 \frac{\partial \log(\mathbf{T_2})^\vee }{\partial \mathbf{T}}
\right|_{T=\D^{-1} \Pone^{-1} \Ptwo}
}
\quad
\underbrace{
\left. \frac{\partial A e^{\E_2} }{\partial \E_2} \right|_{\begin{subarray}{l} \E_2=0\\ A= \D^{-1} \Pone ^{-1} \Ptwo\end{subarray}}
}_{\text{See Eq.~\ref{eq:se2.dDexpe_de}}}
\nonumber
\\
\end{eqnarray}

\bibliographystyle{plain}
\bibliography{myReferences}

\end{document}